\documentclass[10pt,journal, twocolumn]{IEEEtran}
\ifCLASSOPTIONcompsoc
\else
\fi

\ifCLASSINFOpdf
\else
\fi

\usepackage[table]{xcolor}
\usepackage{times}
\usepackage{epsfig}
\usepackage{graphicx}
\usepackage{amsmath}
\usepackage{amssymb}
\usepackage{epsfig}
\usepackage{graphicx}	
\usepackage{amsmath}
\usepackage{amssymb}
\usepackage{bbm}
\usepackage{pifont}% http://ctan.org/pkg/pifont

\usepackage{array,setspace, amsfonts,url,bm, cellspace}
\usepackage{tikz}
\usepackage{epstopdf}
\usetikzlibrary{positioning}
\usetikzlibrary{arrows}
\usepackage{pifont}
\usepackage{multirow, hhline}
\usetikzlibrary{decorations.text}
\usetikzlibrary{shapes.geometric,calc}
\definecolor{mygray}{RGB}{208,208,208}
\definecolor{mymagenta}{RGB}{226,0,116}

\newcolumntype{M}[1]{>{\centering\arraybackslash}m{#1}}

\newcolumntype{C}[1]{>{\centering\let\newline\\\arraybackslash\hspace{0pt}}m{#1}}

\usepackage[pagebackref=true,breaklinks=true,letterpaper=true,colorlinks,bookmarks=false]{hyperref}

\newcommand{\etal}{\textit{et al}.}

\begin{document}

\title{The UU-Net: Reversible Face De-Identification for Visual Surveillance Video Footage}

\author{\IEEEauthorblockN{Hugo Proen\c{c}a,~\IEEEmembership{Senior Member,~IEEE}}
\IEEEcompsocitemizethanks{H. Proen\c{c}a is with the IT: Instituto de Telecomunica\c{c}\~{o}es, Department of Computer Science, University of Beira Interior, Portugal, E-mail: hugomcp@di.ubi.pt}
\thanks{Manuscript received ? ?, 2020; revised ? ?, ?.}}

% The paper headers
%\markboth{IEEE TRANSACTIONS ON CIRCUITS AND SYSTEMS FOR VIDEO TECHNOLOGY,~Vol.~??, No.~??, ??~2020}%
\markboth{arxiv,~2020}%
{Shell \MakeLowercase{\textit{et al.}}: Bare Demo of IEEEtran.cls for Computer Society Journals}

\IEEEtitleabstractindextext{

\begin{abstract}
We propose a reversible face de-identification method for low resolution video data, where landmark-based techniques cannot be reliably used. Our solution is able to generate a photo realistic de-identified stream that meets the data protection regulations and can be publicly released under minimal privacy constraints. Notably, such stream encapsulates all the information required to later reconstruct the original scene, which is useful for scenarios, such as crime investigation, where the identification of the subjects is of most importance. We describe a learning process that jointly optimizes two main components: 1) a \emph{public} module, that receives the raw data and generates the de-identified stream, where the ID information is surrogated in a photo-realistic and seamless way; and 2) a \emph{private} module, designed for legal/security authorities, that analyses the public stream and reconstructs the original scene, disclosing the actual IDs of all the subjects in the scene. The proposed solution is landmarks-free and uses a conditional generative adversarial network to generate synthetic faces that preserve pose, lighting, background information and even facial expressions. Also, we enable full control over the set of soft facial attributes that should be preserved between the raw and de-identified data, which broads the range of  applications for this solution. Our experiments were conducted in three different visual surveillance datasets (BIODI, MARS and P-DESTRE) and showed highly encouraging results. The source code is available at \url{https://github.com/hugomcp/uu-net}.
\end{abstract}

}

\maketitle

\IEEEdisplaynontitleabstractindextext

\IEEEpeerreviewmaketitle

\section{Introduction}

\IEEEPARstart{V}ideo-based surveillance regards \emph{the act of watching a person or a place, esp. a person believed to be involved with criminal activity or a place where criminals gather}\footnote{\url{https://dictionary.cambridge.org/dictionary/english/surveillance}}. While this kind of technologies has been sustaining the growth of social monitoring and control tools, it also hosts crime prevention measures throughout the world,  raising the debates about proper security/privacy balance solutions~\cite{Senior2009}.

Anonymising publicly-recorded video streams has been regarded as a potential solution to privacy issues, in order to comply with  data privacy regulations like GDPR\footnote{\url{https://eur-lex.europa.eu/eli/reg/2016/679/oj}} and CCPA\footnote{\url{https://privacyrights.org/resources/california-consumer-privacy-act-basics}}. In this context, the earliest de-identification techniques obfuscated privacy-sensitive identity information by low-level image processing operations, such as downsampling, blurring or masking (e.g.,~\cite{Boyle2000} and~\cite{Neustaedter2006}). However, these methods also destroy any privacy-insensitive information and decrease the photo realism of the resulting streams, which compromises their utility.

Recently, more sophisticated de-identification techniques were proposed (e.g.,~\cite{Dufaux2008},~\cite{Maximov2020} and~\cite{Samarzija2014}) based in active appearance models (AAMs) and facial landmarks. In particular, conditional Generative Adversarial Networks (cGANs) have become a popular way to control the appearance of the synthesised data, with various applications reported in the literature, from cross-domain/view image synthesis, text-to-image translation to fashion synthesis (e.g., \cite{Isola2017}, \cite{Regmi2018}, \cite{Zhang2017} and~\cite{Zhu2017}).

\begin{figure}[ht!]
\begin{center}
\begin{tikzpicture}

\draw [black, rounded corners, thick] (-2.15, -2.75) rectangle (6.5, 2.1);

\draw (-1.0,0.9-0.15) node(n1)  {\includegraphics[width=1.0 cm]{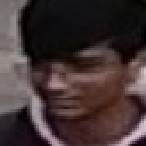}};   
\draw (-1.15,0.75-0.15) node(n1)  {\includegraphics[width=1.0 cm]{imgs/illustration_x1}};   
\draw (-1.3,0.6-0.15) node(n1)  {\includegraphics[width=1.0 cm]{imgs/illustration_x1}};     
  
\draw (2.35,0.9-0.15) node(n1)  {\includegraphics[width=1.0 cm]{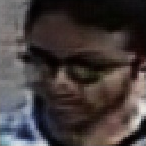}};    
\draw (2.2,0.75-0.15) node(n1)  {\includegraphics[width=1.0 cm]{imgs/illustration_a1}};    
\draw (2.05,0.6-0.15) node(n1)  {\includegraphics[width=1.0 cm]{imgs/illustration_a1}};    

\draw (5.7, 0.9-0.15) node(n1)  {\includegraphics[width=1.0 cm]{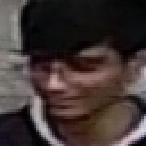}};    
\draw (5.55, 0.75-0.15) node(n1)  {\includegraphics[width=1.0 cm]{imgs/illustration_r1}};     
\draw (5.4, 0.6-0.15) node(n1)  {\includegraphics[width=1.0 cm]{imgs/illustration_r1}};     

\def\deltaX{0.1}

\fill [black!60!green] (0+\deltaX, 0) rectangle (0.1+\deltaX, 1.25);     
\fill [black!60!green] (0.125+\deltaX, 0) rectangle (0.225+\deltaX, 0.95); 
\fill [black!60!green] (0.25+\deltaX, 0) rectangle (0.35+\deltaX, 0.75); 
\fill [black!60!green] (0.375+\deltaX, 0) rectangle (0.475+\deltaX, 0.45); 
\fill [black!60!green] (0.5+\deltaX, 0) rectangle (0.6+\deltaX, 0.75); 
\fill [black!60!green] (0.625+\deltaX, 0) rectangle (0.725+\deltaX, 0.95); 
\fill [black!60!green] (0.750+\deltaX, 0) rectangle (0.850+\deltaX, 1.25);     
\draw [-] (0.11+\deltaX, 1.1) -- (0.74+\deltaX, 1.1);
\draw [-] (0.235+\deltaX, 0.85) -- (0.615+\deltaX, 0.85);
\draw [-] (0.36+\deltaX, 0.65) -- (0.49+\deltaX, 0.65);
\draw [black, fill=white] (0.125+\deltaX, 0.075) rectangle (0.725+\deltaX, 0.375);     
\draw (0.5+\deltaX, 0.225) node[rectangle] {\scriptsize{\textbf{U}$_{e}$}};

\draw  [black] (-1.15, 1.8) node {\scriptsize{\textbf{Raw Stream}}};    
\draw  [black] (-1.15, 1.55) node {\scriptsize{\textbf{x}$_{i,j,k}$}};    
\draw  [black] (2.2, 1.8) node {\scriptsize{\textbf{De-identified (Published) Stream}}};    
\draw  [black] (2.2, 1.55) node {\scriptsize{\textbf{a}$_{i,j,k}$}};    
\draw  [black] (5.55, 1.8) node {\scriptsize{\textbf{Reversed Stream}}};    
\draw  [black] (5.55, 1.55) node {\scriptsize{\textbf{r$_{i,j,k}$}}};

\draw [very thick, ->] (-0.4, 0.4) -- (0, 0.4);
\draw [very thick, ->] (1.05, 0.4) -- (1.45, 0.4);
\draw [very thick, ->] (2.95, 0.4) -- (3.35, 0.4);
\draw [very thick, ->] (4.4, 0.4) -- (4.8, 0.4);

%Comparators
\draw [dashed] (-1.5, -0.5) -- (2.0, -0.5);
\draw  (2, -0.25) -- (2.0, -0.5);
\draw  (-1.5, -0.25) -- (-1.5, -0.5);

\draw [dashed] (-1.05, -0.35) -- (5.5, -0.35);
\draw  (5.5, -0.25) -- (5.5, -0.35);
\draw  (-1.05, -0.25) -- (-1.05, -0.35);

\draw [dashed] (2.6, -0.3) -- (3.05, 0.15);
\draw [dashed] (2.95, 0.25) -- (3.05, 0.15);
\draw [dashed] (2.6, -0.3) -- (2.5, -0.2);

\draw [fill, black](-0.5, -0.75)circle(0.13);
\draw [white] (-0.5,-0.75) node {\scriptsize{1}};          

\draw [fill, black](-0.0, -0.75)circle(0.13);
\draw [white] (-0.0,-0.75) node {\scriptsize{5}};        

\draw [fill, black](0.5, -0.75)circle(0.13);
\draw [white] (0.5,-0.75) node {\scriptsize{6}};

\draw [fill, black](4.5, -0.6)circle(0.13);
\draw [white] (4.5,-0.6) node {\scriptsize{2}};

\draw [fill, black](2.95, -0.35) circle(0.13);
\node at (2.95, -0.35) [white, rotate=45]  {\scriptsize{\textbf{3}}};      

\draw [fill, black](3.15, -0.1) circle(0.13);
\node at (3.15, -0.1) [white, rotate=45]  {\scriptsize{\textbf{4}}};      

\def\deltaY{-0.75}

\draw [fill, black](-1.9,-0.6+\deltaY)circle(0.13);
\draw [white] (-1.9,-0.6+\deltaY) node {\scriptsize{1}};          
\draw  [black] (-0.0, -0.6+\deltaY) node {\scriptsize{Anonymization: ID(\textbf{x}$_.$) $\neq$ ID(\textbf{a}$_.$)}};    

\draw [fill, black](2.05,-0.6-0.5+\deltaY)circle(0.13);
\draw [white] (2.05,-0.6-0.5+\deltaY) node {\scriptsize{4}};       
\draw  [black] (4.375, -0.6-0.5+\deltaY) node {\scriptsize{Temp. Consist.: ID(\textbf{a}$_{i,j,k}$) = ID(\textbf{a}$_{i,j,k'}$)}};    

\draw [fill, black](-1.9,-0.6-0.5+\deltaY)circle(0.13);
\draw [white] (-1.9,-0.6-0.5+\deltaY) node {\scriptsize{3}};       
\draw  [black] (0.075, -0.6-0.5+\deltaY) node {\scriptsize{Diversity : ID(\textbf{a}$_{i,.,.}$) $\neq$ ID(\textbf{a}$_{j,.,.}$)}};    

\draw [fill, black](2.05,-0.6-0.0+\deltaY)circle(0.13);
\draw [white] (2.05,-0.6-0.0+\deltaY) node {\scriptsize{2}};       
\draw  [black] (4.15, -0.6-0.0+\deltaY) node {\scriptsize{Reversibility: \textbf{r}$_.$ = \textbf{U}$_d$\Big(\textbf{U}$_e$(\textbf{x}$_.$)\Big) = \textbf{x}$_.$}};    

\draw [fill, black](-1.9,-0.6-1+\deltaY)circle(0.13);
\draw [white] (-1.9,-0.6-1+\deltaY) node {\scriptsize{5}};       
\draw  [black] (0.075, -0.6-1+\deltaY) node {\scriptsize{Pose Consistency: (\textbf{x$_{i,j,k}$, a$_{i,j,k}$})}};

\draw [fill, black](2.05,-0.6-1+\deltaY)circle(0.13);
\draw [white] (2.05,-0.6-1+\deltaY) node {\scriptsize{6}};       
\draw  [black] (4.275, -0.6-1+\deltaY) node {\scriptsize{Background Consist. : (\textbf{x$_{i,j,k}$, a$_{i,j,k}$})}};    

\def\deltaX{3.5}

\fill [black!60!red] (0+\deltaX, 0) rectangle (0.1+\deltaX, 1.25);     
\fill [black!60!red] (0.125+\deltaX, 0) rectangle (0.225+\deltaX, 0.95); 
\fill [black!60!red] (0.25+\deltaX, 0) rectangle (0.35+\deltaX, 0.75); 
\fill [black!60!red] (0.375+\deltaX, 0) rectangle (0.475+\deltaX, 0.45); 
\fill [black!60!red] (0.5+\deltaX, 0) rectangle (0.6+\deltaX, 0.75); 
\fill [black!60!red] (0.625+\deltaX, 0) rectangle (0.725+\deltaX, 0.95); 
\fill [black!60!red] (0.750+\deltaX, 0) rectangle (0.850+\deltaX, 1.25);     
\draw [-] (0.11+\deltaX, 1.1) -- (0.74+\deltaX, 1.1);
\draw [-] (0.235+\deltaX, 0.85) -- (0.615+\deltaX, 0.85);
\draw [-] (0.36+\deltaX, 0.65) -- (0.49+\deltaX, 0.65);
\draw [black, fill=white] (0.125+\deltaX, 0.075) rectangle (0.725+\deltaX, 0.375);     
\draw (0.5+\deltaX, 0.225) node[rectangle] {\scriptsize{\textbf{U}$_{d}$}};  

\end{tikzpicture}
    \caption{Key properties demanded to a reversible video de-identifier: the identity of every face in the public stream should be surrogated (1), while keeping pose (5) and background (6) information to assure seamless transitions and photo realism. Also, the de-identified faces should be diverse among identities (3) and consistent (4) across multiple frames of a sequence. Finally, it should be possible to reconstruct the original stream (2) exclusively based in the publicly released data.}
        \label{fig:key}
    \end{center}
\end{figure}

As illustrated in Fig.~\ref{fig:key}, reversible video de-identification is a challenging task. Not only the original stream needs to be seamless modified, keeping concerns about distortions or other visual artefacts, but also the identity of the subjects must be obfuscated in a visually pleasant way, while considering constraints such as the background information, pose and lighting conditions. 

In this work, we consider that \emph{faces} are the most sensitive identifiers in public data. We propose a reversible video face de-identification solution based in a two-phase adversarial learning process. In inference time, our model is decomposed into two disjoint parts: 1) an ~\emph{encoder} \textbf{U}$_{e}$ that receives the raw data and generates their de-identified version, ensuring IDs de-identification and temporal consistency, while preserving pose, lighting, background information and even facial expressions. This produces the \emph{public} stream, guaranteeing its usability for analytics tools and social media; and 2) a \emph{decoder} \textbf{U}$_{d}$ that is available only to authorities and reconstructs the original scene exclusively based in the publicly available stream. Note that neither the original stream nor any kind of sensitive meta-information are ever stored or transmitted over the network, assuring the individuals' right to privacy (for public exposure purposes), while still enabling the disclosure of the actual IDs in a crime scene.   

The root of our solution is a cGAN composed of two sequential U-shaped models~\cite{Ronneberger2015} (hence UU-Net), used respectively for de-identification/reconstruction purposes. At first, a multi-label pairwise CNN classifier is inferred, to perceive the matching labels (ID, \emph{gender}, \emph{ethnicity}, \emph{hairstyle} and \emph{age}) between pairs of images. This network is further used in the main learning phase, which works under the adversarial learning paradigm:  the \emph{generator} network considers the pairwise responses and attempts to fool a \emph{PatchGAN}~\cite{Isola2017}  discriminator, responsible to distinguish between the raw, anonymised and reconstructed faces. Depending of a weights (\emph{positive}, \emph{negative} ou \emph{null}) given to every component of the pairwise discriminator responses, we keep full control over the appearance of the anonymised faces,  i.e., we are able to determine the labels that should agree/disagree/be independent between the raw and anonymised faces. 

Considering that our method was designed to work in poor resolution data, we kept it \emph{landmarks-free} and independent of a previous face alignment based in fiducial points. Instead, it exclusively depends of a low-resolution face detector (e.g.,~\cite{Ren2017}~\cite{Deng2019}). Then, in inference time, once the generator \textbf{U}$_{e}$ creates the anonymised faces, we use image steganography~\cite{Denemark2016} to seamlessly hide information of the bounding boxes in the public  stream. This information is used for reconstruction purposes, to define the regions-of-interest that will be reversed by the decoder \textbf{U}$_{d}$. In summary, we provide the following contributions:

\begin{itemize}
\item we propose a two-stage learning process and a simple architecture to de-identify sequences of facial images in poor resolution video streams;
\item based on the responses provided by an image pairwise analyzer, we keep full control over the properties of the anonymized faces, which broads the range of potential applications for our solution; 
\item using image steganography techniques, we encapsulate the anonymised faces and the corresponding regions-of-interests in the public video stream, that can be released without compromising the individuals' right to privacy in public spaces; 
\item based exclusively in the publicly available data, the second part of our model is able to reconstruct the original scenes and disclosure the actual ID of the subjects in the scene. The idea is that this module is exclusively available to legal authorities, in crime scene investigation scenarios.\\
\end{itemize}

The remainder of this paper is organized as follows: Section~\ref{sec:Related} summarizes the most relevant research in the scope of the paper. Section~\ref{sec:Proposed} provides a detailed description of the proposed method. Section~\ref{sec:Results} discusses the results of our empirical evaluation, and the conclusions are given in Section~\ref{sec:Conclusions}.

\section{Related Work}
\label{sec:Related}

This section summarizes the existing works in the image-based/video face de-identification context.

\subsection{Image-Based Face De-identification}

The earliest methods used simple image processing operations, such as blacking-out, pixelation or blurring (e.g.,~\cite{Boyle2000} and~\cite{Neustaedter2006}), and yielded poor realistic anonymised faces. Later, Blanz \etal~\cite{Blanz2004} estimated the shape, pose and illumination in pairs of faces, and fitted morphable 3-D models to each one, rendering new faces by transferring parameters between a source and a target model. \cite{Phillips2005} proposed an eigenvector-based solution in which faces are reconstructed by a fraction of the \emph{eigenface} vectors, such that ID information is lost. Similarly, Seo \etal~\cite{Seo2008}'s method was based on watermarking, hashing and PCA representations of data. Bitouk \etal~\cite{Bitouk2008} proposed a method that replaces a target by a gallery element, selected according to its similarity to the query. Gross \etal~\cite{Gross2009} used multi-factor models that unify linear, bilinear and quadratic data fitting solutions, but requiring a AAM to provide landmarks information.

The \emph{k}-Same algorithm~\cite{Newton2005}  provided the rationale for various techniques. Considering the \emph{k}-anonymity model~\cite{Sweeney2002}, linear combinations of  the gallery elements are obtained per probe, creating realistic anonymised data that depends on the alignment between the gallery/probe elements. Du \etal~\cite{Du2014} used samples of the gallery data to change the query image, obtianing "average" faces that lack in terms of photo realism. There are various recent methods still based in this concept, such as the k-Same-Net~\cite{Meden2018} and the attributes preserving approaches due to Jourabloo \etal~\cite{Jourabloo2015} and Yan \etal~\cite{Yan2019}.

Upon the deep learning breakthrough, Korshunova \etal~\cite{Korshunova2017} learned one generative model per identity. Also, by restricting the output patches to gallery elements of the same identity, this solution limits the variability of the results.  Starting from segmented human silhouettes, Brkic \etal~\cite{Brkic2017} proposed a model where obfuscation depends on the masked input data used. \emph{DeepPrivacy}~\cite{Hukkelas2019} anonymises facial images while retaining the original data distribution, for photo realism purposes. This model uses a cGAN, the central part of the face and pose information that is encoded in additional channels of the input data. Li and Lyu~\cite{Li2019} used a face attribute transfer model to preserve the consistency of non-identity attributes between the input and anonymised data. Sun \etal~\cite{Sun2018} combined parametric face synthesis techniques and GANs, keeping control over the facial parameters while adding fine details and realism into the resulting images. This approach depends of a computationally demanding face alignment step. Yamac \etal~\cite{Yamac2019} introduced a reversible privacy-preserving compression method, that combines multi-level encryption with compressive sensing. Finally,  Gu \etal~\cite{Gu2019} described a generative adversarial learning scheme based in image data and passwords that are fed to the models by  additional input channels. The idea is to train a generative model that reconstructs the original input only when the original password is given as input of the reversal phase. 

\subsection{Face De-identification in Videos}

The earliest approach was due to Dufaux \etal~\cite{Dufaux2008}, which scrambled the quantized transform coefficient of 4 $\times$ 4 facial blocks by random flipping/permutations, allowing reversibility but completely failing in photo realism. Agrawal and Narayanan~\cite{Agrawal2011} performed 3D segmentation (in space and time) and then blurred data in both domains to prevent reversal. Dale \etal~\cite{Dale2011} used 3D multilinear models to track the facial appearance of pairs of source/target videos. Using 3D geometry, they warp the source to the target face and refined the source to match the target appearance, keeping concerns about the temporal consistency of the result. Samarzija and Ribaric~\cite{Samarzija2014} grouped the gallery faces according to ID/pose information, representing each cluster by an AAM. Queries are matched to each AAM and the best match is used as anonymised data.

Ren \etal~\cite{Ren2018} proposed a GAN-based video face anonymizer where the de-identified versions of the input are obtained such that preserve action information. Gafni \etal~\cite{Gafni2019} proposed a feed-forward encoder-decoder architecture that fuses the input data to the masked output of a U-net backbone. This method requires facial landmarks to produce visually acceptable high resolution data. Sun \etal~\cite{Sun2018b} proposed a GAN-based solution that partially changes the face texture, according to facial landmarks and head pose information that are given as input. Bao \etal~\cite{Bao2018} proposed a GAN-based framework to synthesise a face from two input images, one used for identity and the other for style attributes preservation. Similarly, Shen and Liu~\cite{Shen2017} described a solution for editing facial attributes that is much similar to He \etal~\cite{He2019}'s solution. However, both solutions lack in terms of the temporal consistency across the different frames. Maximov \etal~\cite{Maximov2020} proposed the CIAGAN, also based on conditional GANs, to obtain de-identified versions of the input, while keeping control of the soft biometric features of the output data. This method produces realistic high resolution images, yet it also requires the availability of facial landmarks for proper alignment.

There are also various examples for real-time video privacy protection, where anonymity is assured by face masking~\cite{Schiff2009}, cryptographic obscuration~\cite{Chattopadhyay2007}, encryption~\cite{Winkler2010} or blurring~\cite{Mrityunjay2011}.   

\subsection{Face Attributes Transfer}

Face de-identification is closely related to transferring (swapping) attributes in human faces, which has been also motivating several works.  Zong \etal~\cite{Zhong2016} used mid-level features of a CNN as disentangled representations of facial features, while Cao \etal~\cite{Cao2018} proposed a Multi-task CNN with partially shared layers that learn facial attributes. Xiao \etal~\cite{Xiao2018} proposed the ELEGANT framework, that infers a disentangled representation in a latent space, where the various components refer different facial attributes in a quasi-independent way. The Attribute-GAN~\cite{He2019} and Style-GAN~\cite{Karras2019} are other good examples of generative models that change facial features, while retaining all the other types of information. As described in the next section, the method proposed in this paper inherits several insights from both~\cite{He2019} and~\cite{Karras2019}. 

\section{The UU-Net: Reversible Face De-Identification in Video Data}
\label{sec:Proposed}

\begin{figure*}[ht!]
\begin{center}
\begin{tikzpicture}

\draw [black, rounded corners, thick] (0, 0) rectangle (18.0, -10.1);     

\draw  [black] (2.75, -0.2) node {\small{\textbf{Learning Phase 1}}};    

%%%%%%%%%%%%%%%%%%%%%%%%%%%%%%%%%%%%%%%%%%%%
%	RECOGNIZER

\def\posX{0.7}
\def\posY{-1.05}
\draw (\posX,\posY) node(n1)  {\includegraphics[width=0.75 cm]{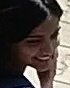}};   
\draw [black!30!yellow, very thick] (\posX-0.39, \posY+0.49) rectangle (\posX+0.39, \posY-0.49);

\def\posX{0.7}
\def\posY{-2.15}
\draw (\posX,\posY) node(n2)  {\includegraphics[width=0.75 cm]{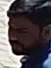}};   
\draw [black!30!yellow, thick] (\posX-0.39, \posY+0.49) rectangle (\posX+0.39, \posY-0.49);    

\def\posX{1.75}
\def\posY{-2.2}
\fill [yellow!20!red] (\posX, \posY) rectangle (\posX+0.1, \posY+1.25);     
\fill [yellow!20!red] (\posX+0.125, \posY+0.1) rectangle (\posX+0.225, \posY+1.15); 
\fill [yellow!20!red] (\posX+0.25, \posY+0.2) rectangle (\posX+0.35, \posY+1.05); 
\fill [yellow!20!red] (\posX+0.375, \posY+0.3) rectangle (\posX+0.475, \posY+0.95); 
\fill [yellow!20!red] (\posX+0.5, \posY+0.4) rectangle (\posX+0.6, \posY+0.85); 
\fill [yellow!20!red] (\posX+0.625, \posY+0.5) rectangle (\posX+0.725, \posY+0.75); 
\fill [yellow!20!red] (\posX+0.75, \posY+0.55) rectangle (\posX+0.85, \posY+0.7); 
\draw [black, fill=white] (\posX+0.135, \posY+0.475) rectangle (\posX+0.65, \posY+0.775);     
\draw (\posX+0.42, \posY+0.625) node[rectangle] {\scriptsize{\textbf{D}$_{a}$}};  

\draw [->, thick] (n1.east) |- (\posX,-1.5);
\draw [->, thick] (n2.east) |- (\posX,-1.75);

\def\posY{-1.5}
\def\posX{0.85}
\def\step{-0.35}
\draw [black, fill=white] (\posX+2, \posY+0.5) rectangle (\posX+2.35, \posY+0.85);     
\draw (\posX+2.175, \posY+0.675) node[rectangle] {\scriptsize{0}};  
\draw  [black] (\posX+2.65, \posY+0.675) node {\scriptsize{ID}};    
\draw [black, fill=white] (\posX+2, \posY+\step+0.5) rectangle (\posX+2.35, \posY+\step+0.85);     
\draw (\posX+2.175, \posY+\step+0.675) node[rectangle] {\scriptsize{1}};  
\draw  [black] (\posX+2.85, \posY+\step+0.675) node {\scriptsize{Gender}};    
\draw [black, fill=white] (\posX+2, \posY+2*\step+0.5) rectangle (\posX+2.35, \posY+2*\step+0.85);     
\draw (\posX+2.175, \posY+2*\step+0.675) node[rectangle] {\scriptsize{1}};  
\draw  [black] (\posX+2.95, \posY+2*\step+0.675) node {\scriptsize{Ethnicity}};    
\draw (\posX+2.175, \posY+3*\step+0.675) node[rectangle] {\scriptsize{$\vdots$}};  
\draw [black, fill=white] (\posX+2, \posY+4.45*\step+0.5) rectangle (\posX+2.35, \posY+4.45*\step+0.85);     
\draw (\posX+2.175, \posY+4.45*\step+0.675) node[rectangle] {\scriptsize{0}};  
\draw  [black] (\posX+2.95, \posY+4.45*\step+0.675) node {\scriptsize{Hairstyle}};    
\draw [white, fill=yellow!60] (4.75-0.25, -1.5-0.25) rectangle (4.75+0.25, -1.5+0.25);     
\node at (4.75, -1.5) [black, rotate=0]  {{$\mathcal{L}_{\text{ce}}$}};      

\draw [dashed, very thick] (0.25, -3) -- (5.5, -3);
\draw [dashed, very thick] (5.5, -3) -- (5.5, -0.25);

%%%%%%%%%%%%%%%%%%%%%%%%%%%%%%%%%%%%%%%%%%%%

\draw  [black] (11.75, -0.2) node {\small{\textbf{Learning Phase 2}}};    

\def\posX{6.55}
\def\posY{-2.25}
\draw (\posX+0.2,\posY+0.2) node(n1)  {\includegraphics[width=0.75 cm]{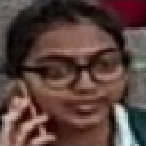}};   
\draw (\posX+0.1,\posY+0.1) node(n1)  {\includegraphics[width=0.75 cm]{imgs/illustration_x2}};   
\draw (\posX,\posY) node(n1)  {\includegraphics[width=0.75 cm]{imgs/illustration_x2}};   
\draw  [black] (\posX+0.2, \posY+0.2+0.65) node {\scriptsize{\textbf{x$_{i,j,k}$}}};    

\draw [->, thick] (7.25, -2.05) |- (8.65,-2.05);

\def\posX{8.75}
\def\posY{-2.625}
\fill [black!60!green] (0+\posX, 0+\posY) rectangle (0.1+\posX, 1.25+\posY);     
\fill [black!60!green] (0.125+\posX, 0+\posY) rectangle (0.225+\posX, 0.95+\posY); 
\fill [black!60!green] (0.25+\posX, 0+\posY) rectangle (0.35+\posX, 0.75+\posY); 
\fill [black!60!green] (0.375+\posX, 0+\posY) rectangle (0.475+\posX, 0.45+\posY); 
\fill [black!60!green] (0.5+\posX, 0+\posY) rectangle (0.6+\posX, 0.75+\posY); 
\fill [black!60!green] (0.625+\posX, 0+\posY) rectangle (0.725+\posX, 0.95+\posY); 
\fill [black!60!green] (0.750+\posX, 0+\posY) rectangle (0.850+\posX, 1.25+\posY);     
\draw [-] (0.11+\posX, 1.1+\posY) -- (0.74+\posX, 1.1+\posY);
\draw [-] (0.235+\posX, 0.85+\posY) -- (0.615+\posX, 0.85+\posY);
\draw [-] (0.36+\posX, 0.65+\posY) -- (0.49+\posX, 0.65+\posY);
\draw [black, fill=white] (0.125+\posX, 0.075+\posY) rectangle (0.725+\posX, 0.375+\posY);     
\draw (0.5+\posX, 0.225+\posY) node[rectangle] {\scriptsize{\textbf{U}$_{e}$}};  

\draw [->, thick] (9.7, -2.05) |- (11.3,-2.05);

\def\posX{11.75}
\def\posY{-2.25}
\draw (\posX+0.2,\posY+0.2) node(n1)  {\includegraphics[width=0.75 cm]{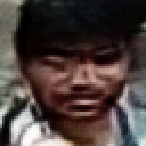}};   
\draw (\posX+0.1,\posY+0.1) node(n1)  {\includegraphics[width=0.75 cm]{imgs/illustration_a2}};   
\draw (\posX,\posY) node(n1)  {\includegraphics[width=0.75 cm]{imgs/illustration_a2}};   
\draw  [black] (\posX+0.2, \posY+0.2+0.65) node {\scriptsize{\textbf{a$_{i,j,k}$}}};    

\draw [->, thick] (12.45, -2.05) |- (13.65,-2.05);

\def\posX{13.75}
\def\posY{-2.625}
\fill [black!60!red] (0+\posX, 0+\posY) rectangle (0.1+\posX, 1.25+\posY);     
\fill [black!60!red] (0.125+\posX, 0+\posY) rectangle (0.225+\posX, 0.95+\posY); 
\fill [black!60!red] (0.25+\posX, 0+\posY) rectangle (0.35+\posX, 0.75+\posY); 
\fill [black!60!red] (0.375+\posX, 0+\posY) rectangle (0.475+\posX, 0.45+\posY); 
\fill [black!60!red] (0.5+\posX, 0+\posY) rectangle (0.6+\posX, 0.75+\posY); 
\fill [black!60!red] (0.625+\posX, 0+\posY) rectangle (0.725+\posX, 0.95+\posY); 
\fill [black!60!red] (0.750+\posX, 0+\posY) rectangle (0.850+\posX, 1.25+\posY);     
\draw [-] (0.11+\posX, 1.1+\posY) -- (0.74+\posX, 1.1+\posY);
\draw [-] (0.235+\posX, 0.85+\posY) -- (0.615+\posX, 0.85+\posY);
\draw [-] (0.36+\posX, 0.65+\posY) -- (0.49+\posX, 0.65+\posY);
\draw [black, fill=white] (0.125+\posX, 0.075+\posY) rectangle (0.725+\posX, 0.375+\posY);     
\draw (0.5+\posX, 0.225+\posY) node[rectangle] {\scriptsize{\textbf{U}$_{d}$}};  

\def\posX{16.75}
\def\posY{-2.25}
\draw (\posX+0.2,\posY+0.2) node(n1)  {\includegraphics[width=0.75 cm]{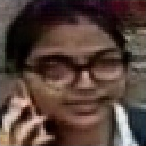}};   
\draw (\posX+0.1,\posY+0.1) node(n1)  {\includegraphics[width=0.75 cm]{imgs/illustration_r2}};   
\draw (\posX,\posY) node(n1)  {\includegraphics[width=0.75 cm]{imgs/illustration_r2}};   
\draw  [black] (\posX+0.2, \posY+0.2+0.65) node {\scriptsize{\textbf{r$_{i,j,k}$}}};    

\draw [white, fill=yellow!60] (13.75-0.5, -0.65-0.25) rectangle (13.75+0.5, -0.65+0.25);     
\node at (13.75, -0.65) [black, rotate=0]  {{$\mathcal{L}_{\text{mse}}$}};     
\draw [dotted, thick] (6.75, -1.25) |- (13.25, -0.65);
\draw [dotted, thick] (14.25, -0.65) -| (16.75, -1.25);

\draw [white, fill=yellow!60] (9.75-0.5, -1-0.25) rectangle (9.75+0.5, -1+0.25);     
\node at (9.75, -1.0) [black, rotate=0]  {{$\mathcal{L}_{\text{dis}}$}};     
\draw [dotted, thick] (6.75, -1.25) |- (9.25, -1);
\draw [dotted, thick] (10.25, -1) -| (11.75, -1.25);

\draw [->, thick] (14.75, -2.05) |- (16.25,-2.05);

%%%%%%%%%%%%%%%%%%%%%%%%%%%%%%%%%%%%%%%
% FACENESS

\def\posX{14.75}
\def\posY{-5.2}
\fill [yellow!60!red] (\posX, \posY) rectangle (\posX+0.1, \posY+1.25);     
\fill [yellow!60!red] (\posX+0.125, \posY+0.1) rectangle (\posX+0.225, \posY+1.15); 
\fill [yellow!60!red] (\posX+0.25, \posY+0.2) rectangle (\posX+0.35, \posY+1.05); 
\fill [yellow!60!red] (\posX+0.375, \posY+0.3) rectangle (\posX+0.475, \posY+0.95); 
\fill [yellow!60!red] (\posX+0.5, \posY+0.4) rectangle (\posX+0.6, \posY+0.85); 
\fill [yellow!60!red] (\posX+0.625, \posY+0.5) rectangle (\posX+0.725, \posY+0.75); 
\fill [yellow!60!red] (\posX+0.75, \posY+0.55) rectangle (\posX+0.85, \posY+0.7); 
\draw [black, fill=white] (\posX+0.135, \posY+0.475) rectangle (\posX+0.65, \posY+0.775);     
\draw (\posX+0.42, \posY+0.625) node[rectangle] {\scriptsize{\textbf{D}$_{f}$}};  

\draw (\posX+1.525, \posY+1.0) node[rectangle] {\scriptsize{1: \textbf{x}$_.$}};  
\draw (\posX+1.7, \posY+0.25) node[rectangle] {\scriptsize{-1: \textbf{a}$_.\vert$ \textbf{r}$_.$}};  

\draw [->] (\posX+0.9, \posY+0.6) -- (\posX+1.15,\posY+0.9);
\draw [->] (\posX+0.9, \posY+0.6) -- (\posX+1.15,\posY+0.35);

\draw [fill, black](\posX-0.5,\posY+0.6)circle(0.05);
\draw [->, thick] (\posX-0.5,\posY+0.6) -- (\posX-0.0,\posY+0.6);

\draw [white, fill=yellow!60] (17.25-0.5, -4.5-0.25) rectangle (17.25+0.5, -4.5+0.25);     
\node at (17.25, -4.5) [black, rotate=0]  {{$\mathcal{L}_{\text{adv}}$}};

%%%%%%%%%%%%%%%%%%%%%%%%%%%%%%%%%%%%%%%%%%%%%%%%
\def\posX{1.0}
\def\posY{-5.2}
\fill [yellow!20!red] (\posX, \posY) rectangle (\posX+0.1, \posY+1.25);     
\fill [yellow!20!red] (\posX+0.125, \posY+0.1) rectangle (\posX+0.225, \posY+1.15); 
\fill [yellow!20!red] (\posX+0.25, \posY+0.2) rectangle (\posX+0.35, \posY+1.05); 
\fill [yellow!20!red] (\posX+0.375, \posY+0.3) rectangle (\posX+0.475, \posY+0.95); 
\fill [yellow!20!red] (\posX+0.5, \posY+0.4) rectangle (\posX+0.6, \posY+0.85); 
\fill [yellow!20!red] (\posX+0.625, \posY+0.5) rectangle (\posX+0.725, \posY+0.75); 
\fill [yellow!20!red] (\posX+0.75, \posY+0.55) rectangle (\posX+0.85, \posY+0.7); 
\draw [black, fill=white] (\posX+0.135, \posY+0.475) rectangle (\posX+0.65, \posY+0.775);     
\draw (\posX+0.42, \posY+0.625) node[rectangle] {\scriptsize{\textbf{D}$_{a}$}};  
\draw [fill, black](\posX-0.5,\posY+0.25)circle(0.05);
\draw [fill, black](\posX-0.5,\posY+0.95)circle(0.05);
\draw [->, thick] (\posX-0.5,\posY+0.25) -- (\posX+0.0,\posY+0.25);
\draw  [black] (\posX-0.35, \posY+0.0) node {\scriptsize{\textbf{a$_{i,j,k}$}}};    
\draw [->, thick] (\posX-0.5,\posY+0.95) -- (\posX+0.0,\posY+0.95);
\draw  [black] (\posX-0.35, \posY+0.7) node {\scriptsize{\textbf{x$_{i,j,k}$}}};

\def\posY{-4.8}
\def\posX{0.3}
\def\step{-0.35}

\draw  [black] (\posX+2.175, \posY+1.0) node {\scriptsize{\textbf{s}}};    
\draw [black, fill=gray!60] (\posX+2, \posY+0.5) rectangle (\posX+2.35, \posY+0.85);     
\draw (\posX+2.175, \posY+0.675) node[rectangle] {\scriptsize{-1}};  
\draw [black, fill=gray!60] (\posX+2, \posY+\step+0.5) rectangle (\posX+2.35, \posY+\step+0.85);     
\draw (\posX+2.175, \posY+\step+0.675) node[rectangle] {\scriptsize{0}};  
\draw [black, fill=gray!60] (\posX+2, \posY+2*\step+0.5) rectangle (\posX+2.35, \posY+2*\step+0.85);     
\draw (\posX+2.175, \posY+2*\step+0.675) node[rectangle] {\scriptsize{1}};  
\draw (\posX+2.175, \posY+3*\step+0.675) node[rectangle] {\scriptsize{$\vdots$}};  
\draw [black, fill=gray!60] (\posX+2, \posY+4.45*\step+0.5) rectangle (\posX+2.35, \posY+4.45*\step+0.85);     
\draw (\posX+2.175, \posY+4.45*\step+0.675) node[rectangle] {\scriptsize{1}};

\draw [white, fill=yellow!60] (\posX+3.0-0.45, \posY-0.05) rectangle (\posX+3.0+0.45, \posY+0.45);     
\node at (\posX+3.0, \posY+0.2) [black, rotate=0]  {{$\mathcal{L}_{\text{ano}}$}};

\def\posX{-0.1}
\draw [black, fill=white] (\posX+2, \posY+0.5) rectangle (\posX+2.35, \posY+0.85);     
\draw (\posX+2.175, \posY+0.675) node[rectangle] {\scriptsize{0}};  
\draw [black, fill=white] (\posX+2, \posY+\step+0.5) rectangle (\posX+2.35, \posY+\step+0.85);     
\draw (\posX+2.175, \posY+\step+0.675) node[rectangle] {\scriptsize{1}};  
\draw [black, fill=white] (\posX+2, \posY+2*\step+0.5) rectangle (\posX+2.35, \posY+2*\step+0.85);     
\draw (\posX+2.175, \posY+2*\step+0.675) node[rectangle] {\scriptsize{0}};  
\draw (\posX+2.175, \posY+3*\step+0.675) node[rectangle] {\scriptsize{$\vdots$}};  
\draw [black, fill=white] (\posX+2, \posY+4.45*\step+0.5) rectangle (\posX+2.35, \posY+4.45*\step+0.85);     
\draw (\posX+2.175, \posY+4.45*\step+0.675) node[rectangle] {\scriptsize{1}};  

%%%%%%%%%%%%%%%%%%%%%%%%%%%%%%%%%%%%%%%%%%%%%%%%

%%%%%%%%%%%%%%%%%%%%%%%%%%%%%%%%%%%%%%%%%%%%%%%%
\def\posX{5.35}
\def\posY{-5.2}
\fill [yellow!20!red] (\posX, \posY) rectangle (\posX+0.1, \posY+1.25);     
\fill [yellow!20!red] (\posX+0.125, \posY+0.1) rectangle (\posX+0.225, \posY+1.15); 
\fill [yellow!20!red] (\posX+0.25, \posY+0.2) rectangle (\posX+0.35, \posY+1.05); 
\fill [yellow!20!red] (\posX+0.375, \posY+0.3) rectangle (\posX+0.475, \posY+0.95); 
\fill [yellow!20!red] (\posX+0.5, \posY+0.4) rectangle (\posX+0.6, \posY+0.85); 
\fill [yellow!20!red] (\posX+0.625, \posY+0.5) rectangle (\posX+0.725, \posY+0.75); 
\fill [yellow!20!red] (\posX+0.75, \posY+0.55) rectangle (\posX+0.85, \posY+0.7); 
\draw [black, fill=white] (\posX+0.135, \posY+0.475) rectangle (\posX+0.65, \posY+0.775);     
\draw (\posX+0.42, \posY+0.625) node[rectangle] {\scriptsize{\textbf{D}$_{a}$}};  
\draw [fill, black](\posX-0.5,\posY+0.25)circle(0.05);
\draw [fill, black](\posX-0.5,\posY+0.95)circle(0.05);
\draw [->, thick] (\posX-0.5,\posY+0.25) -- (\posX+0.0,\posY+0.25);
\draw  [black] (\posX-0.35, \posY+0.0) node {\scriptsize{\textbf{a$_{i,j,k'}$}}};    
\draw [->, thick] (\posX-0.5,\posY+0.95) -- (\posX+0.0,\posY+0.95);
\draw  [black] (\posX-0.35, \posY+0.7) node {\scriptsize{\textbf{a$_{i,j,k}$}}};    

\def\posY{-4.8}
\def\posX{4.5}
\def\step{-0.35}
%\draw [black, fill=gray!60] (\posX+2, \posY+0.5) rectangle (\posX+2.35, \posY+0.85);     
%\draw (\posX+2.175, \posY+0.675) node[rectangle] {\scriptsize{-1}};  
%\draw [black, fill=gray!60] (\posX+2, \posY+\step+0.5) rectangle (\posX+2.35, \posY+\step+0.85);     
%\draw (\posX+2.175, \posY+\step+0.675) node[rectangle] {\scriptsize{0}};  
%\draw [black, fill=gray!60] (\posX+2, \posY+2*\step+0.5) rectangle (\posX+2.35, \posY+2*\step+0.85);     
%\draw (\posX+2.175, \posY+2*\step+0.675) node[rectangle] {\scriptsize{1}};  
%\draw (\posX+2.175, \posY+3*\step+0.675) node[rectangle] {\scriptsize{$\vdots$}};  
%\draw [black, fill=gray!60] (\posX+2, \posY+4.45*\step+0.5) rectangle (\posX+2.35, \posY+4.45*\step+0.85);     
%\draw (\posX+2.175, \posY+4.45*\step+0.675) node[rectangle] {\scriptsize{1}};  
%\draw [white, fill=yellow!60] (\posX+3.15-0.45, \posY-0.25) rectangle (\posX+3.15+0.45, \posY+0.25);     
%\node at (\posX+3.15, \posY) [black, rotate=0]  {{$\mathcal{L}_{\text{con}}$}};     
%
%

%%%

\draw [black, fill=white] (\posX+2, \posY+0.5) rectangle (\posX+2.35, \posY+0.85);     
\draw (\posX+2.175, \posY+0.675) node[rectangle] {\scriptsize{0}};  
\draw [black, fill=white] (\posX+2, \posY+\step+0.5) rectangle (\posX+2.35, \posY+\step+0.85);     
\draw (\posX+2.175, \posY+\step+0.675) node[rectangle] {\scriptsize{0}};  
\draw [black, fill=white] (\posX+2, \posY+2*\step+0.5) rectangle (\posX+2.35, \posY+2*\step+0.85);     
\draw (\posX+2.175, \posY+2*\step+0.675) node[rectangle] {\scriptsize{1}};  
\draw (\posX+2.175, \posY+3*\step+0.675) node[rectangle] {\scriptsize{$\vdots$}};  
\draw [black, fill=white] (\posX+2, \posY+4.45*\step+0.5) rectangle (\posX+2.35, \posY+4.45*\step+0.85);     
\draw (\posX+2.175, \posY+4.45*\step+0.675) node[rectangle] {\scriptsize{1}};

\draw [white, fill=yellow!60] (\posX+3.0-0.45, \posY-0.05) rectangle (\posX+3.0+0.45, \posY+0.45);     
\node at (\posX+3.0, \posY+0.2) [black, rotate=0]  {{$\mathcal{L}_{\text{con}}$}};

%\def\posX{4.35}
%\draw [black, fill=white] (\posX+2, \posY+0.5) rectangle (\posX+2.35, \posY+0.85);     
%\draw (\posX+2.175, \posY+0.675) node[rectangle] {\scriptsize{1}};  
%\draw [black, fill=white] (\posX+2, \posY+\step+0.5) rectangle (\posX+2.35, \posY+\step+0.85);     
%\draw (\posX+2.175, \posY+\step+0.675) node[rectangle] {\scriptsize{-1}};  
%\draw [black, fill=white] (\posX+2, \posY+2*\step+0.5) rectangle (\posX+2.35, \posY+2*\step+0.85);     
%\draw (\posX+2.175, \posY+2*\step+0.675) node[rectangle] {\scriptsize{-1}};  
%\draw (\posX+2.175, \posY+3*\step+0.675) node[rectangle] {\scriptsize{$\vdots$}};  
%\draw [black, fill=white] (\posX+2, \posY+4.45*\step+0.5) rectangle (\posX+2.35, \posY+4.45*\step+0.85);     
%\draw (\posX+2.175, \posY+4.45*\step+0.675) node[rectangle] {\scriptsize{1}};  

%%%%%%%%%%%%%%%%%%%%%%%%%%%%%%%%%%%%%%%%%%%%%%%%

\def\posX{9.7}
\def\posY{-5.2}
\fill [yellow!20!red] (\posX, \posY) rectangle (\posX+0.1, \posY+1.25);     
\fill [yellow!20!red] (\posX+0.125, \posY+0.1) rectangle (\posX+0.225, \posY+1.15); 
\fill [yellow!20!red] (\posX+0.25, \posY+0.2) rectangle (\posX+0.35, \posY+1.05); 
\fill [yellow!20!red] (\posX+0.375, \posY+0.3) rectangle (\posX+0.475, \posY+0.95); 
\fill [yellow!20!red] (\posX+0.5, \posY+0.4) rectangle (\posX+0.6, \posY+0.85); 
\fill [yellow!20!red] (\posX+0.625, \posY+0.5) rectangle (\posX+0.725, \posY+0.75); 
\fill [yellow!20!red] (\posX+0.75, \posY+0.55) rectangle (\posX+0.85, \posY+0.7); 
\draw [black, fill=white] (\posX+0.135, \posY+0.475) rectangle (\posX+0.65, \posY+0.775);     
\draw (\posX+0.42, \posY+0.625) node[rectangle] {\scriptsize{\textbf{D}$_{a}$}};  
\draw [fill, black](\posX-0.5,\posY+0.25)circle(0.05);
\draw [fill, black](\posX-0.5,\posY+0.95)circle(0.05);
\draw [->, thick] (\posX-0.5,\posY+0.25) -- (\posX+0.0,\posY+0.25);
\draw  [black] (\posX-0.35, \posY+0.0) node {\scriptsize{\textbf{a$_{i',.,.}$}}};    
\draw [->, thick] (\posX-0.5,\posY+0.95) -- (\posX+0.0,\posY+0.95);
\draw  [black] (\posX-0.35, \posY+0.7) node {\scriptsize{\textbf{a$_{i,.,.}$}}};    

\def\posY{-4.8}
\def\posX{8.8}
\def\step{-0.35}
%\draw [black, fill=gray!60] (\posX+2, \posY+0.5) rectangle (\posX+2.35, \posY+0.85);     
%\draw (\posX+2.175, \posY+0.675) node[rectangle] {\scriptsize{-1}};  
%\draw [black, fill=gray!60] (\posX+2, \posY+\step+0.5) rectangle (\posX+2.35, \posY+\step+0.85);     
%\draw (\posX+2.175, \posY+\step+0.675) node[rectangle] {\scriptsize{0}};  
%\draw [black, fill=gray!60] (\posX+2, \posY+2*\step+0.5) rectangle (\posX+2.35, \posY+2*\step+0.85);     
%\draw (\posX+2.175, \posY+2*\step+0.675) node[rectangle] {\scriptsize{1}};  
%\draw (\posX+2.175, \posY+3*\step+0.675) node[rectangle] {\scriptsize{$\vdots$}};  
%\draw [black, fill=gray!60] (\posX+2, \posY+4.45*\step+0.5) rectangle (\posX+2.35, \posY+4.45*\step+0.85);     
%\draw (\posX+2.175, \posY+4.45*\step+0.675) node[rectangle] {\scriptsize{1}};  
%\draw [white, fill=yellow!60] (\posX+3.15-0.45, \posY-0.25) rectangle (\posX+3.15+0.45, \posY+0.25);     
%\node at (\posX+3.15, \posY) [black, rotate=0]  {{$\mathcal{L}_{\text{div}}$}};     

%%%

\draw [black, fill=white] (\posX+2, \posY+0.5) rectangle (\posX+2.35, \posY+0.85);     
\draw (\posX+2.175, \posY+0.675) node[rectangle] {\scriptsize{0}};  
\draw [black, fill=white] (\posX+2, \posY+\step+0.5) rectangle (\posX+2.35, \posY+\step+0.85);     
\draw (\posX+2.175, \posY+\step+0.675) node[rectangle] {\scriptsize{0}};  
\draw [black, fill=white] (\posX+2, \posY+2*\step+0.5) rectangle (\posX+2.35, \posY+2*\step+0.85);     
\draw (\posX+2.175, \posY+2*\step+0.675) node[rectangle] {\scriptsize{1}};  
\draw (\posX+2.175, \posY+3*\step+0.675) node[rectangle] {\scriptsize{$\vdots$}};  
\draw [black, fill=white] (\posX+2, \posY+4.45*\step+0.5) rectangle (\posX+2.35, \posY+4.45*\step+0.85);     
\draw (\posX+2.175, \posY+4.45*\step+0.675) node[rectangle] {\scriptsize{1}};

\draw [white, fill=yellow!60] (\posX+3.0-0.45, \posY-0.05) rectangle (\posX+3.0+0.45, \posY+0.45);     
\node at (\posX+3.0, \posY+0.2) [black, rotate=0]  {{$\mathcal{L}_{\text{div}}$}};

\draw  [black] (5.75, -6.35) node {\small{Shared Layers}};    
\draw [dotted, thick] (6.75, -6.35) -| (10.1, -5);
\draw [dotted, thick] (4.75, -6.35) -| (1.4, -5);
\draw [dotted, thick] (5.75, -6.15) -- (5.75, -5.0);

\draw [->, dashed, thick] (16.75,-2.9) -- (16.75, -3.75) -- (14.25, -3.75) -- (14.25,-4.5);
\draw [->, dashed, thick] (11.95,-2.9) -- (11.95, -3.25) -- (13.25, -3.25) -- (14.1,-4.6);
\draw [->, dashed, thick] (6.85,-2.9) -- (6.85, -3.5) -- (12.5, -3.5) -- (13.25, -4.75) -- (14.0,-4.75) -- (14.15, -4.65);

\draw [->, thick] (6.45,-2.9) -- (6.45, -3.6) -- (0.5, -3.6) -- (0.5,-4.15);
\draw [->, thick] (11.65,-2.9) -- (11.65, -3.2) -- (0.15, -3.2) -- (0.15, -4.95) -- (0.44,-4.95);

\draw [fill, black](8.5, -3.2 )circle(0.05);
\draw [->, thick] (8.5,-3.2) -- (8.5, -4.25) -- (9.1, -4.25);
\draw [->, thick] (8.5,-3.2) -- (8.5, -4.95) -- (9.1, -4.95);

\draw [fill, black](4, -3.2 )circle(0.05);
\draw [->, thick] (4,-3.2) -- (4, -4.25) -- (4.76, -4.25);
\draw [->, thick] (4,-3.2) -- (4, -4.95) -- (4.76, -4.95);

\draw [white, fill=yellow!60] (11.5, -6.0) rectangle (17.5, -6.5);     
\node at (14.5, -6.25) [black, rotate=0]  {{$\mathcal{L}= \sum_{i \in \{\text{mse, adv, ano, con, dis, div}\}} \omega_i~\mathcal{L}_{i}$}};     

%%%%%%%%%%%%%%%%%%%%%%%%%%%%%%%%%%%%%%%%%%%%%%%%

\draw [dashed, very thick] (0.25, -7) -- (17.75, -7);
\draw  [black] (4.5, -7.2) node {\small{\textbf{Inference}}};

\def\posX{0.65}
\def\posY{-9.0}
\draw (\posX+0.2,\posY+0.2) node(n1)  {\includegraphics[width=0.75 cm]{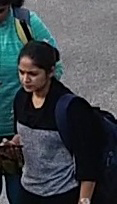}};   
\draw (\posX+0.1,\posY+0.1) node(n1)  {\includegraphics[width=0.75 cm]{imgs/full_body_x}};   
\draw (\posX,\posY) node(n1)  {\includegraphics[width=0.75 cm]{imgs/full_body_x}};   

\def\posX{2.5}
\def\posY{-9}
\draw [black, rounded corners, thick, fill=white] (\posX-0.8, \posY-0.5) rectangle (\posX+0.8, \posY+0.5);     
\draw  [black] (\posX, \posY+0.15) node {\scriptsize{Head}};    
\draw  [black] (\posX, \posY-0.15) node {\scriptsize{Detection~\cite{Ren2017}}};    

\draw [->, thick] (1.25, \posY) |- (1.6,\posY);

\def\posX{4.25}
\def\posY{-9}
\draw (\posX+0.2,\posY+0.2) node(n1)  {\includegraphics[width=0.75 cm]{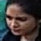}};   
\draw (\posX+0.1,\posY+0.1) node(n1)  {\includegraphics[width=0.75 cm]{imgs/x}};   
\draw (\posX,\posY) node(n1)  {\includegraphics[width=0.75 cm]{imgs/x}};   
\draw  [black] (\posX+0.2, \posY+0.2+0.65) node {\scriptsize{\textbf{x$_{i,j,k}$}}};   

\draw [->, thick] (3.35, \posY) |- (3.8,\posY);

\def\posX{5.35}
\def\posY{-9.5}
\fill [black!60!green] (0+\posX, 0+\posY) rectangle (0.1+\posX, 1.25+\posY);     
\fill [black!60!green] (0.125+\posX, 0+\posY) rectangle (0.225+\posX, 0.95+\posY); 
\fill [black!60!green] (0.25+\posX, 0+\posY) rectangle (0.35+\posX, 0.75+\posY); 
\fill [black!60!green] (0.375+\posX, 0+\posY) rectangle (0.475+\posX, 0.45+\posY); 
\fill [black!60!green] (0.5+\posX, 0+\posY) rectangle (0.6+\posX, 0.75+\posY); 
\fill [black!60!green] (0.625+\posX, 0+\posY) rectangle (0.725+\posX, 0.95+\posY); 
\fill [black!60!green] (0.750+\posX, 0+\posY) rectangle (0.850+\posX, 1.25+\posY);     
\draw [-] (0.11+\posX, 1.1+\posY) -- (0.74+\posX, 1.1+\posY);
\draw [-] (0.235+\posX, 0.85+\posY) -- (0.615+\posX, 0.85+\posY);
\draw [-] (0.36+\posX, 0.65+\posY) -- (0.49+\posX, 0.65+\posY);
\draw [black, fill=white] (0.125+\posX, 0.075+\posY) rectangle (0.725+\posX, 0.375+\posY);     
\draw (0.5+\posX, 0.225+\posY) node[rectangle] {\scriptsize{\textbf{U}$_{e}$}};

\def\posX{5.5}
\def\posY{-9}
\draw [black, rounded corners, thick, fill=white] (\posX+1.2, \posY-0.5) rectangle (\posX+2.8, \posY+0.5);     
\draw  [black] (\posX+2, \posY+0.15) node {\scriptsize{Steganography}};    
\draw  [black] (\posX+2, \posY-0.15) node {\scriptsize{(encoding)~\cite{Denemark2016} }};    

\draw [->, thick] (4.9, \posY) |- (5.3,\posY);

\def\posX{9.15}
\def\posY{-9}
\draw (\posX+0.2,\posY+0.2) node(n1)  {\includegraphics[width=0.75 cm]{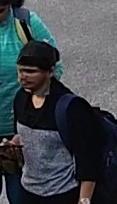}};   
\draw (\posX+0.1,\posY+0.1) node(n1)  {\includegraphics[width=0.75 cm]{imgs/full_body_a}};   
\draw (\posX,\posY) node(n1)  {\includegraphics[width=0.75 cm]{imgs/full_body_a}};   

\def\posX{9.0}
\def\posY{-9}
\draw [black, rounded corners, thick, fill=white] (\posX+1.2, \posY-0.5) rectangle (\posX+2.8, \posY+0.5);     
\draw  [black] (\posX+2, \posY+0.15) node {\scriptsize{Steganography}};    
\draw  [black] (\posX+2, \posY-0.15) node {\scriptsize{(decoding)}};    

\draw [->, thick] (6.22, \posY) |- (6.6,\posY);

\def\posX{12.65}
\def\posY{-9}
\draw (\posX+0.2,\posY+0.2) node(n1)  {\includegraphics[width=0.75 cm]{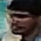}};   
\draw (\posX+0.1,\posY+0.1) node(n1)  {\includegraphics[width=0.75 cm]{imgs/a}};   
\draw (\posX,\posY) node(n1)  {\includegraphics[width=0.75 cm]{imgs/a}};   
\draw  [black] (\posX+0.2, \posY+0.2+0.65) node {\scriptsize{\textbf{a$_{i,j,k}$}}};   

\draw [->, thick] (8.38, \posY+0.0) -- (8.5, \posY+0.0) -- (8.5, \posY+0.5) -- (8.72,\posY+0.5);

\draw [->, thick] (1.35, -9) -- (1.35, \posY-0.9)  -- (8.5, \posY-0.9) -- (8.5, \posY-0.5) -- (8.72,\posY-0.5);

\def\posX{13.75}
\def\posY{-9.5}
\fill [black!60!red] (0+\posX, 0+\posY) rectangle (0.1+\posX, 1.25+\posY);     
\fill [black!60!red] (0.125+\posX, 0+\posY) rectangle (0.225+\posX, 0.95+\posY); 
\fill [black!60!red] (0.25+\posX, 0+\posY) rectangle (0.35+\posX, 0.75+\posY); 
\fill [black!60!red] (0.375+\posX, 0+\posY) rectangle (0.475+\posX, 0.45+\posY); 
\fill [black!60!red] (0.5+\posX, 0+\posY) rectangle (0.6+\posX, 0.75+\posY); 
\fill [black!60!red] (0.625+\posX, 0+\posY) rectangle (0.725+\posX, 0.95+\posY); 
\fill [black!60!red] (0.750+\posX, 0+\posY) rectangle (0.850+\posX, 1.25+\posY);     
\draw [-] (0.11+\posX, 1.1+\posY) -- (0.74+\posX, 1.1+\posY);
\draw [-] (0.235+\posX, 0.85+\posY) -- (0.615+\posX, 0.85+\posY);
\draw [-] (0.36+\posX, 0.65+\posY) -- (0.49+\posX, 0.65+\posY);
\draw [black, fill=white] (0.125+\posX, 0.075+\posY) rectangle (0.725+\posX, 0.375+\posY);     
\draw (0.5+\posX, 0.225+\posY) node[rectangle] {\scriptsize{\textbf{U}$_{d}$}};  

\def\posX{15.5}
\def\posY{-9}
\draw (\posX+0.2,\posY+0.2) node(n1)  {\includegraphics[width=0.75 cm]{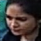}};   
\draw (\posX+0.1,\posY+0.1) node(n1)  {\includegraphics[width=0.75 cm]{imgs/r}};   
\draw (\posX,\posY) node(n1)  {\includegraphics[width=0.75 cm]{imgs/r}};   
\draw  [black] (\posX+0.2, \posY+0.2+0.65) node {\scriptsize{\textbf{r$_{i,j,k}$}}};   

\draw [->, thick] (9.75, \posY) -- (10.15,\posY);

\draw [->, thick] (11.86, \posY) -- (12.25,\posY);

\draw [->, thick] (13.3, \posY) -- (13.7,\posY);

\draw [->, thick] (14.65, \posY) -- (15.08,\posY);

\draw [->, thick] (16.15, \posY) -- (16.4, \posY) -- (16.4, \posY+0.5) -- (16.575,\posY + 0.5);

\draw [->, thick] (9.85, -9) -- (9.85, \posY-0.9)  -- (16.4, \posY-0.9) -- (16.4, \posY-0.5) -- (16.575,\posY-0.5);

\def\posX{17.00}
\def\posY{-9}
\draw (\posX+0.2,\posY+0.2) node(n1)  {\includegraphics[width=0.75 cm]{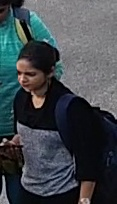}};   
\draw (\posX+0.1,\posY+0.1) node(n1)  {\includegraphics[width=0.75 cm]{imgs/full_body_r}};   
\draw (\posX,\posY) node(n1)  {\includegraphics[width=0.75 cm]{imgs/full_body_r}};

\fill [green!20, rounded corners, thick, opacity=0.3] (0.15, -7.6) rectangle (9.85, -10.05);    

\draw  [black] (0.9, -7.4) node {\small{{Raw Stream}}};     

\draw  [black] (9.25, -7.4) node {\small{{Published Stream}}};     

\draw  [black] (17.0, -7.4) node {\small{{Private Stream}}};     

\fill [red!20, rounded corners, thick, opacity=0.3] (10.05, -7.6) rectangle (17.85, -10.05);     

\end{tikzpicture}
    \caption{Cohesive perspective of the proposed method. During phase 1, we learn a pairwise attributes matcher \textbf{D}$_a$ that determines the shared labels between image pairs. Next, a double-U sequential architecture is proposed, where the first part \textbf{U}$_e$ receives the raw data \textbf{x}and produces the de-identified versions \textbf{a}, complemented by the reverser \textbf{U}$_d$ that exclusively analyzes the de-identified data and reconstructs the original samples \textbf{x}. The pairwise matcher is used as basis of the anonymization, temporal consistency and diversity losses, along with an adversarial discriminator \textbf{D}$_a$ that enforces the \emph{facial appearance} of the generated data. In inference time, a state-of-the-art head detector is used to crop the regions that fed the \textbf{U}$_e$ network. Next, using steganography techniques, the anonymized regions-of-interest are hidden in the published stream, released with minimal privacy concerns. Finally, such public streams are used by  the reverser network \textbf{U}$_d$, available to legal authorities, that is able to reconstruct the original scenes and disclosure the actual identities in the scene.}
        \label{fig:proposedMethod}
    \end{center}
\end{figure*}

A global perspective of our method is shown in Fig.~\ref{fig:proposedMethod}. The learning process is divided into two phases: 1) we start by obtaining an auxiliary pairwise attribute matcher \textbf{D}$_{a}$ that predicts the common labels (ID, gender, ethnicity, age and hairstyle) between image pairs; and 2) use the \textbf{D}$_{a}$  responses to constraint the properties of the de-identified samples in the adversarial learning phase, along with an adversarial discriminator \textbf{D}$_{f}$ that distinguishes between the input images \textbf{x} and the outputs of the generator (\textbf{U}$_{e}$ and \textbf{U}$_{d}$). 

\subsection{Learning Phase 1: \emph{'Same'}/\emph{'Different'} Pairwise Attributes Classifier}

Let \textbf{x}$^{\textbf{l}}_{i, j, k}$ represent the k$^{th}$ frame from the j$^{th}$ sequence of the i$^{th}$ subject in the learning set. Also, let \textbf{a}$^{\textbf{l}}_{i, j, k}$ represent the corresponding de-identified data and \textbf{r}$^{\textbf{l}}_{i, j, k}$ the reconstructed version. $\textbf{l} \in \mathbb{N}^t$ denotes the ground-truth attributes from \textbf{x}. We consider $t=4$, for \{\emph{ID}, \emph{gender}, \emph{ethnicity}, \emph{hairstyle}\} labels. For every pair of images \textbf{x$^{\textbf{l}}$}/\textbf{x}$'^{\textbf{l}'}$, we define a binary vector $\textbf{b}$ zeroed in the positions of the disagreeing labels between \textbf{l} and \textbf{l}':

\begin{equation}
\textbf{b} = \big[\mathbbm{1}_{\{l^1 == l'^1\}}, \ldots, \mathbbm{1}_{\{l^t == l'^t\}} \big],
\end{equation}
where ''$==$'' stands for the equality test operator and $\mathbbm{1}$ is the characteristic function. The attribute classifier \textbf{D}$_{a}$: $\mathbb{N}^n  \times  \mathbb{N}^n \rightarrow \mathbb{N}^t$ receives a pair of images (each of length n) and predicts their common labels:

\begin{equation}
\textbf{\^{b}} = \textbf{D}_a(\textbf{x}^{\textbf{l}}, \textbf{x}'^{\textbf{l}'}).
\end{equation}

We use a cross-entropy loss for the pairwise attribute classifier \textbf{D}$_{a}$, which is trained using the ground-truth \textbf{b} and predicted \textbf{\^{b}} attributes, according to:

\begin{equation}
\mathcal{L}_{\text{ce}} = \mathbb{E}_{\textbf{b}, \textbf{\^{b}}}~-\textbf{b}^T\log(\textbf{\^{b}}) - ({\overrightarrow{\textbf{1}}}^T-\textbf{b}^T)\log(1-\textbf{\^{b}}),
\end{equation}
with ${\overrightarrow{\textbf{1}}}$ representing an all-ones vector of $t$ components, and the logarithmic operation being applied component-wise to every element of its input. The inferred classifier is given by:

\begin{equation}
 \textbf{D}^*_{a} = \arg \min_{\textbf{D}_{a}} \mathcal{L}_{\text{ce}}.
\end{equation}

\subsection{Learning Phase 2: Reversible De-Identification}

\begin{figure}[ht!]
\begin{center}
\begin{tikzpicture}

\draw [black, rounded corners, thick] (0, 0) rectangle (8.5, -2.0);     

\def\posY{-1.2}
\def\posX{0.05}
\def\step{-0.35}
\draw [black] (\posX+2, \posY+0.5) rectangle (\posX+2.35, \posY+0.85);     
\draw (\posX+2.175, \posY+0.675) node[rectangle] {\scriptsize{0}};  
\draw  [black] (\posX+2.175, \posY+0.975) node {\tiny{{\textbf{Hs}}}};     

\draw [black] (\posX+2+\step, \posY+0.5) rectangle (\posX+\step+2.35, \posY+0.85);     
\draw (\posX+2.175+\step, \posY+0.675) node[rectangle] {\scriptsize{0}};  
\draw  [black] (\posX+2.175+\step, \posY+0.975) node {\tiny{{\textbf{Age}}}};     

\draw [black] (\posX+2+2*\step, \posY+0.5) rectangle (\posX+2*\step+2.35, \posY+0.85);     
\draw (\posX+2.175+2*\step, \posY+0.675) node[rectangle] {\scriptsize{0}};  
\draw  [black] (\posX+2.175+2*\step, \posY+0.975) node {\tiny{{\textbf{Eth}}}};     

\draw [black] (\posX+2+3*\step, \posY+0.5) rectangle (\posX+3*\step+2.35, \posY+0.85);     
\draw (\posX+2.175+3*\step, \posY+0.675) node[rectangle] {\scriptsize{1}};  
\draw  [black] (\posX+2.175+3*\step, \posY+0.975) node {\tiny{{\textbf{Ge}}}};     

\draw [black] (\posX+2+4*\step, \posY+0.5) rectangle (\posX+2*\step+2.35, \posY+0.85);     
\draw (\posX+2.175+4*\step, \posY+0.675) node[rectangle] {\scriptsize{-1}};  
\draw  [black] (\posX+2.175+4*\step, \posY+0.975) node {\tiny{{\textbf{ID}}}};   
\draw  [black] (\posX+1.8+4*\step, \posY+0.675) node {\scriptsize{{\textbf{s}:}}};       

\draw  [black] (3.5, -0.5) node {\Large{$\rightarrow$}};     

\draw  [black] (6.25, -0.35) node {\scriptsize{''\emph{De-identify the data, keeping}}};     
\draw  [black] (6.25, -0.65) node {\scriptsize{\emph{the 'gender' of the original samples}''}};    

\def\posY{-2.2}

\draw [black] (\posX+2, \posY+0.5) rectangle (\posX+2.35, \posY+0.85);     
\draw (\posX+2.175, \posY+0.675) node[rectangle] {\scriptsize{-1}};  
\draw  [black] (\posX+2.175, \posY+0.975) node {\tiny{{\textbf{Hs}}}};     

\draw [black] (\posX+2+\step, \posY+0.5) rectangle (\posX+\step+2.35, \posY+0.85);     
\draw (\posX+2.175+\step, \posY+0.675) node[rectangle] {\scriptsize{0}};  
\draw  [black] (\posX+2.175+\step, \posY+0.975) node {\tiny{{\textbf{Age}}}};     

\draw [black] (\posX+2+2*\step, \posY+0.5) rectangle (\posX+2*\step+2.35, \posY+0.85);     
\draw (\posX+2.175+2*\step, \posY+0.675) node[rectangle] {\scriptsize{0}};  
\draw  [black] (\posX+2.175+2*\step, \posY+0.975) node {\tiny{{\textbf{Eth}}}};     

\draw [black] (\posX+2+3*\step, \posY+0.5) rectangle (\posX+3*\step+2.35, \posY+0.85);     
\draw (\posX+2.175+3*\step, \posY+0.675) node[rectangle] {\scriptsize{0}};  
\draw  [black] (\posX+2.175+3*\step, \posY+0.975) node {\tiny{{\textbf{Ge}}}};     

\draw [black] (\posX+2+4*\step, \posY+0.5) rectangle (\posX+2*\step+2.35, \posY+0.85);     
\draw (\posX+2.175+4*\step, \posY+0.675) node[rectangle] {\scriptsize{-1}};  
\draw  [black] (\posX+2.175+4*\step, \posY+0.975) node {\tiny{{\textbf{ID}}}};     
\draw  [black] (\posX+1.8+4*\step, \posY+0.675) node {\scriptsize{{\textbf{s}:}}};

\draw  [black] (3.5, -0.5-1.0) node {\Large{$\rightarrow$}};     

\draw  [black] (6.25, -0.35-1.0) node {\scriptsize{''\emph{De-identify the data, changing}}};     
\draw  [black] (6.25, -0.65-1.0) node {\scriptsize{\emph{the 'hairstyle' of the original samples}''}};    

\end{tikzpicture}
    \caption{Illustration of the effect that different \textbf{s} parameterizations have in the de-identified data. We keep control of the soft labels in the de-identified faces, imposing agreement/disagreement or even make values independent with respect to the original data. }
        \label{fig:sign}
    \end{center}
\end{figure}
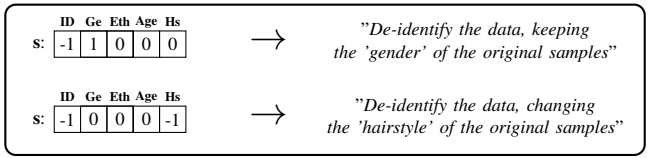

The next step is the main learning phase, where the de-identifier and reconstructor networks are inferred. We start by defining a vector $\textbf{s} \in \{-1, 0, 1\}^t$ where the positive coefficients denote labels that should be consistent between the image pairs, the negative coefficients represent labels that should disagree and the zeros determine independence between labels (Fig.~\ref{fig:sign}). $\textbf{s}$ weights the responses provided \textbf{D}$_a$ and serves to broad the range of applications for the proposed method, depending of the desired properties for the de-identified data. In every case, the ID position of \textbf{s} is always set to -1, guaranteeing the de-identification feature.
  
The first term is the \textbf{reconstruction loss} $\mathcal{L}_{\text{mse}}$, that guarantees the  fidelity of the samples reconstructed by $\textbf{U}_d$ with respect to $\textbf{x}$. To this end, not only the reconstructor $\textbf{U}_d$ should perfectly reconstruct  $\textbf{x}$, but the de-identifier $\textbf{U}_e$  should also encapsulate hidden features in $\textbf{a}_.$ that enable such reconstruction. It is given by:

\begin{equation}
\mathcal{L}_{\text{mse}} = || \textbf{x}_. - \textbf{U}_d \big( \textbf{U}_e(\textbf{x}_.)\big) ||_2.
\end{equation}

The \textbf{adversarial loss} is introduced by a \emph{face plausibility} discriminator $\textbf{D}_f$, that is responsible to distinguish between the input samples \textbf{x} and their encoded counterparts, either the de-identified $\textbf{a}_. =  \textbf{U}_e(\textbf{x}_.)$ or the reconstructed samples  $\textbf{r}_. =  \textbf{U}_d(\textbf{a}_.) =    \textbf{U}_d\Big(\textbf{U}_e(\textbf{x}_.)\Big)$ In practice, this component forces the encoded data to have \emph{facial appearance}, as both parts of generator \textbf{U}$_e$ and \textbf{U}$_d$ will attempt to fool $\textbf{D}_f$ during the adversarial learning process. From the discriminator perspective, this loss is formulated as: 

\begin{multline}
\mathcal{L}_{\text{adv$_1$}} = - 2.\mathbb{E}_{\textbf{x}}~\textbf{D}_f(\textbf{x}_.) + \mathbb{E}_{\textbf{a}_.}~\textbf{D}_f(\textbf{a}_.) + \mathbb{E}_{\textbf{r}_.}~\textbf{D}_f(\textbf{r}_.),  \\ \text{s.t.}~|| \textbf{D}_f ||_\infty \leq \delta_{\text{gp}}, 
\end{multline}
where $||.||_{\infty}$ denotes the maximum gradient allowed $\delta_{\text{gp}}$ to avoid mode collapse and enhance training stability, according to the WGAN-GP scheme~\cite{Gulrajani2017}. The optimal discriminator is formulated as:

\begin{equation}
 \textbf{D}^*_{f} = \arg \min_{\textbf{D}_{f}} \mathcal{L}_{\text{adv$_1$}}.
\end{equation}

From the encoder perspective, the  corresponding terms in the loss formulation have an opposite sign:
 
\begin{equation}
\mathcal{L}_{\text{adv$_2$}} = - \mathbb{E}_{\textbf{a}}~\textbf{D}_f(\textbf{a}_.) - \mathbb{E}_{\textbf{r}}~\textbf{D}_f(\textbf{r}_.).
\end{equation}

All the remaining terms evolved in the generator use the previously learned pairwise discriminator $\textbf{D}_a(.,.)$. The \textbf{anonymization loss} forces that pairs of \textbf{x}$_.$/\textbf{a}$_.$ elements follow the attribute configuration determined by $\textbf{s}$:

\begin{equation}
\mathcal{L}_{\text{ano}} = \mathbb{E}_{\textbf{x}_{i,j,k}, \textbf{a}_{i,j,k}} \textbf{s} \odot \Big(2. \textbf{D}_a(\textbf{x}_{i,j,k}, \textbf{a}_{i,j,k}) -1\Big), 
\end{equation} 
where $\odot$ denotes the inner product and the $\big(2.\textbf{D}_a(., .) -1\big)$ term maps the output of \textbf{D}$_a$ into the [-1, 1] scale. Similarly, the \textbf{temporal consistency} loss guarantees that any samples of the same subject sequence $\textbf{a}_{i,j,k}/\textbf{a}_{i,j,k'}$ share all the attributes, for photo realism purposes:

 \begin{equation}
\mathcal{L}_{\text{con}} =  - \mathbb{E}_{\textbf{a}_{i,j,k}, \textbf{a}_{i,j,k'}} \overrightarrow{\textbf{1}} \odot \textbf{D}_a(\textbf{a}_{i,j,k}, \textbf{a}_{i,j,k'}), 
\end{equation} 
with $\overrightarrow{\textbf{1}} \in \mathbb{N}^t$ denoting an all-ones row vector of $t$ components. 

The \textbf{diversity loss} assures that different sequences of the same subject $\textbf{a}_{i,j,.}/\textbf{a}_{i,j',.}$ are mapped to different virtual identities, to avoid any malicious inference of subjects and scenes properties:
      
 \begin{equation}
\mathcal{L}_{\text{div}} = \mathbb{E}_{\textbf{x}_{i,j,.}, \textbf{a}_{i,j,.}} \overrightarrow{\textbf{1}} \odot \textbf{D}_a(\textbf{a}_{i,j,.}, \textbf{a}_{i,j',.}).
\end{equation}
 
Finally, the \textbf{distribution loss} assures that the color distributions of \textbf{a} elements and the corresponding original samples \textbf{x} are similar, which augments the photo realism of the results and the resemblance of the original lighting conditions:

 \begin{equation}
\mathcal{L}_{\text{dis}} = \mathbb{E}_{\textbf{x}_{i,j,k}, \textbf{a}_{i,j,k}}  \chi_{h(\textbf{x}), h(\textbf{a})}^2, 
\end{equation} 
where $h(.)$ represents  the histogram operator and $\chi_{\textbf{v}^{(1)},\textbf{v}^{(2)}}^2$ denotes the Chi-square distance between the distributions of $(\textbf{v}^{(1)}, \textbf{v}^{(2)})$ given by $\sum_i \frac{(\textbf{v}^{(1)}_i-\textbf{v}^{(2)}_i)^2}{\textbf{v}^{(1)}_i+\textbf{v}^{(2)}_i}$, with $\textbf{v}^{(.)}_i$ is the i$^{th}$ bin density.

Overall, the full loss function is  the weighted sum of the above described terms:

\begin{multline}
 \textbf{U}^*_{e},  \textbf{U}^*_{d} = \arg \min_{\textbf{U}_{e}, \textbf{U}_{d}} \omega_{\text{mse}} \mathcal{L}_{\text{mse}} + \omega_{\text{adv}} \mathcal{L}_{\text{adv$_2$}} + \\ \omega_{\text{ano}} \mathcal{L}_{\text{ano}} + \omega_{\text{con}} \mathcal{L}_{\text{con}}  + \omega_{\text{div}} \mathcal{L}_{\text{div}} + \omega_{\text{dis}} \mathcal{L}_{\text{dis}},
\end{multline}
where $\omega_.$ are the hyper-parameters that weight the importance of each term in the learning process (details about the values used are given in Section~\ref{sec:Results}).

\subsection{Image Steganography}

Steganography is used in this work for two purposes: 1) to avoid that the head detector is used both in the de-identification and reconstruction phases; and, most importantly, 2) to assure that the region-of-interest (ROI) to reconstruct each \textbf{a}$_.$ is the same used to previously crop \textbf{x}$_.$. This is a sensitive point, as we observed a decrease in the fidelity of the reconstructed data, in case of misalignments between the ROIs cropped for corresponding \textbf{x}$_.$/\textbf{a}$_.$ elements. Hence, we encapsulate the output of the head detector~\cite{Ren2017} in the public stream, using the following protocol:

\begin{align}
  \text{message} & := \text{n + '',' + } [ \text{bound. box} ]_n  \nonumber \\   
  \text{bound. box}  & := \text{x + '','' + y + '','' + w + '','' + h + '',''} \nonumber \\ 
  \text{n} & := \{\mathbb{N}\} \nonumber \\  
  \text{x} & := \{\mathbb{N}\} \nonumber \\
  \text{y} & := \{\mathbb{N}\} \nonumber \\
   \text{w} & := \{\mathbb{N}\} \nonumber \\
  \text{h} & := \{\mathbb{N}\} \nonumber, 
\end{align}
where $n$ denotes the number of bounding boxes (faces) in the frame, $[.]_n$ denotes $n$ occurrences of one element, $\{\mathbb{N}\}$ stands for ''\emph{one natural number}'' and $(x, y, w, h)$ provide the top left corner $(x,y)$, plus the width $w$ and height $h$ of the box. 

This way, every frame in the public stream encapsulates (using~\cite{Denemark2016}) the number and position of all head regions (ROIs) in the frame. As an example, the message ''\emph{2,10,16,9,15,25,45,8,14,}" corresponds to two ROIs in a frame, one starting at position $(10,16)$, with dimensions $(9,15)$ and  another starting at position $(25,45)$ with dimensions $(8,14)$.

\subsection{Inference}

For security purposes, the data are de-identified before being published or transmitted through the networks (bottom row of Fig.~\ref{fig:proposedMethod}). This can be done \emph{in situ}, embedded in the camera hardware and starts by the detection~\cite{Ren2017} of the human heads in each frame (using~\cite{He2017} or~\cite{Redmon2018}). Next, the facial images \textbf{x}$_.$ feed the \textbf{U}$_e$ generator, that returns the de-identified versions \textbf{a}$_.$. Such virtual IDs are then overlapped in the each frame of the video sequence, with image steganography~\cite{Denemark2016} used to hidden information about all the ROIs.

The second part of the generator \textbf{U}$_d$ is supposed to be private and exclusively available to authorities. Upon a security incident,  the de-identified stream contains all the information required to reconstruct the original scene and disclosure the actual ID of the subjects in the scene. Using~\cite{Denemark2016}, we get the positions of the bounding boxes of the anonymised faces in every frame and feed \textbf{U}$_d$, that is responsible to reconstruct \textbf{x}. 

\section{Experiments and Results}
\label{sec:Results}

\subsection{Datasets and Empirical Protocol}
\label{ssec:datasets}

Our experiments were conducted in one proprietary (BIODI) and two freely available visual surveillance datasets (MARS and P-DESTRE). 

The BIODI\footnote{\url{http://di.ubi.pt/~hugomcp/BIODI/}} dataset is proprietary of \emph{Tomiworld}$^{\scriptsize{\textregistered}}$\footnote{\url{https://tomiworld.com/}}, and is composed of 849,932 images from 13,876 sequences, taken from 216 indoor/outdoor video surveillance sequences.  All images are manually annotated for 14 labels: \{'gender', 'age', 'height', 'body volume', 'ethnicity', 'hair color' and 'hairstyle', 'beard', 'moustache', 'glasses' and 'clothing' (x4)\}. As this set is not annotated for ID,  the face recognition experiments were exclusively performed in the remaining sets. MARS~\cite{Zheng2016} contains 1,261 IDs from around 20,000 tracklets, automatically extracted by the Deformable Part Model~\cite{Felzenszwalb2010} detector and the GMMCP~\cite{Dehghan2015} tracker. In this set, the soft labels \{'gender', 'ethnicity''\} were automatically inferred by the Matlab SDK  for \emph{Face++}\footnote{\url{http://www.faceplusplus.com/}} system, and the \emph{ethnicity} was manually annotated. Finally, the P-DESTRE~\cite{Kumar2020} dataset  provides video sequences of pedestrians in outdoor environments (taken from UAVs), and is fully annotated at the frame level, for ID and 16 soft labels: {'gender', 'age', 'height', 'body volume', 'ethnicity', 'hair colour', 'hairstyle', 'beard', 'moustache', 'glasses', 'head accessories', 'body accessories', 'action' and 'clothing information' (x3)}. It contains 253 identities and over 14.8M bounding boxes.

\begin{figure}[ht!]
\begin{center}
\begin{tikzpicture}

\def\sizeImg{1.35}
\def\deltaY{-2.4}

\draw (0*\sizeImg,0) node(segment_ok)  {\includegraphics[width=\sizeImg cm]{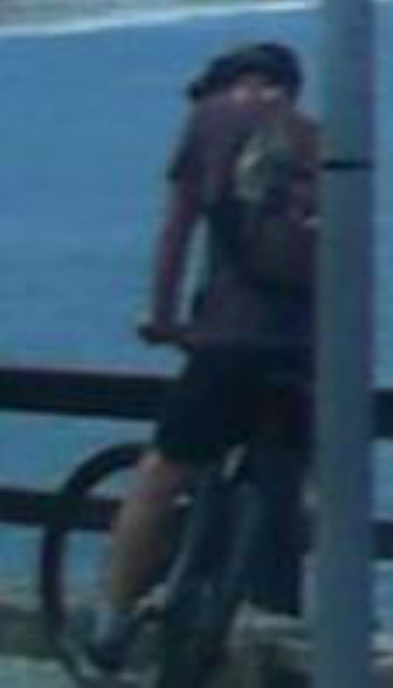}};  	
\draw (1*\sizeImg,0) node(segment_ok)  {\includegraphics[width=\sizeImg cm]{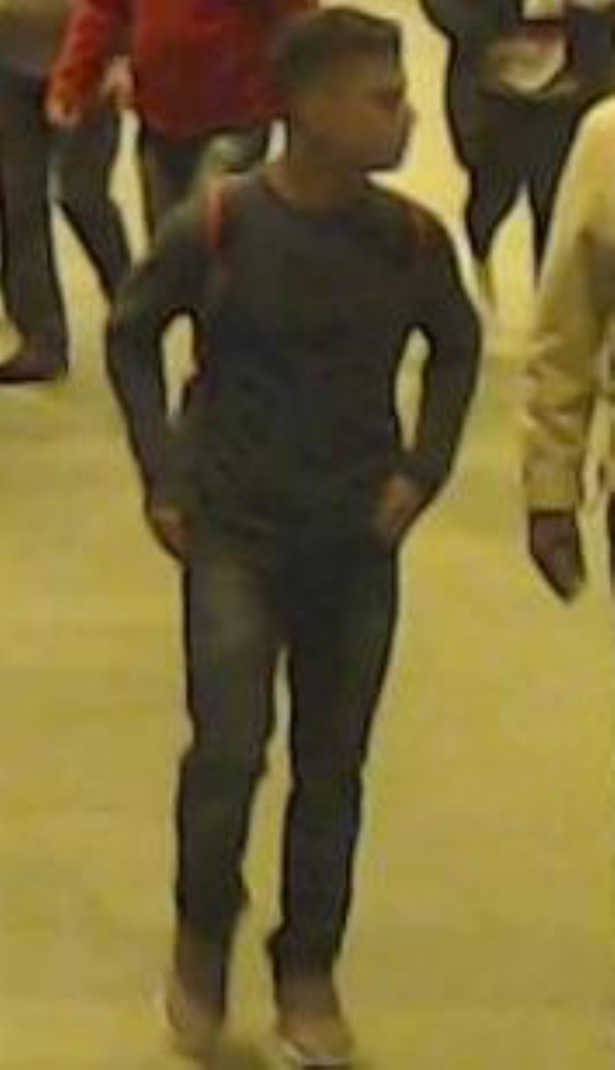}};  	
\draw (2*\sizeImg,0) node(segment_ok)  {\includegraphics[width=\sizeImg cm]{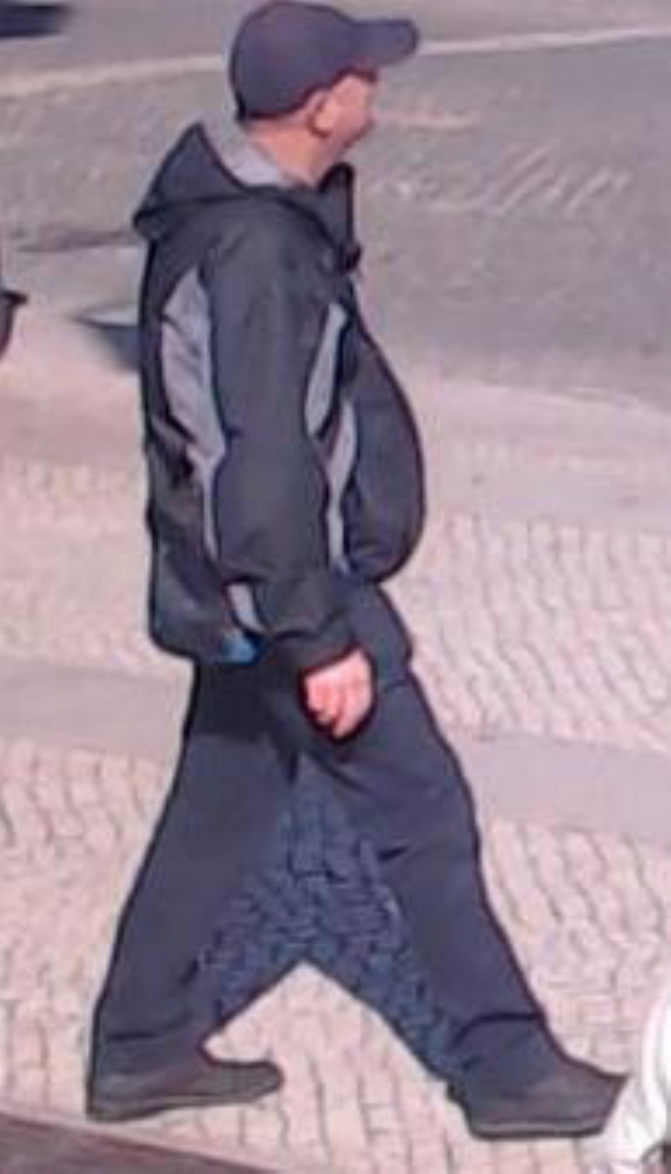}};  	
\draw (3*\sizeImg,0) node(segment_ok)  {\includegraphics[width=\sizeImg cm]{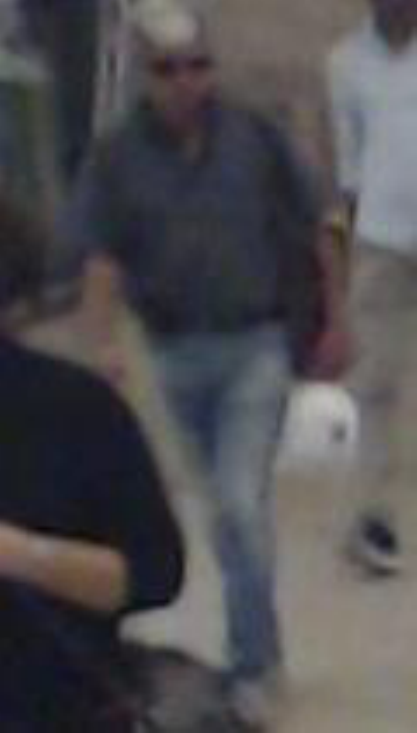}};  	
\draw (4*\sizeImg,0) node(segment_ok)  {\includegraphics[width=\sizeImg cm]{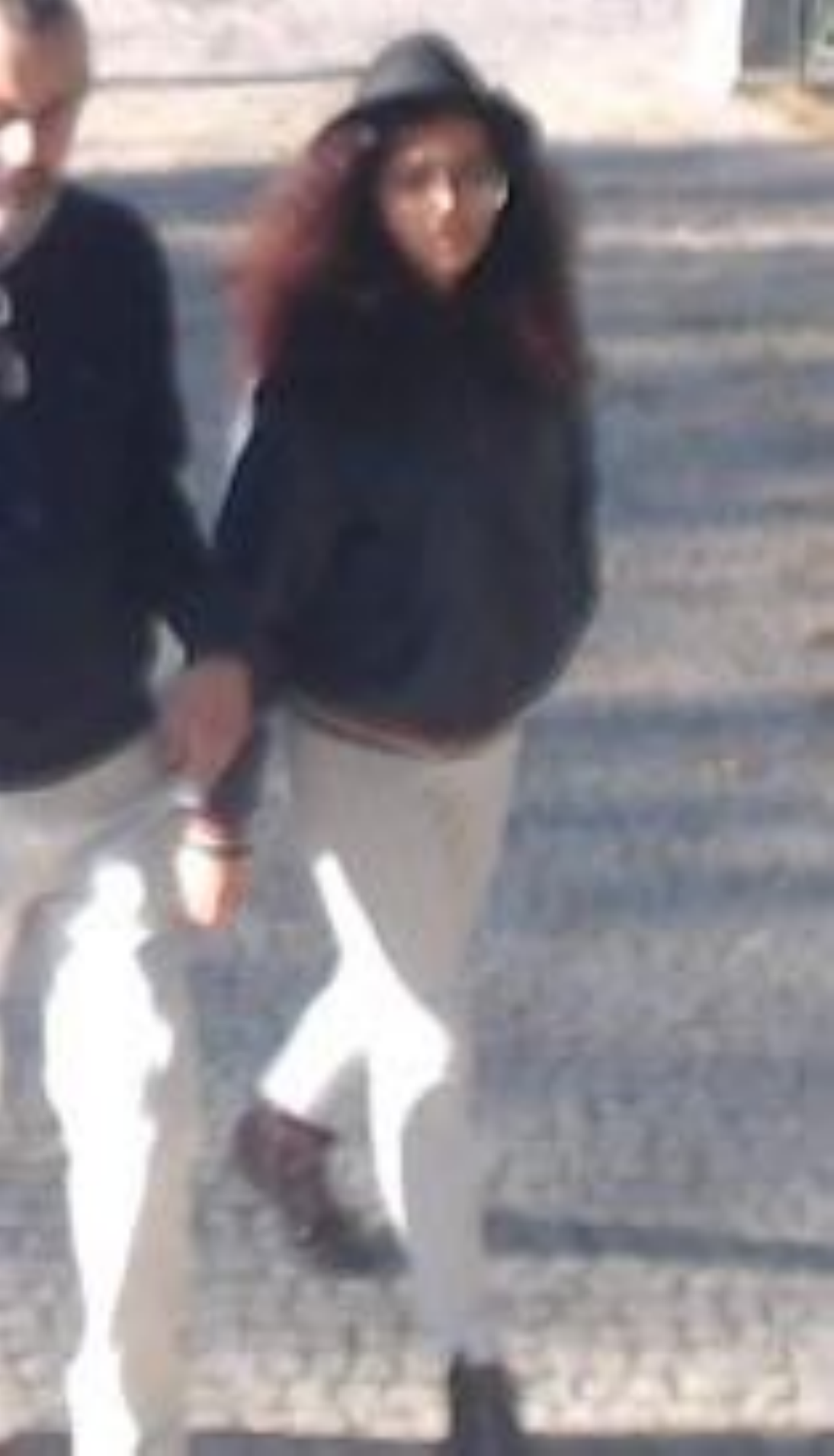}};  	
\draw (5*\sizeImg,0) node(segment_ok)  {\includegraphics[width=\sizeImg cm]{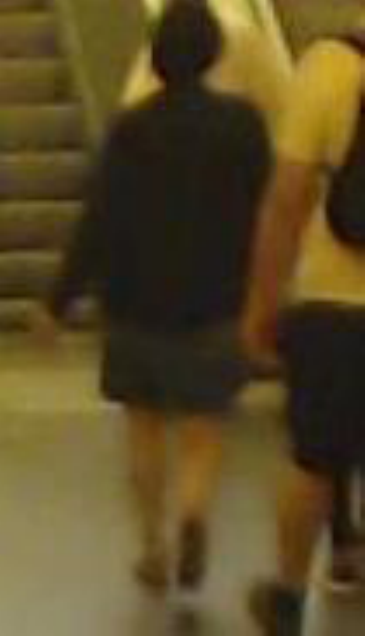}};  	
\draw (-1.0, 0) node[rectangle, rotate=90] {\small{\textbf{BIODI}}};  

\draw (0*\sizeImg,0+1.22*\deltaY) node(segment_ok)  {\includegraphics[width=\sizeImg cm]{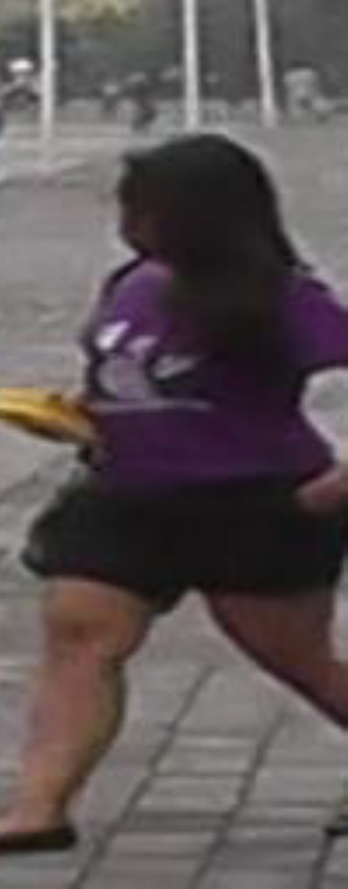}};  	
\draw (1*\sizeImg,0+1.22*\deltaY) node(segment_ok)  {\includegraphics[width=\sizeImg cm]{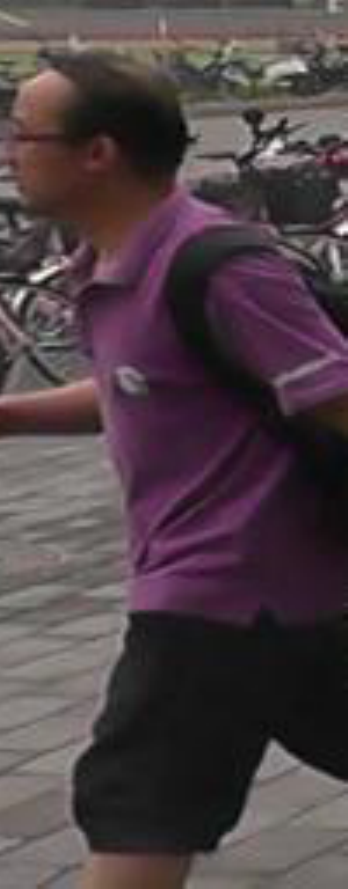}};  	
\draw (2*\sizeImg,0+1.22*\deltaY) node(segment_ok)  {\includegraphics[width=\sizeImg cm]{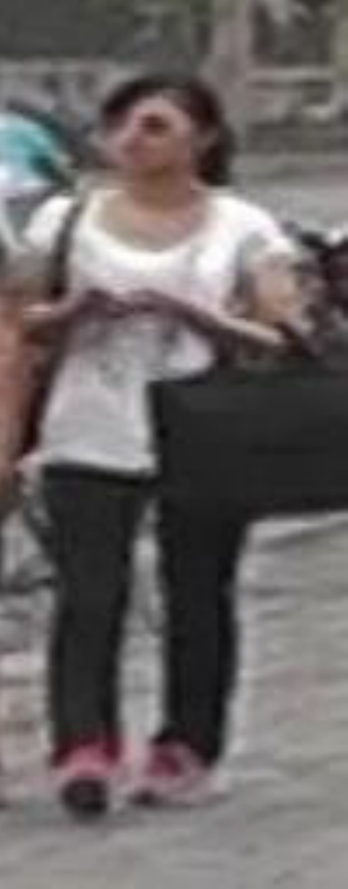}};  	
\draw (3*\sizeImg,0+1.22*\deltaY) node(segment_ok)  {\includegraphics[width=\sizeImg cm]{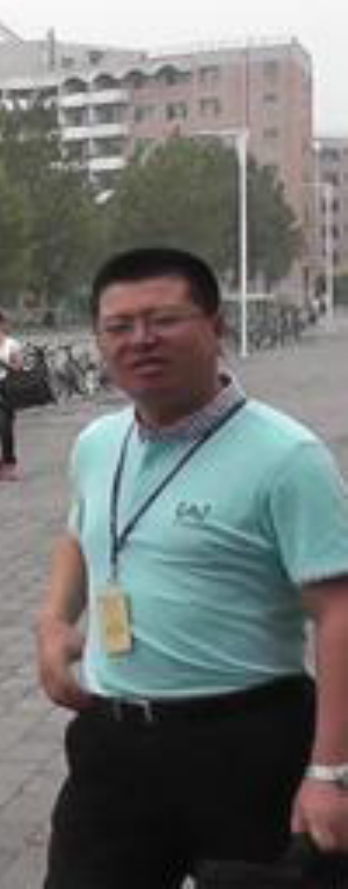}};  	
\draw (4*\sizeImg,0+1.22*\deltaY) node(segment_ok)  {\includegraphics[width=\sizeImg cm]{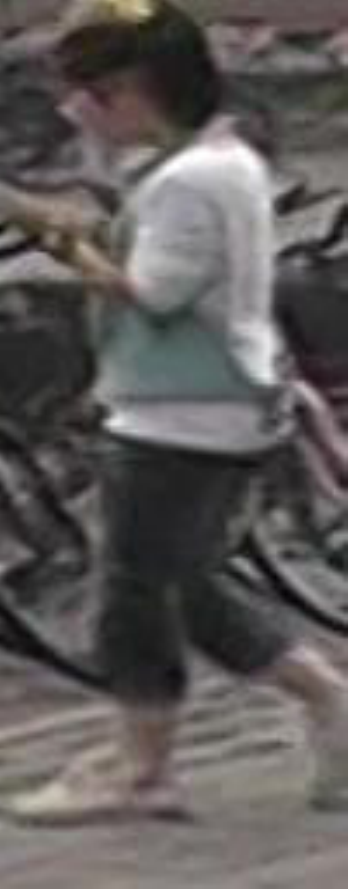}};  	
\draw (5*\sizeImg,0+1.22*\deltaY) node(segment_ok)  {\includegraphics[width=\sizeImg cm]{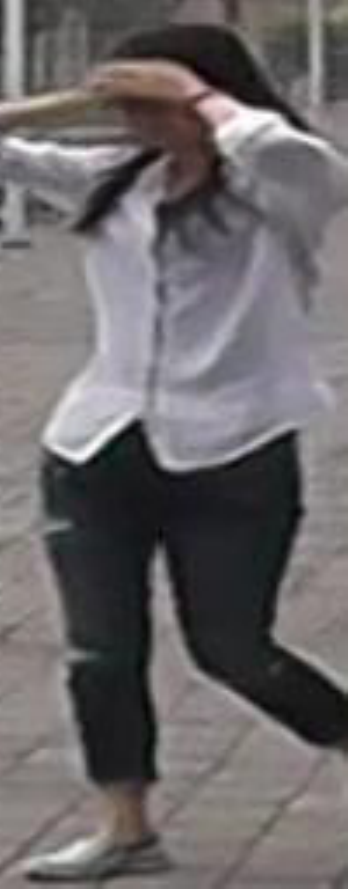}};  	
\draw (-1.0, 0+1.22*\deltaY) node[rectangle, rotate=90] {\small{\textbf{MARS}}};  

\draw (0*\sizeImg,0+2.665*\deltaY) node(segment_ok)  {\includegraphics[width=\sizeImg cm]{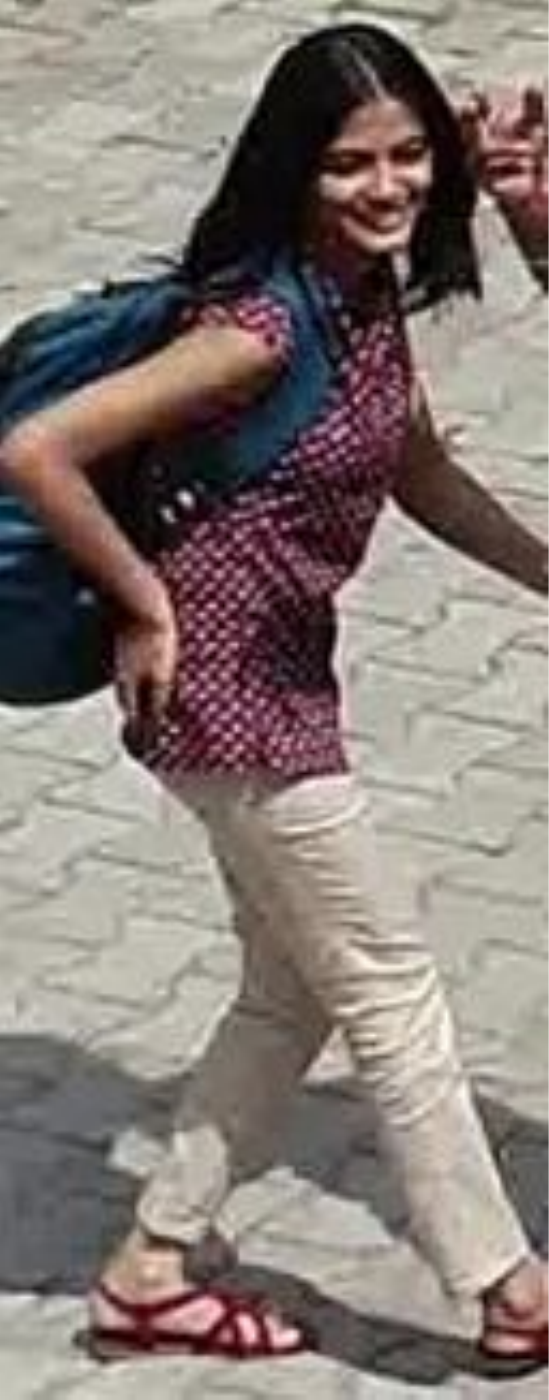}};  	
\draw (1*\sizeImg,0+2.665*\deltaY) node(segment_ok)  {\includegraphics[width=\sizeImg cm]{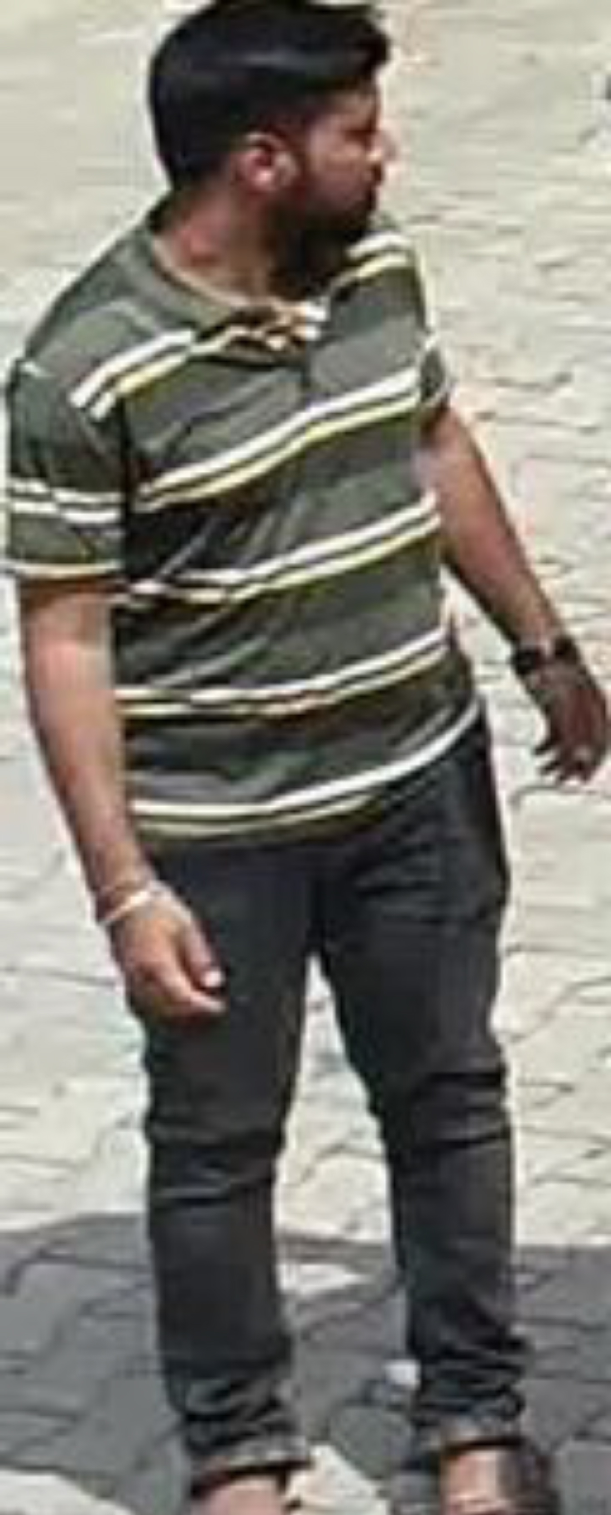}};  	
\draw (2*\sizeImg,0+2.665*\deltaY) node(segment_ok)  {\includegraphics[width=\sizeImg cm]{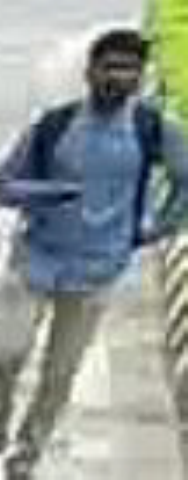}};  	
\draw (3*\sizeImg,0+2.665*\deltaY) node(segment_ok)  {\includegraphics[width=\sizeImg cm]{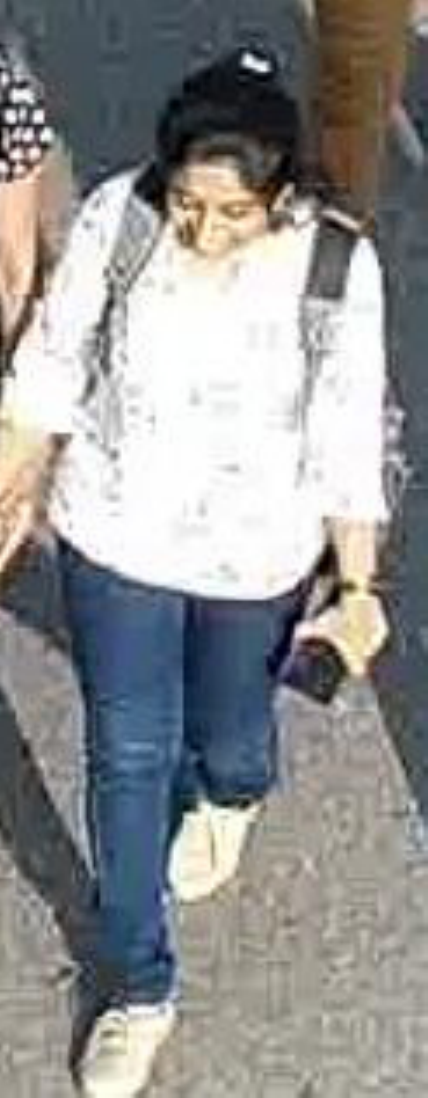}};  	
\draw (4*\sizeImg,0+2.665*\deltaY) node(segment_ok)  {\includegraphics[width=\sizeImg cm]{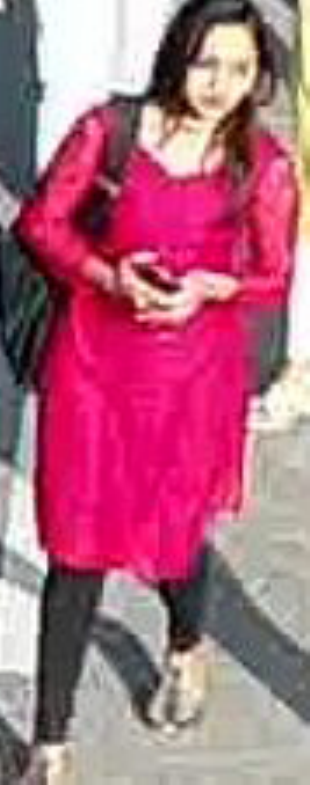}};  	
\draw (5*\sizeImg,0+2.665*\deltaY) node(segment_ok)  {\includegraphics[width=\sizeImg cm]{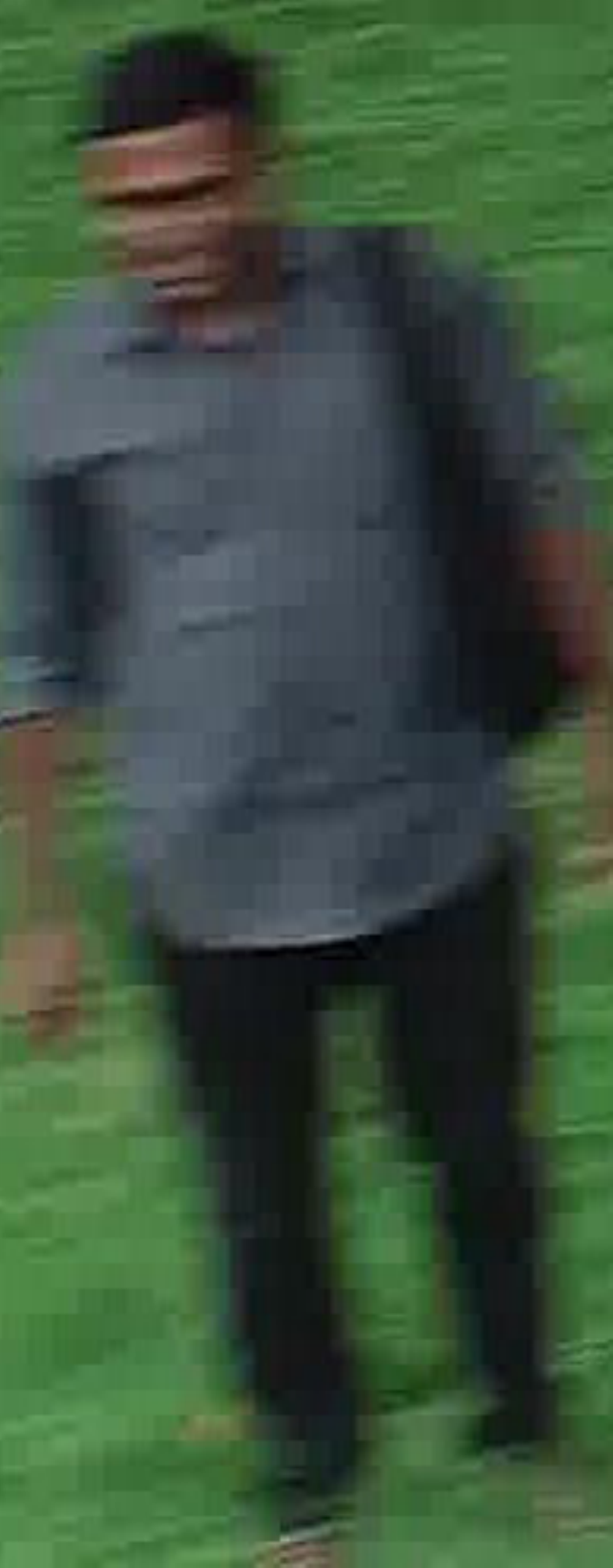}};  	
\draw (-1.0, 0+2.665*\deltaY) node[rectangle, rotate=90] {\small{\textbf{P-DESTRE}}};  

\end{tikzpicture}
    \caption{Datasets used in the empirical validation of the method proposed in this paper. From top to bottom rows, images of the BIODI, MARS and P-DESTRE are shown.}
        \label{fig:Datasets}
    \end{center}
\end{figure}
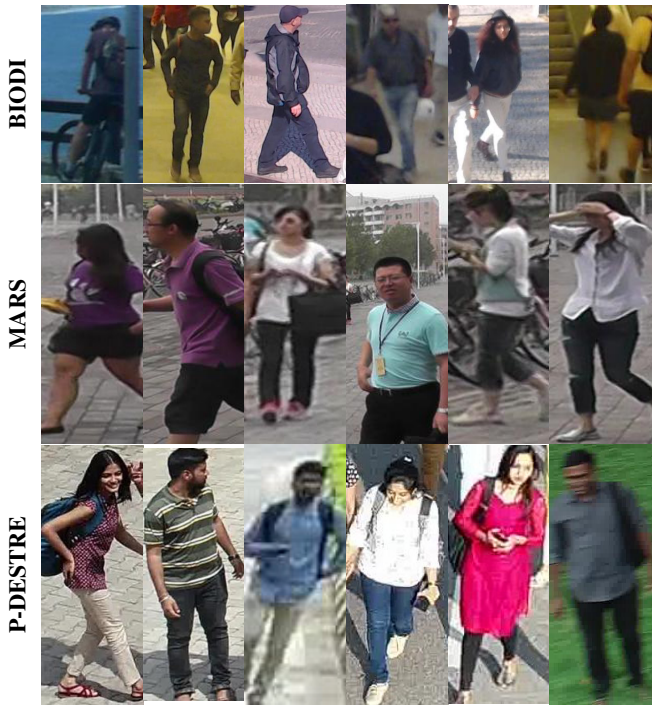

The image pairwise matcher \textbf{D}$_a$ uses a classic VGG-like architecture, detailed in Table~\ref{tab:CNNs}. A different model was inferred independently for each label (\emph{ID}, \emph{gender}, \emph{ethnicity}, \emph{age} and \emph{hairstyle}). Then, during inference, the pairs of RGB images to be matched were resized and concatenated along the depth axis, resulting in $64 \times 64 \times 6$ inputs, from where the 5-dimensional binary output vectors were inferred.  

For the second learning phase, both the encoder \textbf{U}$_e$ and decoder \textbf{U}$_d$ models share the well known \emph{U-Net} architecture~\cite{Ronneberger2015}, with a minor adaptation to receive $64 \times 64 \times 4$ (encoder) and $64 \times 64 \times 3$ (decoder) data. The encoder receives the raw facial images represented in the RGB space (scaled to the unit interval) and a forth channel of random values drew from a uniform distribution in the unit interval $\mathcal{U}(0,1)$. The adversarial discriminative model \textbf{D}$_f$ uses the \emph{PatchGAN}~\cite{Isola2017} architecture. Upon empirical optimization and grounded on the human perception of the generated \textbf{a}$_.$ elements, we set $\delta_{\text{gp}}$=0.01 to assure the stability of the adversarial learning process and the weight parameters: $\omega_{\text{mse}}$=50,  $\omega_{\text{adv}}$=1, $\omega_{\text{ano}}$=1, $\omega_{\text{con}}$=1, $\omega_{\text{dis}}$=1 and $\omega_{\text{div}}$=1. 

\begin{table}[h!]
\centering
     \caption{Architecture of the CNN models used in our experiments. ('nk': number of kernels; 'ks': kernel size; 'st': stride; 'mm': momentum).}
     \label{tab:CNNs}
\begin{tabular}{|p{6cm}|p{2.0cm}|}
\hline
\multicolumn{1}{|c|}{    \cellcolor{gray!50} \scriptsize{\textbf{D}$_a$ model}}  & \multicolumn{1}{c|}{    \cellcolor{gray!50} \scriptsize{\textbf{U}$_e$ model}} \\ \hline
 \multirow{5}{*}{
\begin{tabular}{p{5.7cm}} 
\scriptsize{Input: ($64 \times 64 \times 6$) $\rightarrow$ Convolution (nk: 16, ks: 3 $\times$ 3, st: 2) $\rightarrow$  Batch Normalization (mm: 0.8) $\rightarrow$ LeakyReLU $\rightarrow$ Dropout (0.25) $\rightarrow$ Convolution (nk: 64, ks: 3 $\times$ 3, st: 1) $\rightarrow$ Batch Normalization (mm: 0.8) $\rightarrow$ LeakyReLU $\rightarrow$ Dropout (0.25) $\rightarrow$ Convolution (nk: 128, ks: 3 $\times$ 3, st: 1) $\rightarrow$ Batch Normalization (mm: 0.8) $\rightarrow$ LeakyReLU $\rightarrow$ Dropout (0.25) $\rightarrow$ [Convolution (nk: 64, ks: 3 $\times$ 3, st: 2) $\rightarrow$ BN (mm: 0.8) $\rightarrow$ LeakyReLU $\rightarrow$ Dropout (0.25)] $\times$ 2 $\rightarrow$ Flatten $\rightarrow$ Dense (128) $\rightarrow$ ReLU $\rightarrow$ Dense (t) $\rightarrow$ Sigmoid}
\end{tabular}
 }
 	& \multicolumn{1}{p{2.0cm}|}{\scriptsize{Input: ($64 \times 64 \times 4$) $\rightarrow$ \emph{U-Net}~\cite{Ronneberger2015}}} \\ \cline{2-2}
	 	&\multicolumn{1}{c|}{    \cellcolor{gray!50} \scriptsize{\textbf{U}$_d$ model}}\\ \cline{2-2}
		 	& \multicolumn{1}{p{2.0cm}|}{ \scriptsize{Input: ($64 \times 64 \times 3$) $\rightarrow$\emph{U-Net}~\cite{Ronneberger2015}}}\\ \cline{2-2}
			 	&\multicolumn{1}{c|}{    \cellcolor{gray!50} \scriptsize{\textbf{D}$_f$ model}} \\ \cline{2-2}
       &\multicolumn{1}{p{2.0cm}|}{\scriptsize{Input: ($64 \times 64 \times 3$) $\rightarrow$ \emph{PatchGAN}~\cite{Isola2017}}}\\ \hline
\end{tabular}
\end{table}

To our knowledge, there are no prior works to perform reversible de-identification in video data. Though, we considered the two baselines to compare the face detection effectiveness in de-identified data:  the Super-pixel~\cite{Butler2015} method: that replaces each pixel by the average value of the corresponding super-pixel and the classical Blur-based method~\cite{Ryoo2017}, where images are downsampled to extreme low-resolution and then upsampled back.

\subsection{Face Detection}

\begin{figure}[ht!]
\begin{center}
\begin{tikzpicture}

\fill [gray!50, rounded corners] (1.3, 1.7) rectangle (3.1, 2.1);     
\node at (2.2, 1.9) [black, rotate=0]  {\scriptsize{\textbf{BIODI}}};     

\draw (0,0) node(n1)  {\includegraphics[width=4 cm]{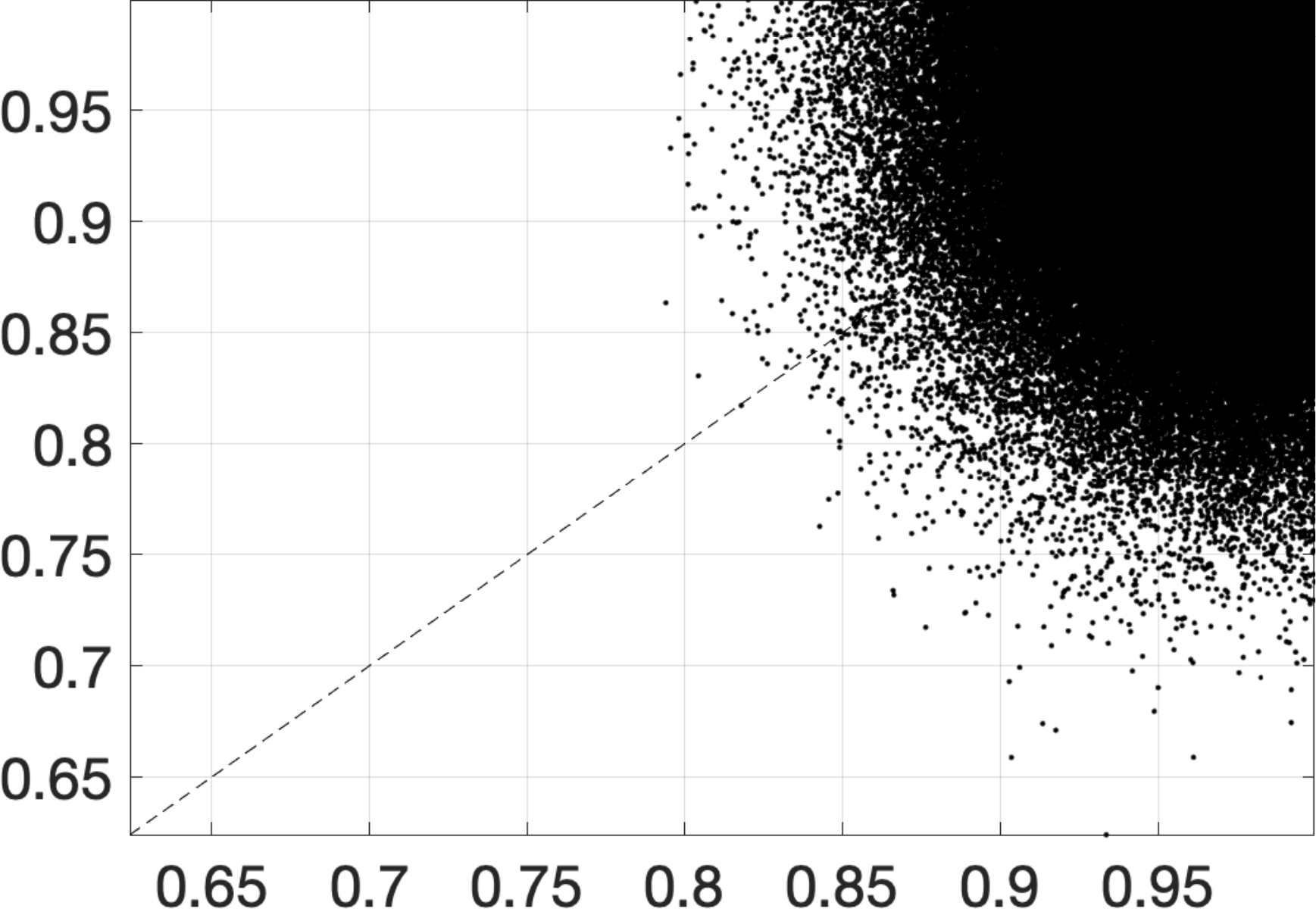}};   
\node at (0, -1.7) [black, rotate=0]  {\scriptsize{f$_{\text{c}}$(\textbf{x})}};     
\node at (-2.2, 0) [black, rotate=90]  {\scriptsize{f$_{\text{c}}$(\textbf{a})}};     

\draw (4.4,0.85) node(n1)  {\includegraphics[width=3.75 cm]{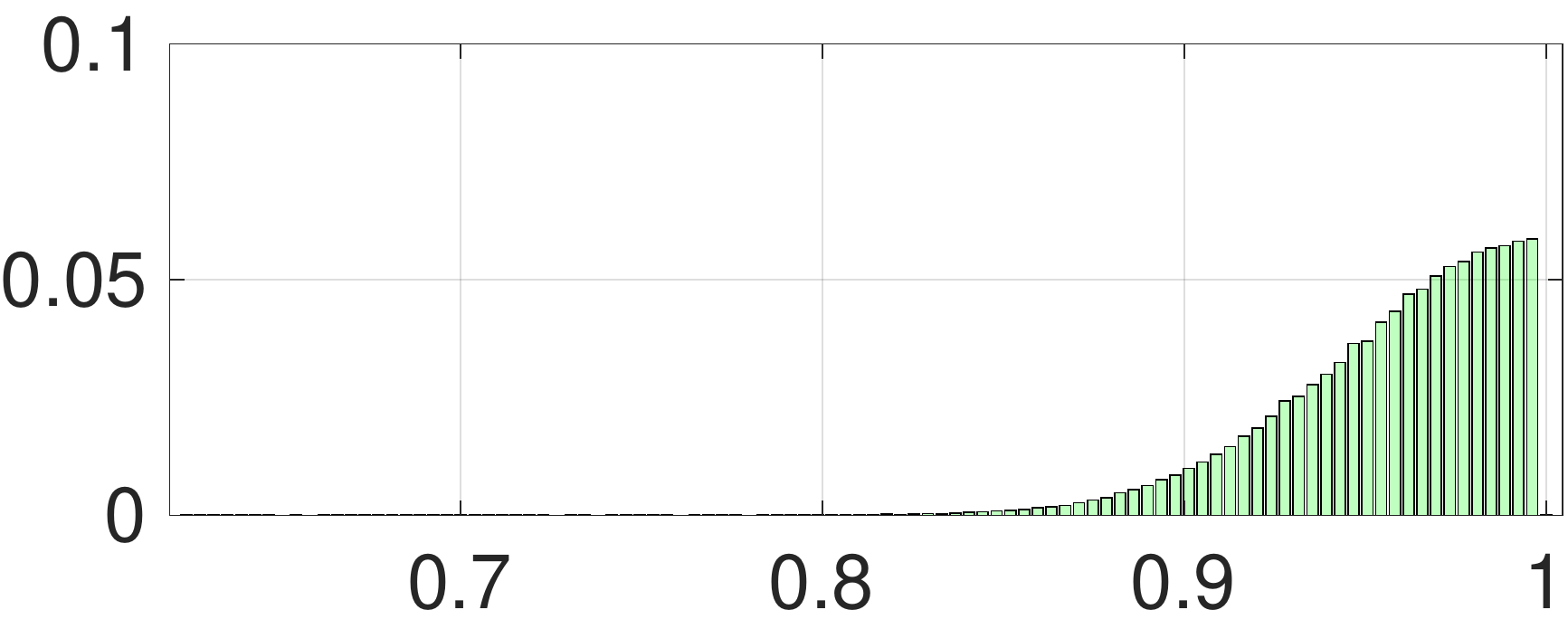}};   
 
\node at (2.4, 0.85) [black, rotate=90]  {\scriptsize{Frequency}};   

\draw (4.4,-0.85) node(n1)  {\includegraphics[width=3.75 cm]{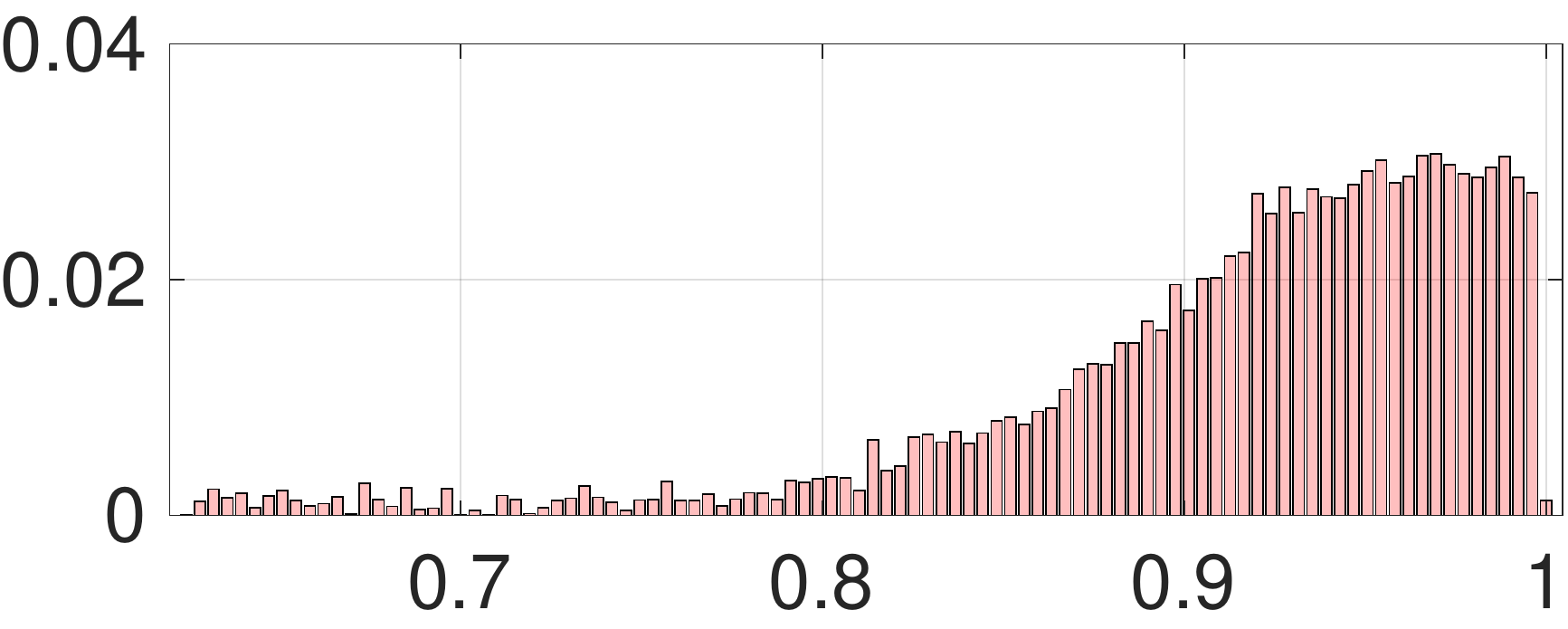}};   
\node at (4.4, -1.7) [black, rotate=0]  {\scriptsize{f$_{\text{c}}$(\textbf{a})}};     
\node at (4.4, -0.0) [black, rotate=0]  {\scriptsize{f$_{\text{c}}$(\textbf{x})}};    
\node at (2.4, -0.85) [black, rotate=90]  {\scriptsize{Frequency}};

\def\deltaY{-4}

\draw (0,0+\deltaY) node(n1)  {\includegraphics[width=4 cm]{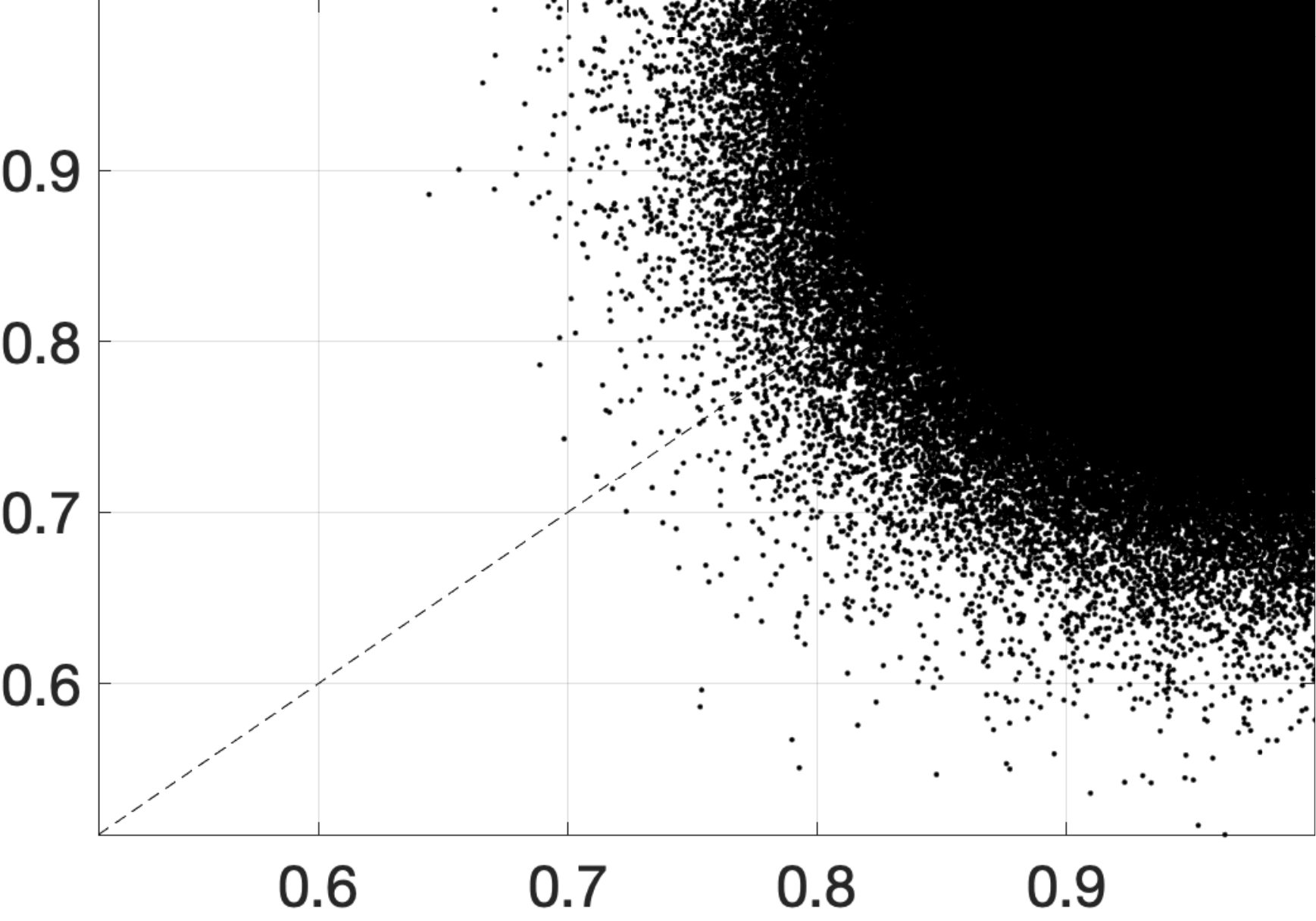}};   
\node at (0, -1.7+\deltaY) [black, rotate=0]  {\scriptsize{f$_{\text{c}}$(\textbf{x})}};     
\node at (-2.2, 0+\deltaY) [black, rotate=90]  {\scriptsize{f$_{\text{c}}$(\textbf{a})}};     

\draw (4.4,0.85+\deltaY) node(n1)  {\includegraphics[width=3.75 cm]{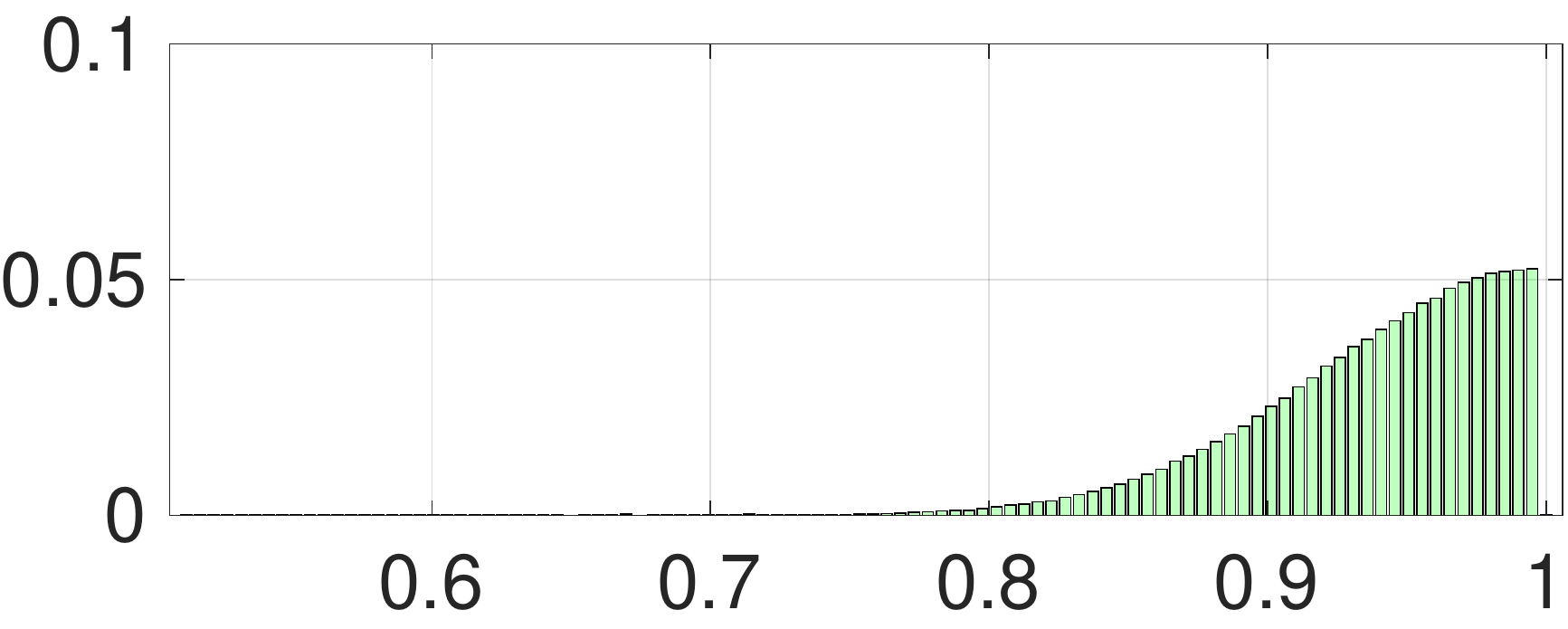}};   
 
\node at (2.4, 0.85+\deltaY) [black, rotate=90]  {\scriptsize{Frequency}};   

\draw (4.4,-0.85+\deltaY) node(n1)  {\includegraphics[width=3.75 cm]{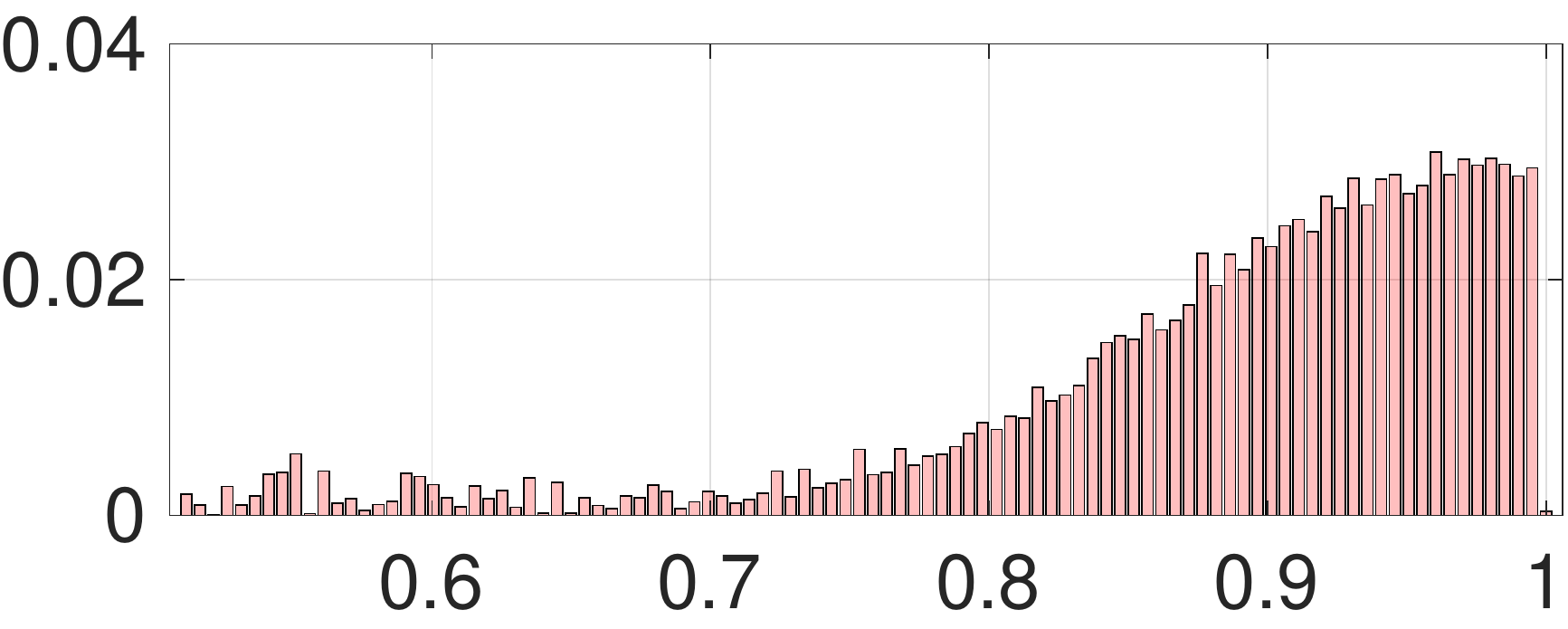}};   
\node at (4.4, -1.7+\deltaY) [black, rotate=0]  {\scriptsize{f$_{\text{c}}$(\textbf{a})}};     
\node at (4.4, -0.0+\deltaY) [black, rotate=0]  {\scriptsize{f$_{\text{c}}$(\textbf{x})}};    
\node at (2.4, -0.85+\deltaY) [black, rotate=90]  {\scriptsize{Frequency}};   

\fill [gray!50, rounded corners] (1.3, 1.7+\deltaY) rectangle (3.1, 2.1+\deltaY);     
\node at (2.2, 1.9+\deltaY) [black, rotate=0]  {\scriptsize{\textbf{MARS}}};     

\def\deltaY{-8}

\draw (0,0+\deltaY) node(n1)  {\includegraphics[width=4 cm]{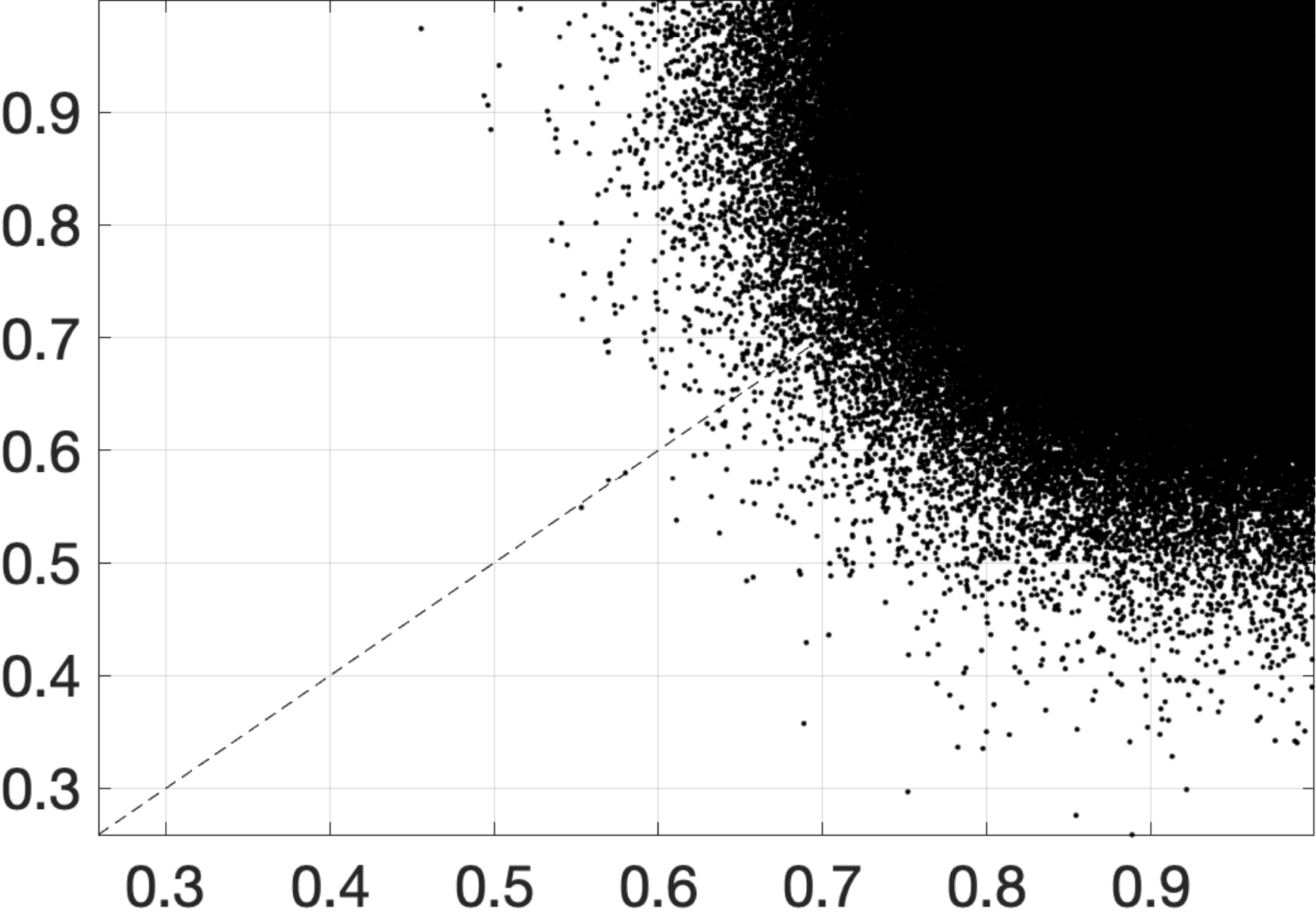}};   
\node at (0, -1.7+\deltaY) [black, rotate=0]  {\scriptsize{f$_{\text{c}}$(\textbf{x})}};     
\node at (-2.2, 0+\deltaY) [black, rotate=90]  {\scriptsize{f$_{\text{c}}$(\textbf{a})}};     

\draw (4.4,0.85+\deltaY) node(n1)  {\includegraphics[width=3.75 cm]{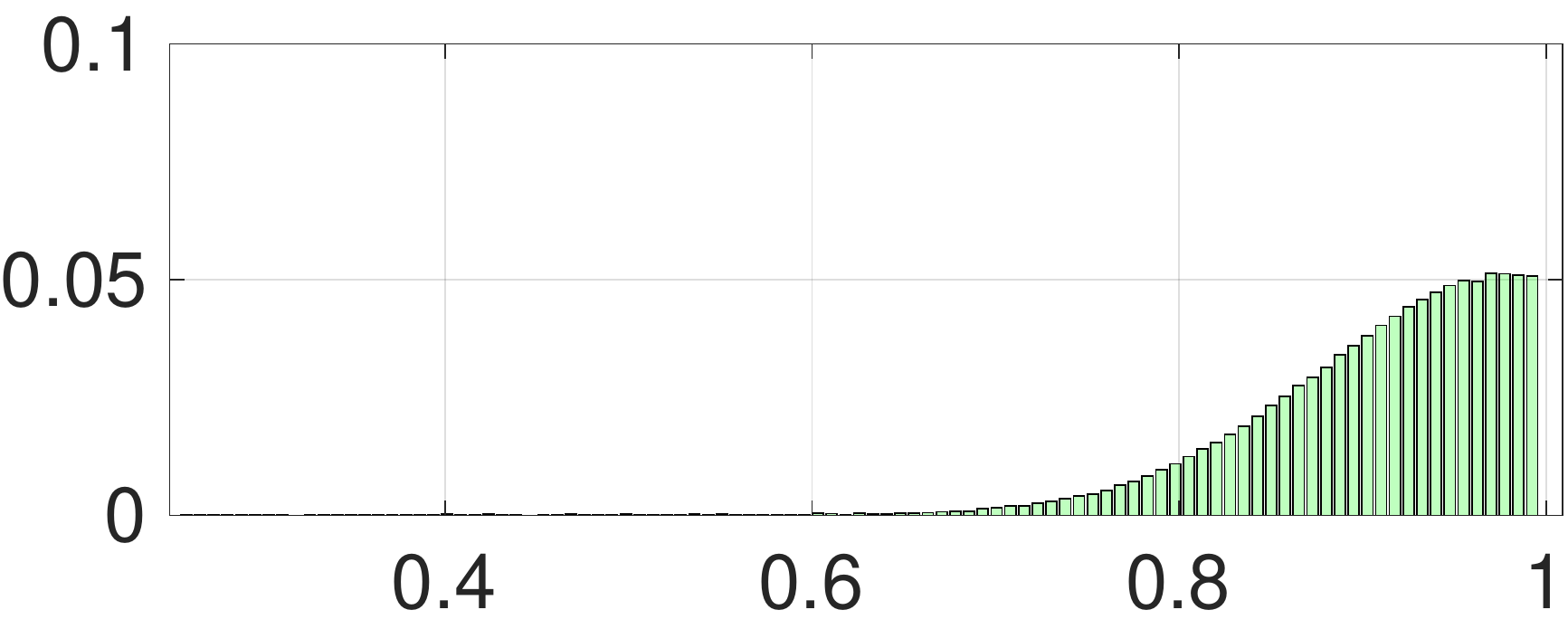}};   
 
\node at (2.4, 0.85+\deltaY) [black, rotate=90]  {\scriptsize{Frequency}};   

\draw (4.4,-0.85+\deltaY) node(n1)  {\includegraphics[width=3.75 cm]{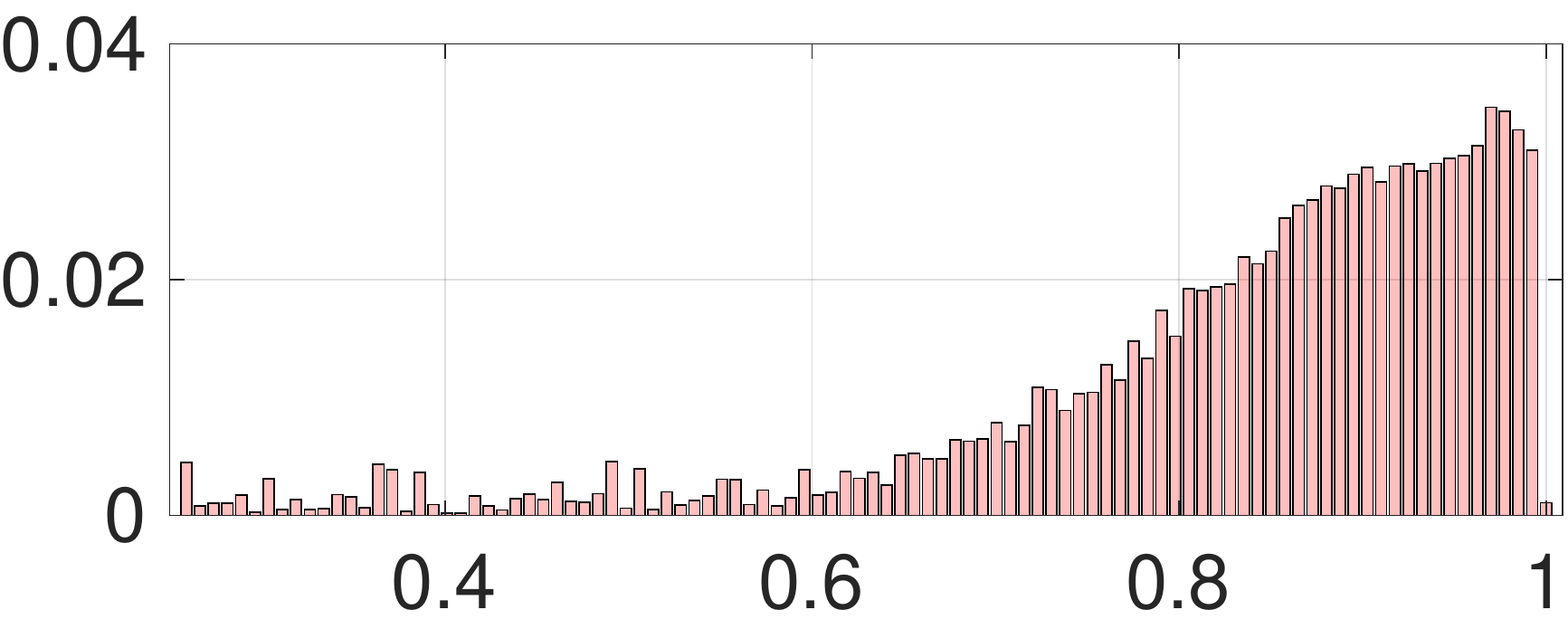}};   
\node at (4.4, -1.7+\deltaY) [black, rotate=0]  {\scriptsize{f$_{\text{c}}$(\textbf{a})}};     
\node at (4.4, -0.0+\deltaY) [black, rotate=0]  {\scriptsize{f$_{\text{c}}$(\textbf{x})}};    
\node at (2.4, -0.85+\deltaY) [black, rotate=90]  {\scriptsize{Frequency}};   

\fill [gray!50, rounded corners] (1.3, 1.7+\deltaY) rectangle (3.1, 2.1+\deltaY);     
\node at (2.2, 1.9+\deltaY) [black, rotate=0]  {\scriptsize{\textbf{P-DESTRE}}};     

\end{tikzpicture}
    \caption{Scatter plot of the correlation between the MTCNN~\cite{Zhang2016} face detection confidence scores f$_{\text{c}}$(.) for \textbf{x}$_.$ and \textbf{a}$_.$ elements, obtained for the BIODI, MARS and P-DESTRE sets (left plots). The corresponding histograms of the \textbf{x}$_.$ and \textbf{a}$_.$ values are given at the right side.}
        \label{fig:faceDetection}
    \end{center}
\end{figure}
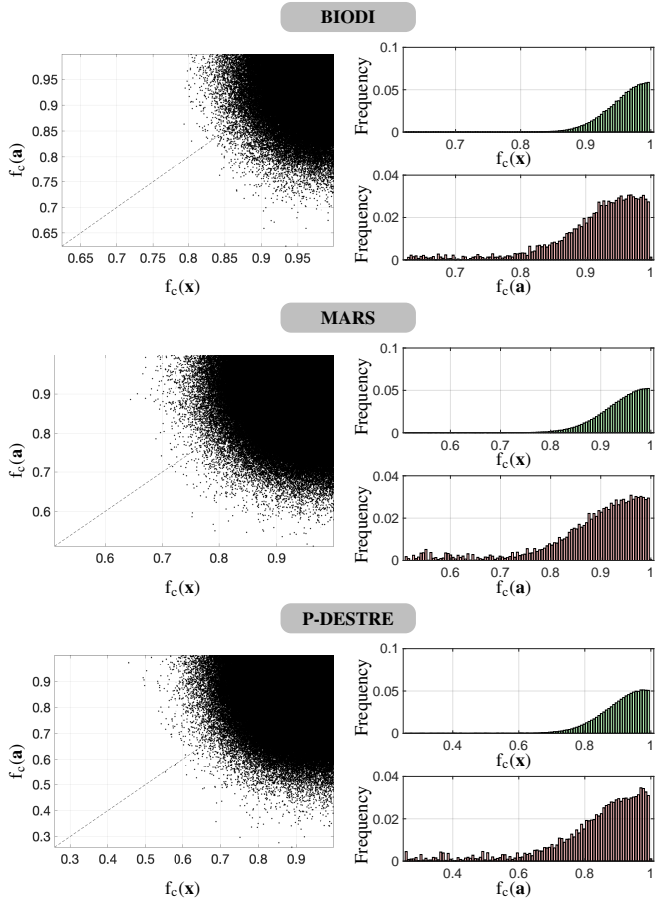

The photo-realism of the de-identified data was evaluated by comparing the face detection scores in the raw \textbf{x}$_.$ and de-identified \textbf{a}$_.$ samples, according to~\cite{Zhang2016}. This method provides a confidence score f$_c$(.) for having a face at a certain position. The results for the three data sets are shown in Fig.~\ref{fig:faceDetection}, where the scatter plots correlate the confidence values between \textbf{x} (horizontal axis) and \textbf{a} elements (vertical axis). The histograms provide the distributions for \textbf{x}/\textbf{a} values.  Overall, the average values for \textbf{a} images decreased about 2.11\% (BIODI), 3.22\% (MARS) and  4.40\% (P-DESTRE) with respect to \textbf{x} elements (BIODI: f$_c$(\textbf{x})= 0.954 $\rightarrow$ f$_c$(\textbf{a})= 0.931, MARS: f$_c$(\textbf{x})= 0.968 $\rightarrow$ f$_c$(\textbf{a})= 0.930 and P-DESTRE: f$_c$(\textbf{x})= 0.918 $\rightarrow$ f$_c$(\textbf{a})= 0.877). Also, minimal levels of linear correlation were observed between the \textbf{x}/\textbf{a} values, with Pearson's coefficients of 0.011 (BIODI),  0.023 (MARS) and 0.025 (P-DESTRE). 

\begin{table}[h!]
\centering
     \caption{Comparison between the face detection performance~\cite{Zhang2016} in the de-identified data, with respect to the baseline detection values. Average $\pm$ standard deviation ''mean Average Precision' values are given.}
     \label{tab:faceDetection}
\begin{tabular}{|l|C{1.3cm}|C{1.2cm}|C{1.2cm}|C{1.2cm}|}
\hline
\textbf{\scriptsize{Method}}  & \textbf{\scriptsize{Params.}}  & \textbf{\scriptsize{BIODI}} & \textbf{\scriptsize{MARS}}  & \textbf{\scriptsize{P-DESTRE}}\\ \hline
\multicolumn{2}{|l|}{\scriptsize{\textbf{Baseline Detection mAP~\cite{Zhang2016} (\textbf{x}$_.$)}}} & \scriptsize{0.82} \tiny{$\pm$ 0.02} & \scriptsize{0.84} \tiny{$\pm$ 0.03} & \scriptsize{0.63} \tiny{$\pm$ 0.01}  \\ \hline
\scriptsize{Proposed} & \scriptsize{$\omega_{\text{mse}}$=50,  $\omega_{\text{adv}}$=1, $\omega_{\text{ano}}$=1, $\omega_{\text{con}}$=1, $\omega_{\text{div}}$=1, $\omega_{\text{dis}}$=1} & \scriptsize{0.73} \tiny{$\pm$ 0.06} & \scriptsize{0.75} \tiny{$\pm$ 0.06} & \scriptsize{0.59} \tiny{$\pm$ 0.10} \\ \hline
\scriptsize{Butler \etal~\cite{Butler2015}} & \scriptsize{8 superpixels} & \scriptsize{0.23} \tiny{$\pm$ 0.06} & \scriptsize{0.21} \tiny{$\pm$ 0.04} & \scriptsize{0.20} \tiny{$\pm$ 0.04} \\ \hline
\scriptsize{Butler \etal~\cite{Butler2015}} & \scriptsize{16 superpixels} & \scriptsize{0.36} \tiny{$\pm$ 0.05} & \scriptsize{0.32} \tiny{$\pm$ 0.04} & \scriptsize{0.22} \tiny{$\pm$ 0.05} \\ \hline

\scriptsize{Ryoo \etal~\cite{Ryoo2017}} & \scriptsize{resolution 5x3} & \scriptsize{0.20} \tiny{$\pm$ 0.03} & \scriptsize{0.19} \tiny{$\pm$ 0.03} & \scriptsize{0.17} \tiny{$\pm$ 0.02} \\ \hline
\scriptsize{Ryoo \etal~\cite{Ryoo2017}} & \scriptsize{resolution 7x4} & \scriptsize{0.31} \tiny{$\pm$ 0.04} & \scriptsize{0.27} \tiny{$\pm$ 0.04} & \scriptsize{0.16} \tiny{$\pm$ 0.06} \\ \hline

\end{tabular}
\end{table}

Table~\ref{tab:faceDetection}  summarizes the face detection effectiveness~\cite{Zhang2016} on \textbf{a}$_.$ data, with respect to the performance in \textbf{x}$_.$. Also, as baselines, we provide the results obtained by two simple de-identification techniques (due to Butler \etal~\cite{Butler2015} and Ryoo \etal~\cite{Ryoo2017}). For these experiments, random samples composed of 90\% of the test samples were created (drew with repetition) and the mean Average Precision (mAP) taken in each split, from where the mean and standard deviation values were  taken. The proposed solution attained mAP values for \textbf{a}$_.$ elements that were about 89\% (BIODI), 89\% (MARS) and 93\% (P-DESTRE) of the values obtained for \textbf{x}$_.$. Errors occurred typically for extreme poor resolution samples where the detection method was still able to find the original face, but not the de-identified version. Also, the de-identified versions tend to feature less details (i.e., lower entropy) than the original samples, which might justify this gap in performance. The remaining techniques got far worse performance, even stressing that both not aim at reversibility.   

\subsection{Face Recognition}

\begin{figure}[ht!]
\begin{center}
\begin{tikzpicture}

\def\deltaY{0}

\node at (0, 1.35) [black, rotate=0]  {\scriptsize{\textbf{x $\leftrightarrow$ x'}}};     
\draw (0,0) node(n1)  {\includegraphics[width=4 cm]{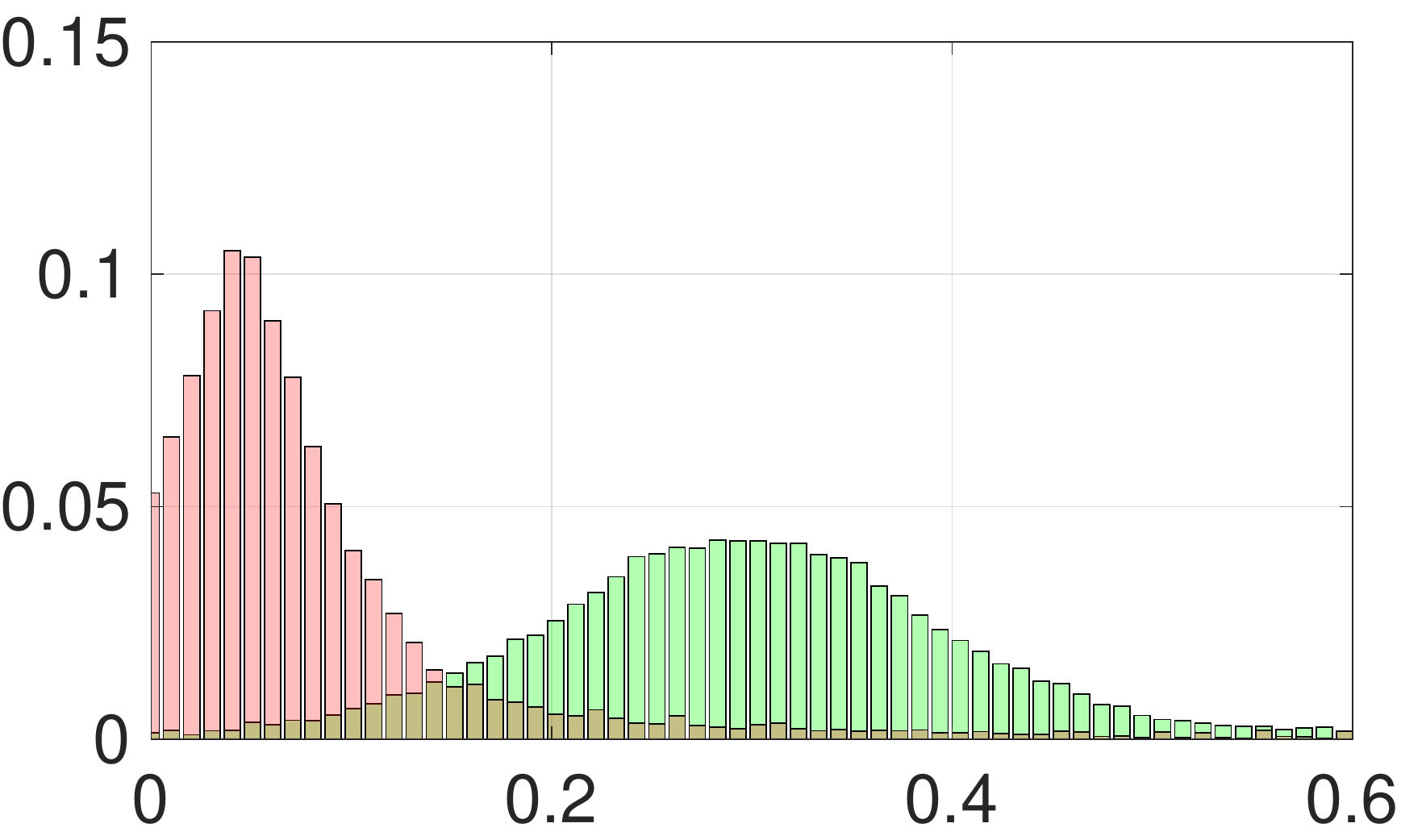}};   
\node at (0, -1.35) [black, rotate=0]  {\scriptsize{Score}};     
\node at (-2.2, 0) [black, rotate=90]  {\scriptsize{Frequency}};     

\node at (4.5, 1.35) [black, rotate=0]  {\scriptsize{\textbf{x $\leftrightarrow$ a'}}};   
\draw (4.5,0) node(n1)  {\includegraphics[width=4 cm]{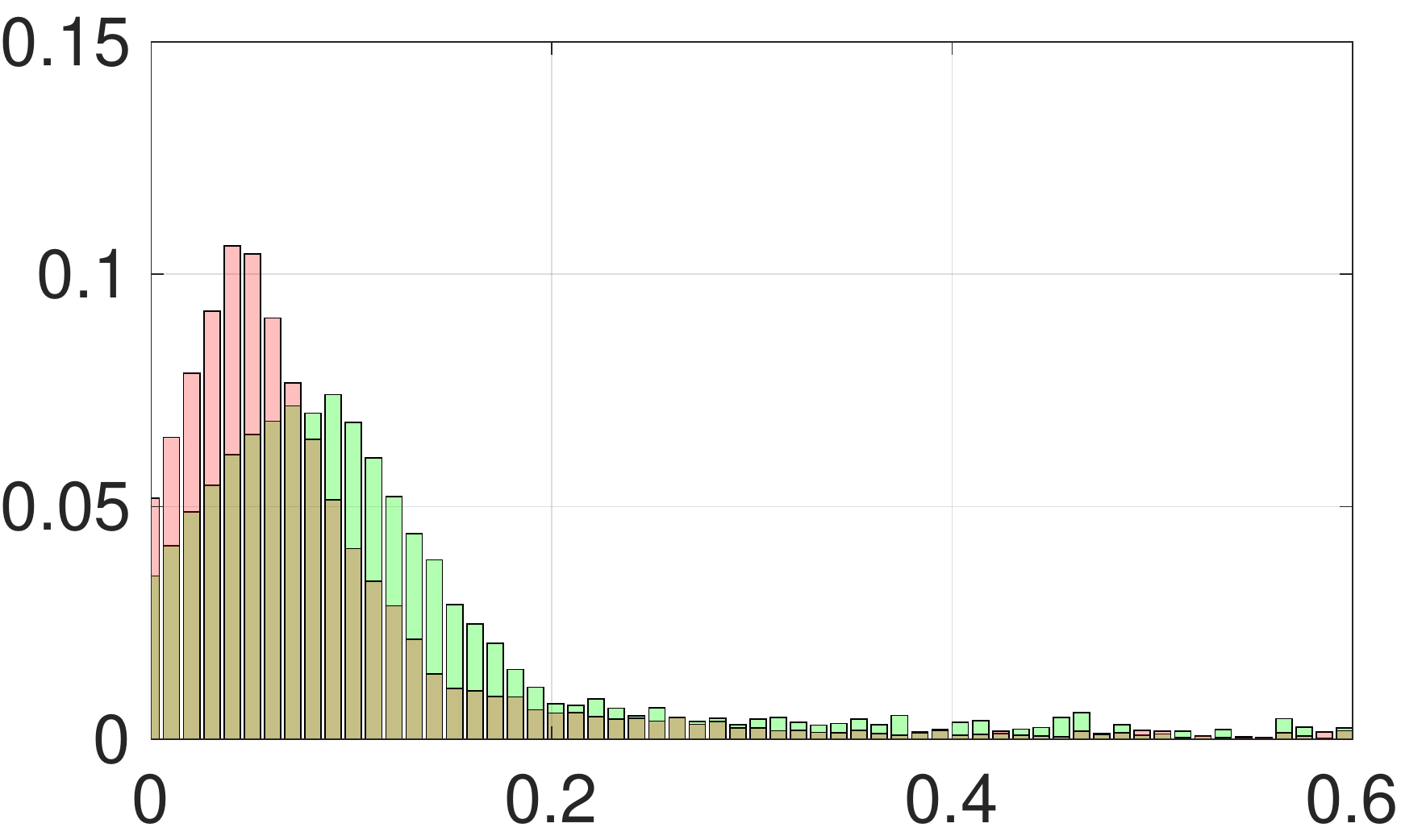}};   
\node at (0+4.5, -1.35) [black, rotate=0]  {\scriptsize{Score}};     
\node at (-2.2+4.5, 0) [black, rotate=90]  {\scriptsize{Frequency}};     

\fill [gray!50, rounded corners] (1.35, 1.4+\deltaY) rectangle (3.15, 1.8+\deltaY);     
\node at (2.25, 1.6+\deltaY) [black, rotate=0]  {\scriptsize{\textbf{MARS}}};     

\def\deltaY{-3.25}

\node at (0, 1.35+\deltaY) [black, rotate=0]  {\scriptsize{\textbf{x $\leftrightarrow$ x'}}};     
\draw (0,0+\deltaY) node(n1)  {\includegraphics[width=4 cm]{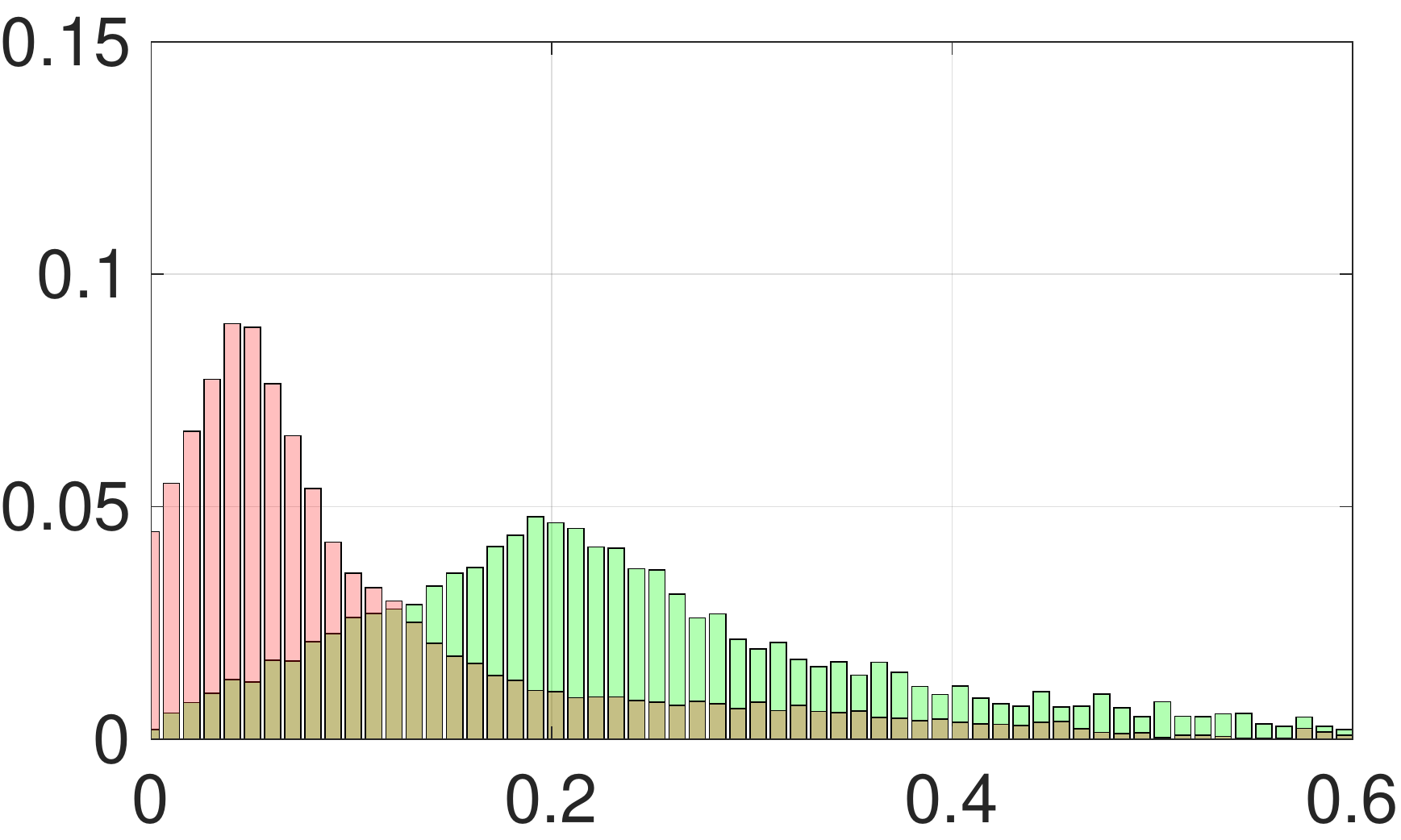}};   
\node at (0, -1.35+\deltaY) [black, rotate=0]  {\scriptsize{Score}};     
\node at (-2.2, 0+\deltaY) [black, rotate=90]  {\scriptsize{Frequency}};     

\node at (4.5, 1.35+\deltaY) [black, rotate=0]  {\scriptsize{\textbf{x $\leftrightarrow$ a'}}};   
\draw (4.5,0+\deltaY) node(n1)  {\includegraphics[width=4 cm]{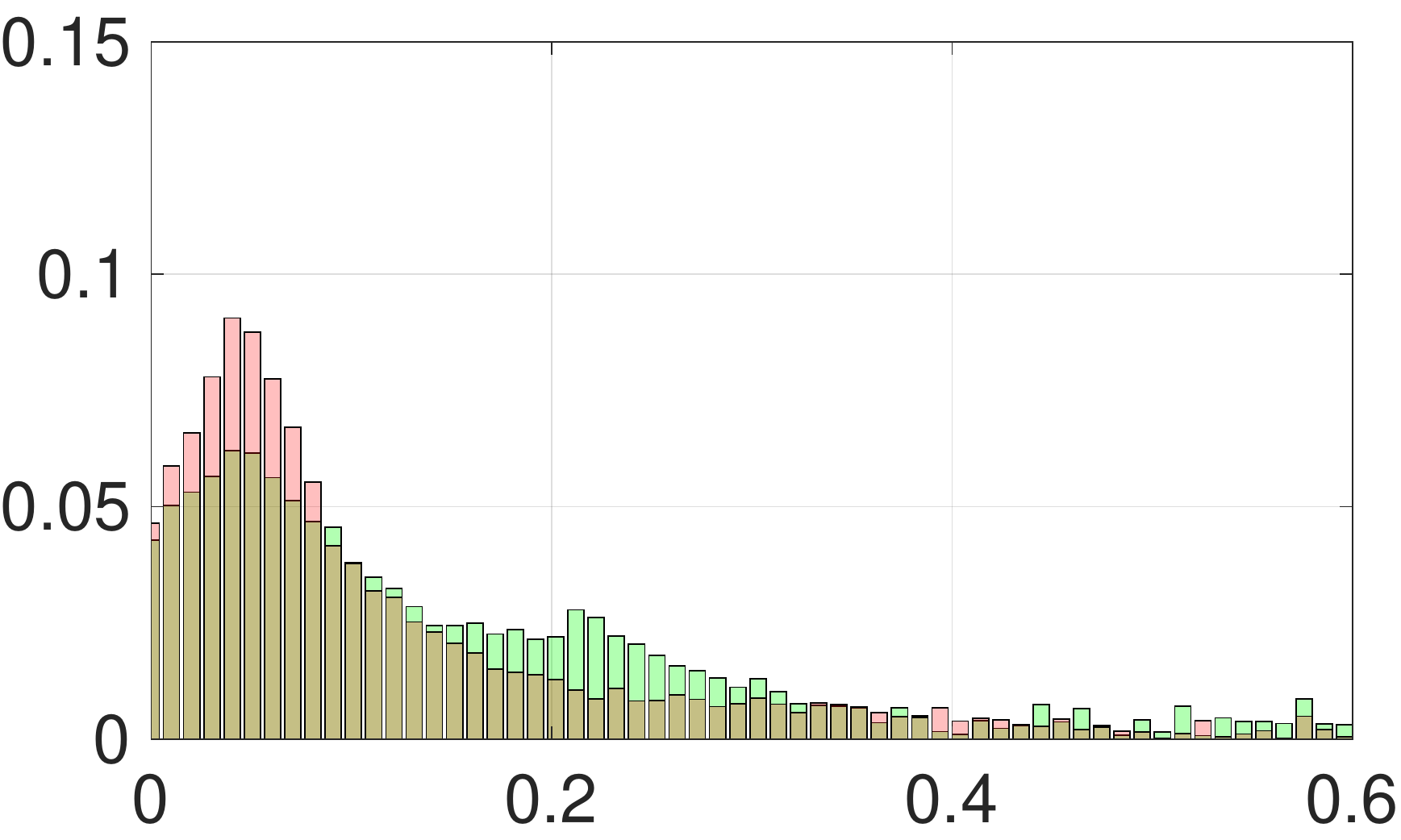}};   
\node at (0+4.5, -1.35+\deltaY) [black, rotate=0]  {\scriptsize{Score}};     
\node at (-2.2+4.5, 0+\deltaY) [black, rotate=90]  {\scriptsize{Frequency}};     

\fill [gray!50, rounded corners] (1.35, 1.4+\deltaY) rectangle (3.15, 1.8+\deltaY);     
\node at (2.25, 1.6+\deltaY) [black, rotate=0]  {\scriptsize{\textbf{P-DESTRE}}};     

\end{tikzpicture}
    \caption{Comparison between the decision environments resulting from the VGG-Face2~\cite{Cao2018b} face recognition method in pairs of images of the MARS and P-DESTRE datasets (\textbf{x} $\leftrightarrow$ \textbf{x}', left plots) and when the second image in each pairwise comparison was de-identified (\textbf{x} $\leftrightarrow$ \textbf{a}, right plots).}
        \label{fig:decisionEnv}
    \end{center}
\end{figure}
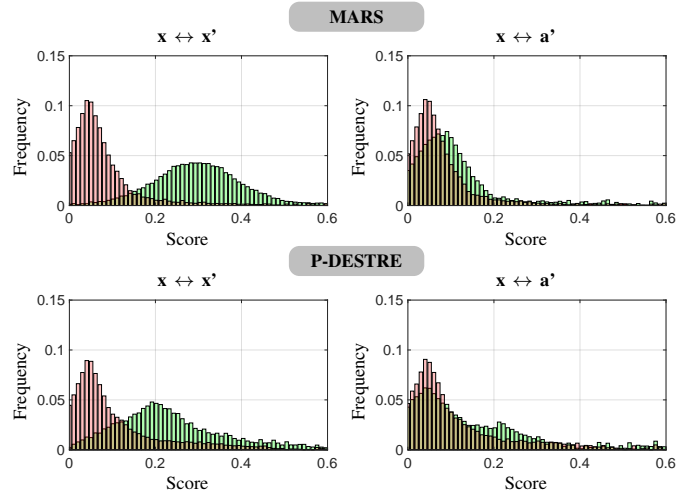

A second important question to address is the possibility of models being simply swapping faces between identities, rather than creating virtual IDs. To verify this hypothesis, we learned two VGG-Face2~\cite{Cao2018b} recognizers (for MARS and for the P-DESTRE sets), including 80\% of the IDs in the learning set (\textbf{x}$_.$ elements). Then, in inference time, we sampled the remaining 20\% IDs and created 50K \emph{impostors} + 10K {genuine} pairwise comparisons (\textbf{x} $\leftrightarrow$ \textbf{x}). This experiment was repeated when de-identifying the second image of each pair (i.e., \textbf{x} $\leftrightarrow$ \textbf{a}). Results are given in Fig.~\ref{fig:decisionEnv}, that compares the decision environments for the MARS (top plots) and P-DESTRE (bottom plots) sets. The green bars correspond to the distributions of the \emph{genuine} scores, while the red bars denote the \emph{impostors} distributions.  The decidability values of the decision environments were also obtained ($d'=\frac{\mu_G-\mu_I}{\sqrt{\sigma_G^2+\sigma_I^2}}$), where $(\mu, \sigma)$ denote the mean and standard deviation statistics and 'G'/'I' stand for the \emph{genuine} and \emph{impostors} pairwise comparisons. Values decreased from 1.885 (MARS) and 0.839 (P-DESTRE) (\textbf{x} $\leftrightarrow$ \textbf{x}) to 0.162 (MARS) and 0.155 (P-DESTRE)(\textbf{x} $\leftrightarrow$ \textbf{a}), with an evident movement of the genuine distributions toward the impostors region. The corresponding decreases in the AUC values were of MARS: 0.962 $\rightarrow$ 0.570 and P-DESTRE: 0.820 $\rightarrow$ 0.568, in both cases turning the identification based in the \textbf{a}$_.$ samples almost equivalent to a random choice. Even though, a slight difference between the right tails of the genuine/impostors distributions was observed in both sets, which was justified by potential overfitting problems, i.e., lack of sufficient learning data to sustain enough variability in the de-identified space of identities. 

\subsection{Temporal Consistency}

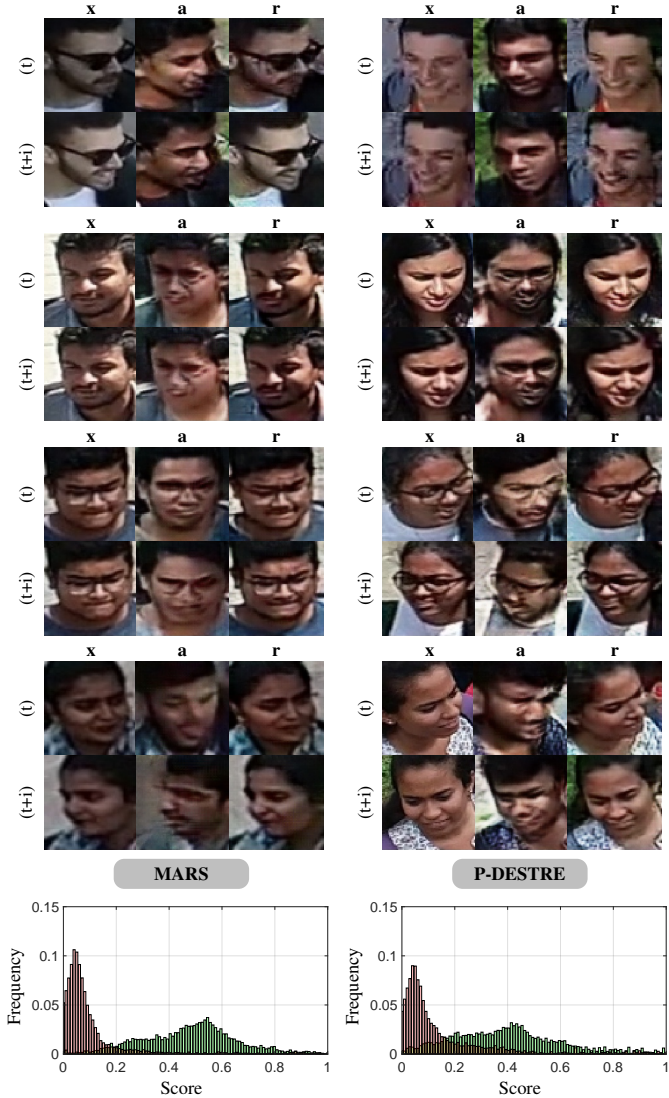
\begin{figure}[ht!]
\begin{center}
\begin{tikzpicture}

\def\sizeImg{1.25}

%%%%%%%%%%%%%%%%%%%%%%%%%%%%%%%%%%%%%%%%%%

\node at (0, 0.75) [black, rotate=0]  {\scriptsize{\textbf{x}}};   
\draw (0,0) node(n1)  {\includegraphics[height=\sizeImg cm]{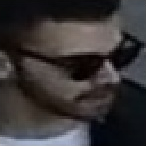}};   
\node at (0.97*\sizeImg, 0.75) [black, rotate=0]  {\scriptsize{\textbf{a}}};   
\draw (0.97*\sizeImg,0) node(n1)  {\includegraphics[height=\sizeImg cm]{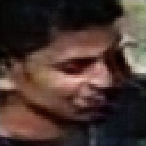}};   
\node at (1.97*\sizeImg, 0.75) [black, rotate=0]  {\scriptsize{\textbf{r}}};   
\draw (1.97*\sizeImg,0) node(n1)  {\includegraphics[height=\sizeImg cm]{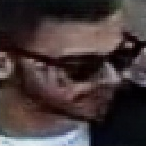}};   

\draw (0,0-1.0*\sizeImg) node(n1)  {\includegraphics[height=\sizeImg cm]{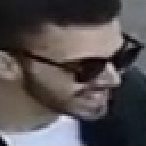}};   
\draw (0.985*\sizeImg,0-1.0*\sizeImg) node(n1)  {\includegraphics[height=\sizeImg cm]{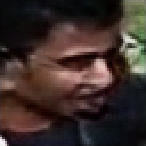}};   
\draw (1.97*\sizeImg,0-1.0*\sizeImg) node(n1)  {\includegraphics[height=\sizeImg cm]{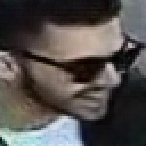}};

\node at (-0.85, 0) [black, rotate=90]  {\scriptsize{(t)}};   
\node at (-0.85, 0-\sizeImg) [black, rotate=90]  {\scriptsize{(t+i)}};

%%%%%%%%%%%%%%%%%%%%%%%%%%%%%%%%%%%%%%%%%%

\def\deltaX{4.5}
\def\deltaY{0}

\node at (0+\deltaX, 0.75+\deltaY) [black, rotate=0]  {\scriptsize{\textbf{x}}};   
\draw (0+\deltaX,0+\deltaY) node(n1)  {\includegraphics[height=\sizeImg cm]{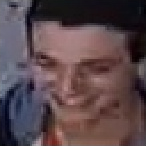}};   
\node at (0.97*\sizeImg+\deltaX, 0.75+\deltaY) [black, rotate=0]  {\scriptsize{\textbf{a}}};   
\draw (0.97*\sizeImg+\deltaX,0+\deltaY) node(n1)  {\includegraphics[height=\sizeImg cm]{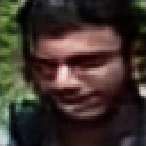}};   
\node at (1.97*\sizeImg+\deltaX, 0.75+\deltaY) [black, rotate=0]  {\scriptsize{\textbf{r}}};   
\draw (1.97*\sizeImg+\deltaX,0+\deltaY) node(n1)  {\includegraphics[height=\sizeImg cm]{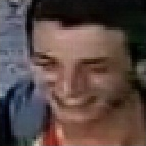}};   

\draw (0+\deltaX,0-1.0*\sizeImg+\deltaY) node(n1)  {\includegraphics[height=\sizeImg cm]{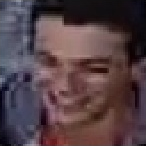}};   
\draw (0.985*\sizeImg+\deltaX,0-1.0*\sizeImg+\deltaY) node(n1)  {\includegraphics[height=\sizeImg cm]{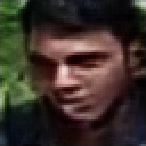}};   
\draw (1.97*\sizeImg+\deltaX,0-1.0*\sizeImg+\deltaY) node(n1)  {\includegraphics[height=\sizeImg cm]{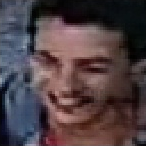}};

\node at (-0.85+\deltaX, 0+\deltaY) [black, rotate=90]  {\scriptsize{(t)}};   
\node at (-0.85+\deltaX, 0-\sizeImg+\deltaY) [black, rotate=90]  {\scriptsize{(t+i)}};   

%%%%%%%%%%%%%%%%%%%%%%%%%%%%%%%%%%%%%%%%%%

\def\deltaX{0}
\def\deltaY{-2.85}

\node at (0+\deltaX, 0.75+\deltaY) [black, rotate=0]  {\scriptsize{\textbf{x}}};   
\draw (0+\deltaX,0+\deltaY) node(n1)  {\includegraphics[height=\sizeImg cm]{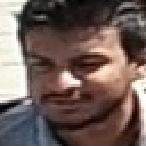}};   
\node at (0.97*\sizeImg+\deltaX, 0.75+\deltaY) [black, rotate=0]  {\scriptsize{\textbf{a}}};   
\draw (0.97*\sizeImg+\deltaX,0+\deltaY) node(n1)  {\includegraphics[height=\sizeImg cm]{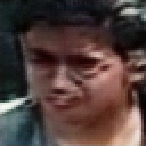}};   
\node at (1.97*\sizeImg+\deltaX, 0.75+\deltaY) [black, rotate=0]  {\scriptsize{\textbf{r}}};   
\draw (1.97*\sizeImg+\deltaX,0+\deltaY) node(n1)  {\includegraphics[height=\sizeImg cm]{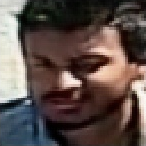}};   

\draw (0+\deltaX,0-1.0*\sizeImg+\deltaY) node(n1)  {\includegraphics[height=\sizeImg cm]{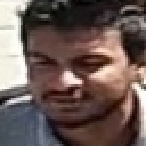}};   
\draw (0.985*\sizeImg+\deltaX,0-1.0*\sizeImg+\deltaY) node(n1)  {\includegraphics[height=\sizeImg cm]{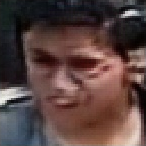}};   
\draw (1.97*\sizeImg+\deltaX,0-1.0*\sizeImg+\deltaY) node(n1)  {\includegraphics[height=\sizeImg cm]{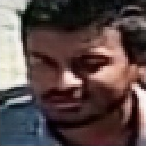}};   

\node at (-0.85+\deltaX, 0+\deltaY) [black, rotate=90]  {\scriptsize{(t)}};   
\node at (-0.85+\deltaX, 0-\sizeImg+\deltaY) [black, rotate=90]  {\scriptsize{(t+i)}};   

%%%%%%%%%%%%%%%%%%%%%%%%%%%%%%%%%%%%%%%%%%

\def\deltaX{4.5}
\def\deltaY{-2.85}

\node at (0+\deltaX, 0.75+\deltaY) [black, rotate=0]  {\scriptsize{\textbf{x}}};   
\draw (0+\deltaX,0+\deltaY) node(n1)  {\includegraphics[height=\sizeImg cm]{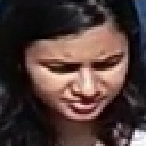}};   
\node at (0.97*\sizeImg+\deltaX, 0.75+\deltaY) [black, rotate=0]  {\scriptsize{\textbf{a}}};   
\draw (0.97*\sizeImg+\deltaX,0+\deltaY) node(n1)  {\includegraphics[height=\sizeImg cm]{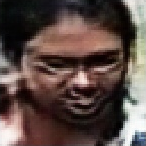}};   
\node at (1.97*\sizeImg+\deltaX, 0.75+\deltaY) [black, rotate=0]  {\scriptsize{\textbf{r}}};   
\draw (1.97*\sizeImg+\deltaX,0+\deltaY) node(n1)  {\includegraphics[height=\sizeImg cm]{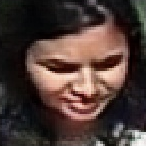}};   

\draw (0+\deltaX,0-1.0*\sizeImg+\deltaY) node(n1)  {\includegraphics[height=\sizeImg cm]{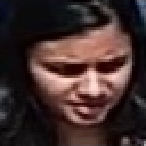}};   
\draw (0.985*\sizeImg+\deltaX,0-1.0*\sizeImg+\deltaY) node(n1)  {\includegraphics[height=\sizeImg cm]{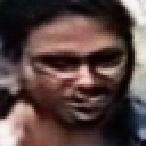}};   
\draw (1.97*\sizeImg+\deltaX,0-1.0*\sizeImg+\deltaY) node(n1)  {\includegraphics[height=\sizeImg cm]{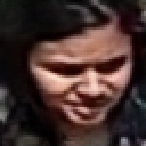}};

\node at (-0.85+\deltaX, 0+\deltaY) [black, rotate=90]  {\scriptsize{(t)}};   
\node at (-0.85+\deltaX, 0-\sizeImg+\deltaY) [black, rotate=90]  {\scriptsize{(t+i)}};   
%%%%%%%%%%%%%%%%%%%%%%%%%%%%%%%%%%%%%%%%%%

\def\deltaX{0}
\def\deltaY{-5.7}

\node at (0+\deltaX, 0.75+\deltaY) [black, rotate=0]  {\scriptsize{\textbf{x}}};   
\draw (0+\deltaX,0+\deltaY) node(n1)  {\includegraphics[height=\sizeImg cm]{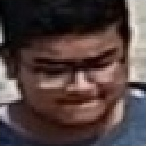}};   
\node at (0.97*\sizeImg+\deltaX, 0.75+\deltaY) [black, rotate=0]  {\scriptsize{\textbf{a}}};   
\draw (0.97*\sizeImg+\deltaX,0+\deltaY) node(n1)  {\includegraphics[height=\sizeImg cm]{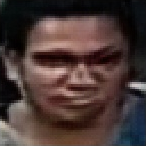}};   
\node at (1.97*\sizeImg+\deltaX, 0.75+\deltaY) [black, rotate=0]  {\scriptsize{\textbf{r}}};   
\draw (1.97*\sizeImg+\deltaX,0+\deltaY) node(n1)  {\includegraphics[height=\sizeImg cm]{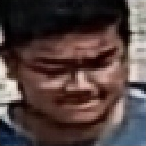}};   

\draw (0+\deltaX,0-1.0*\sizeImg+\deltaY) node(n1)  {\includegraphics[height=\sizeImg cm]{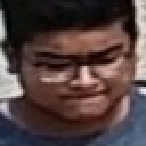}};   
\draw (0.985*\sizeImg+\deltaX,0-1.0*\sizeImg+\deltaY) node(n1)  {\includegraphics[height=\sizeImg cm]{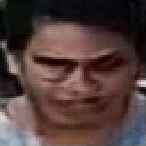}};   
\draw (1.97*\sizeImg+\deltaX,0-1.0*\sizeImg+\deltaY) node(n1)  {\includegraphics[height=\sizeImg cm]{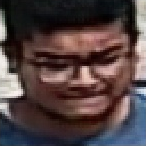}};   

\node at (-0.85+\deltaX, 0+\deltaY) [black, rotate=90]  {\scriptsize{(t)}};   
\node at (-0.85+\deltaX, 0-\sizeImg+\deltaY) [black, rotate=90]  {\scriptsize{(t+i)}};   

%%%%%%%%%%%%%%%%%%%%%%%%%%%%%%%%%%%%%%%%%%

\def\deltaX{4.5}
\def\deltaY{-5.7}

\node at (0+\deltaX, 0.75+\deltaY) [black, rotate=0]  {\scriptsize{\textbf{x}}};   
\draw (0+\deltaX,0+\deltaY) node(n1)  {\includegraphics[height=\sizeImg cm]{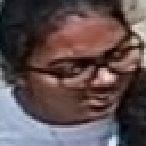}};   
\node at (0.97*\sizeImg+\deltaX, 0.75+\deltaY) [black, rotate=0]  {\scriptsize{\textbf{a}}};   
\draw (0.97*\sizeImg+\deltaX,0+\deltaY) node(n1)  {\includegraphics[height=\sizeImg cm]{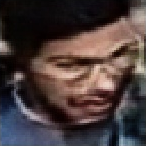}};   
\node at (1.97*\sizeImg+\deltaX, 0.75+\deltaY) [black, rotate=0]  {\scriptsize{\textbf{r}}};   
\draw (1.97*\sizeImg+\deltaX,0+\deltaY) node(n1)  {\includegraphics[height=\sizeImg cm]{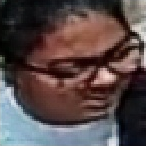}};   

\draw (0+\deltaX,0-1.0*\sizeImg+\deltaY) node(n1)  {\includegraphics[height=\sizeImg cm]{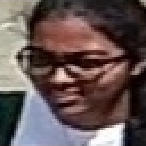}};   
\draw (0.985*\sizeImg+\deltaX,0-1.0*\sizeImg+\deltaY) node(n1)  {\includegraphics[height=\sizeImg cm]{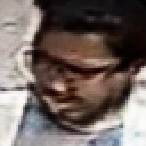}};   
\draw (1.97*\sizeImg+\deltaX,0-1.0*\sizeImg+\deltaY) node(n1)  {\includegraphics[height=\sizeImg cm]{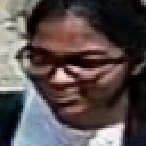}};   

\node at (-0.85+\deltaX, 0+\deltaY) [black, rotate=90]  {\scriptsize{(t)}};   
\node at (-0.85+\deltaX, 0-\sizeImg+\deltaY) [black, rotate=90]  {\scriptsize{(t+i)}};

%%%%%%%%%%%%%%%%%%%%%%%%%%%%%%%%%%%%%%%%%%
\def\deltaX{4.5}
\def\deltaY{-8.55}

\node at (0+\deltaX, 0.75+\deltaY) [black, rotate=0]  {\scriptsize{\textbf{x}}};   
\draw (0+\deltaX,0+\deltaY) node(n1)  {\includegraphics[height=\sizeImg cm]{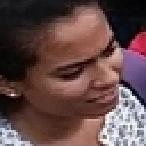}};   
\node at (0.97*\sizeImg+\deltaX, 0.75+\deltaY) [black, rotate=0]  {\scriptsize{\textbf{a}}};   
\draw (0.97*\sizeImg+\deltaX,0+\deltaY) node(n1)  {\includegraphics[height=\sizeImg cm]{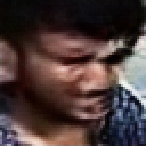}};   
\node at (1.97*\sizeImg+\deltaX, 0.75+\deltaY) [black, rotate=0]  {\scriptsize{\textbf{r}}};   
\draw (1.97*\sizeImg+\deltaX,0+\deltaY) node(n1)  {\includegraphics[height=\sizeImg cm]{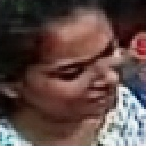}};   

\draw (0+\deltaX,0-1.0*\sizeImg+\deltaY) node(n1)  {\includegraphics[height=\sizeImg cm]{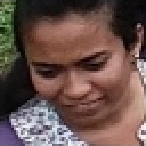}};   
\draw (0.985*\sizeImg+\deltaX,0-1.0*\sizeImg+\deltaY) node(n1)  {\includegraphics[height=\sizeImg cm]{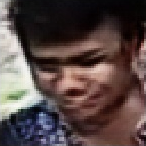}};   
\draw (1.97*\sizeImg+\deltaX,0-1.0*\sizeImg+\deltaY) node(n1)  {\includegraphics[height=\sizeImg cm]{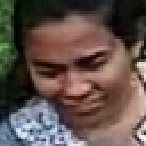}};   

\node at (-0.85+\deltaX, 0+\deltaY) [black, rotate=90]  {\scriptsize{(t)}};   
\node at (-0.85+\deltaX, 0-\sizeImg+\deltaY) [black, rotate=90]  {\scriptsize{(t+i)}};   
%%%%%%%%%%%%%%%%%%%%%%%%%%%%%%%%%%%%%%%%%%

\def\deltaX{0}
\def\deltaY{-8.55}

\node at (0+\deltaX, 0.75+\deltaY) [black, rotate=0]  {\scriptsize{\textbf{x}}};   
\draw (0+\deltaX,0+\deltaY) node(n1)  {\includegraphics[height=\sizeImg cm]{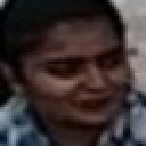}};   
\node at (0.97*\sizeImg+\deltaX, 0.75+\deltaY) [black, rotate=0]  {\scriptsize{\textbf{a}}};   
\draw (0.97*\sizeImg+\deltaX,0+\deltaY) node(n1)  {\includegraphics[height=\sizeImg cm]{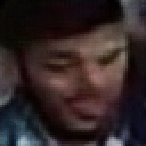}};   
\node at (1.97*\sizeImg+\deltaX, 0.75+\deltaY) [black, rotate=0]  {\scriptsize{\textbf{r}}};   
\draw (1.97*\sizeImg+\deltaX,0+\deltaY) node(n1)  {\includegraphics[height=\sizeImg cm]{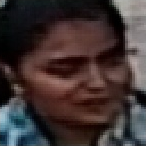}};   

\draw (0+\deltaX,0-1.0*\sizeImg+\deltaY) node(n1)  {\includegraphics[height=\sizeImg cm]{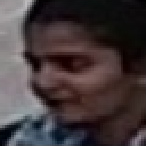}};   
\draw (0.985*\sizeImg+\deltaX,0-1.0*\sizeImg+\deltaY) node(n1)  {\includegraphics[height=\sizeImg cm]{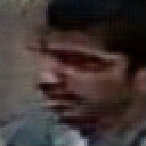}};   
\draw (1.97*\sizeImg+\deltaX,0-1.0*\sizeImg+\deltaY) node(n1)  {\includegraphics[height=\sizeImg cm]{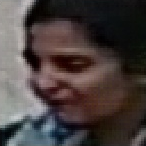}};   

\node at (-0.85+\deltaX, 0+\deltaY) [black, rotate=90]  {\scriptsize{(t)}};   
\node at (-0.85+\deltaX, 0-\sizeImg+\deltaY) [black, rotate=90]  {\scriptsize{(t+i)}};   
%%%%%%%%%%%%%%%%%%%%%%%%%%%%%%%%%%%%%%%%%%

\def\deltaY{-12.25}
\def\deltaX{1.2}

\fill [gray!50, rounded corners] (-0.9+\deltaX, 1.3+\deltaY) rectangle (0.9+\deltaX, 1.7+\deltaY);     
\node at (0+\deltaX, 1.5+\deltaY) [black, rotate=0]  {\scriptsize{\textbf{MARS}}};

\draw (0+\deltaX,0+\deltaY) node(n1)  {\includegraphics[width=4 cm]{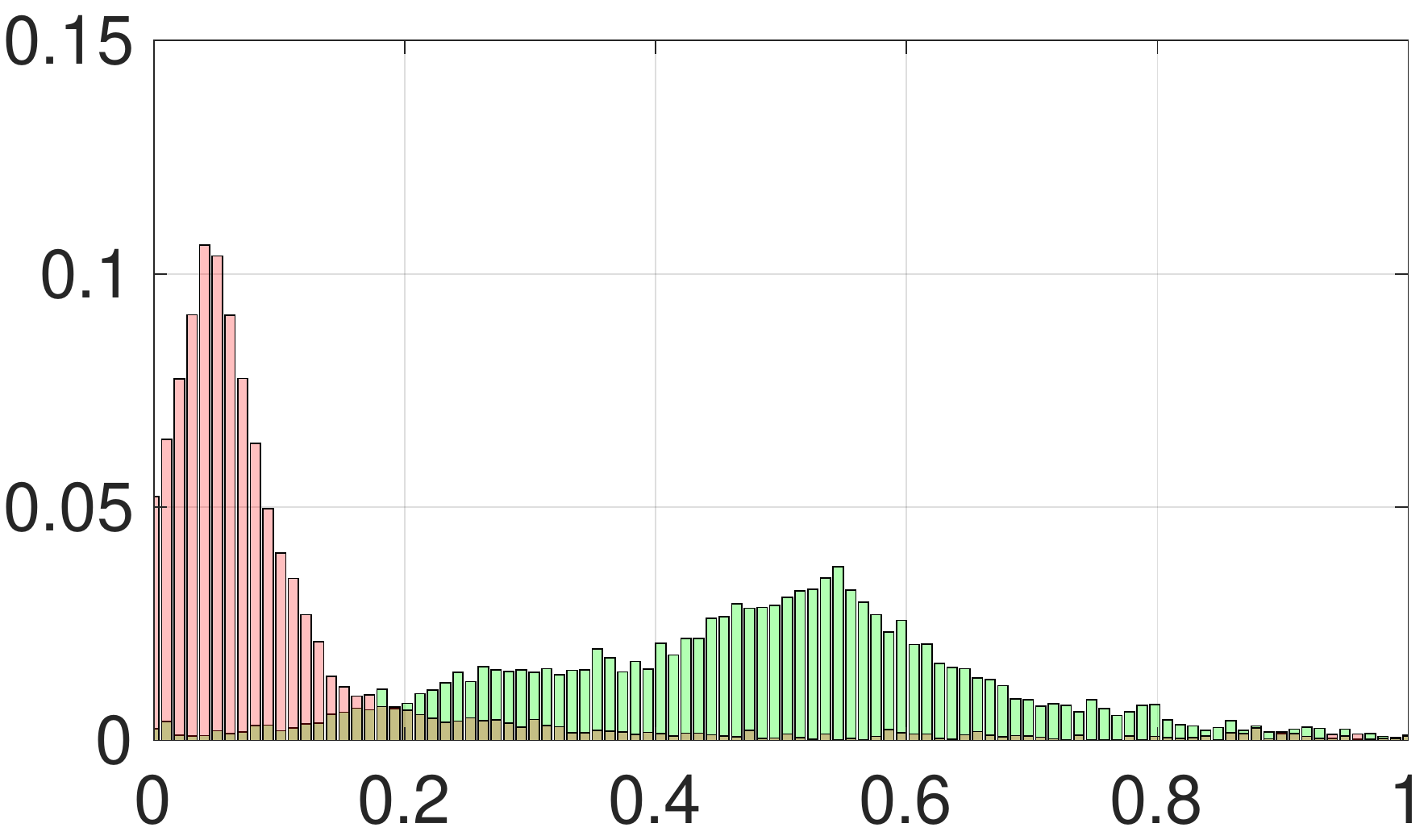}};   
\node at (0+\deltaX, -1.35+\deltaY) [black, rotate=0]  {\scriptsize{Score}};     
\node at (-2.2+\deltaX, 0+\deltaY) [black, rotate=90]  {\scriptsize{Frequency}};

\fill [gray!50, rounded corners] (-0.9+4.5+\deltaX, 1.3+\deltaY) rectangle (0.9+4.5+\deltaX, 1.7+\deltaY);     
\node at (0+4.5+\deltaX, 1.5+\deltaY) [black, rotate=0]  {\scriptsize{\textbf{P-DESTRE}}};     

\draw (4.5+\deltaX,0+\deltaY) node(n1)  {\includegraphics[width=4 cm]{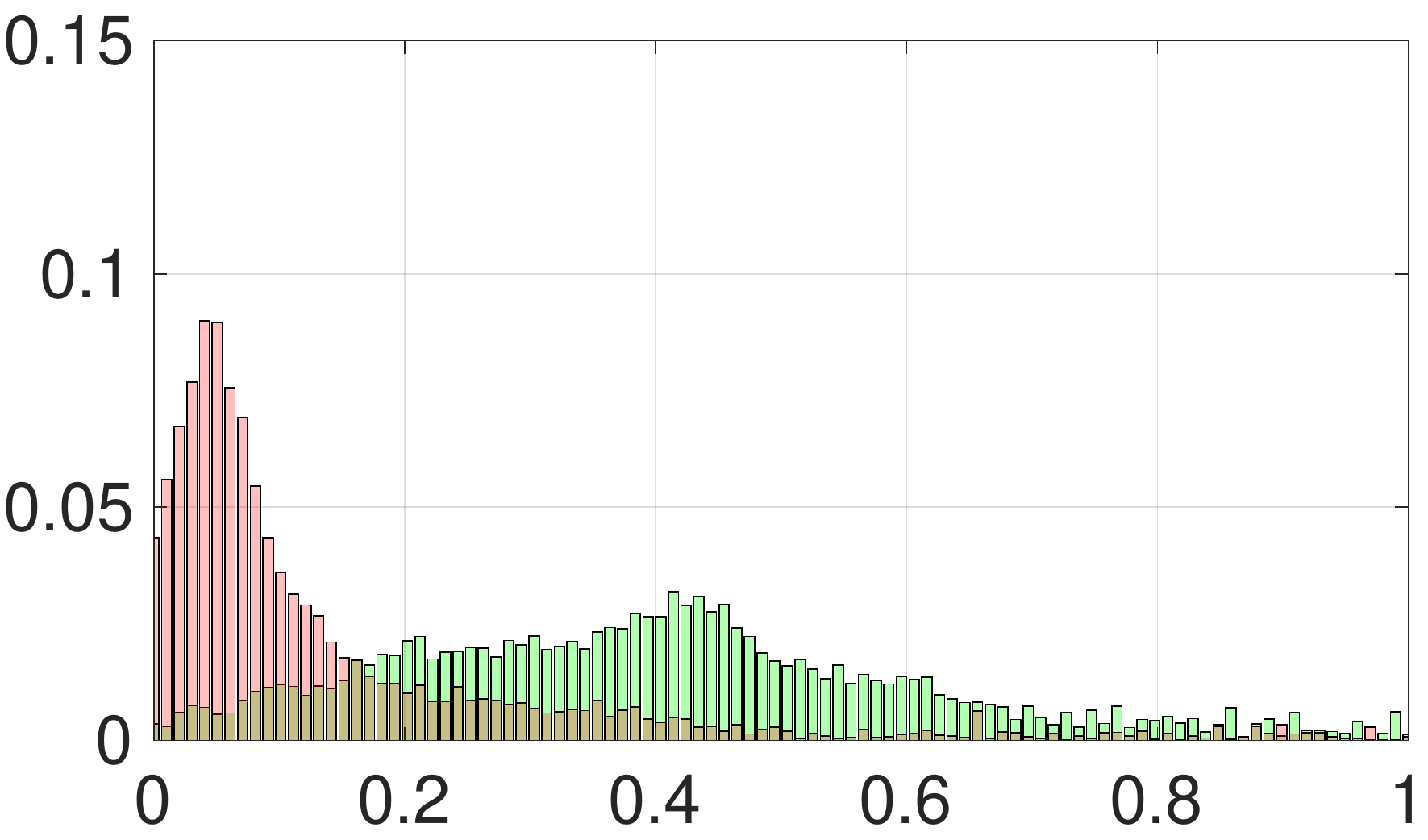}};   
\node at (0+4.5+\deltaX, -1.35+\deltaY) [black, rotate=0]  {\scriptsize{Score}};     
\node at (-2.2+4.5+\deltaX, 0+\deltaY) [black, rotate=90]  {\scriptsize{Frequency}};

\end{tikzpicture}
    \caption{Top rows: Examples that illustrate the temporal consistency required for video de-identification. Each group provides two within-subject examples of a sequence \textbf{x} taken at time $t$ and $t+i, i \geq 1$. The central column in each group provides the de-identified images \textbf{a}, and the reconstructed samples are given at the rightmost column \textbf{r}. The bottom row shows the decision environments obtained when the \emph{genuine} pairwise comparisons were exclusively composed of  \textbf{a}$_{i, j, t}$/\textbf{a}$_{i, j, t+k}$ elements.} 
        \label{fig:temporalConsistency}
    \end{center}
\end{figure}

The temporal consistency of the de-identified samples is of most importance for photo-realism purposes. In a subjective evaluation perspective, Fig.~\ref{fig:temporalConsistency} provides several examples of (\textbf{x}, \textbf{a}, \textbf{r}) elements, obtained for different frames of the same sequence at time $t$ and $t+i$. In all cases, an evident consistency between the features generated for \textbf{a}$_t$ and  \textbf{a}$_{t+i}$ can be observed.

To provide a quantitative measure of temporal consistency, we compared the decision environments for the MARS and P-DESTRE sets when the \emph{genuine} pairwise comparisons were exclusively composed of \textbf{a}$_{i, j, t}$/\textbf{a}$_{i, j, t+k}$ elements, with $k \in \{1,\ldots, s_l\}$ ($s_l$ is the sequence length). The \emph{impostors} distributions were obtained as in the previous experiments and contextualize the genuine scores. The bottom row in Fig.~\ref{fig:temporalConsistency} provides the results for both sets,  where the separation between the \emph{impostors}' and the \emph{genuine} scores is evident. The Kolmogorov-Smirnov test was used to compare both empirical data distributions, with the \emph{null} hypothesis ('\emph{both samples come from the same distribution}') being rejected with asymptotic \emph{p}-values lower than 1$e^{-8}$ in MARS and P-DESTRE.  When comparing these results to the values given in Figs.~\ref{fig:decisionEnv}, note that in the later case, only samples of the same session were considered as \emph{genuine} pairs, which justifies the larger separability between the \emph{genuine}/\emph{impostors} distributions. In both cases, the genuine scores spread in a relatively homogeneous way in the unit interval, yet there is still a fraction of cases ($< 15\%$) where consecutive de-identified elements of one subject suddenly  change their appearance and even soft labels.

%\begin{figure}[ht!]
%\begin{center}
%\begin{tikzpicture}
%
%\def\deltaY{0}
%
%
%\fill [gray!50, rounded corners] (-0.9, 1.3+\deltaY) rectangle (0.9, 1.7+\deltaY);     
%\node at (0, 1.5) [black, rotate=0]  {\scriptsize{\textbf{MARS}}};     
%
%
%\draw (0,0) node(n1)  {\includegraphics[width=4 cm]{imgs/temporal_consistency_MARS}};   
%\node at (0, -1.35) [black, rotate=0]  {\scriptsize{Score}};     
%\node at (-2.2, 0) [black, rotate=90]  {\scriptsize{Frequency}};     
%
%
%\fill [gray!50, rounded corners] (-0.9+4.5, 1.3+\deltaY) rectangle (0.9+4.5, 1.7+\deltaY);     
%\node at (0+4.5, 1.5) [black, rotate=0]  {\scriptsize{\textbf{P-DESTRE}}};     
%
%\draw (4.5,0) node(n1)  {\includegraphics[width=4 cm]{imgs/temporal_consistency_PDESTRE}};   
%\node at (0+4.5, -1.35) [black, rotate=0]  {\scriptsize{Score}};     
%\node at (-2.2+4.5, 0) [black, rotate=90]  {\scriptsize{Frequency}};     
%
%
%\end{tikzpicture}
%    \caption{Temporal consistency experiments: decision environments obtained when the test sets of \emph{genuine} pairwise comparisons were exclusively composed of  \textbf{a}$_{i, j, t}$/\textbf{a}$_{i, j, t+k}$ elements, i.e., the de-identified samples for the frames that compose a subject sequence.}
%        \label{fig:temporal_results}
%    \end{center}
%\end{figure}

\subsection{Soft Labels Consistency/Inter-Session Diversity}

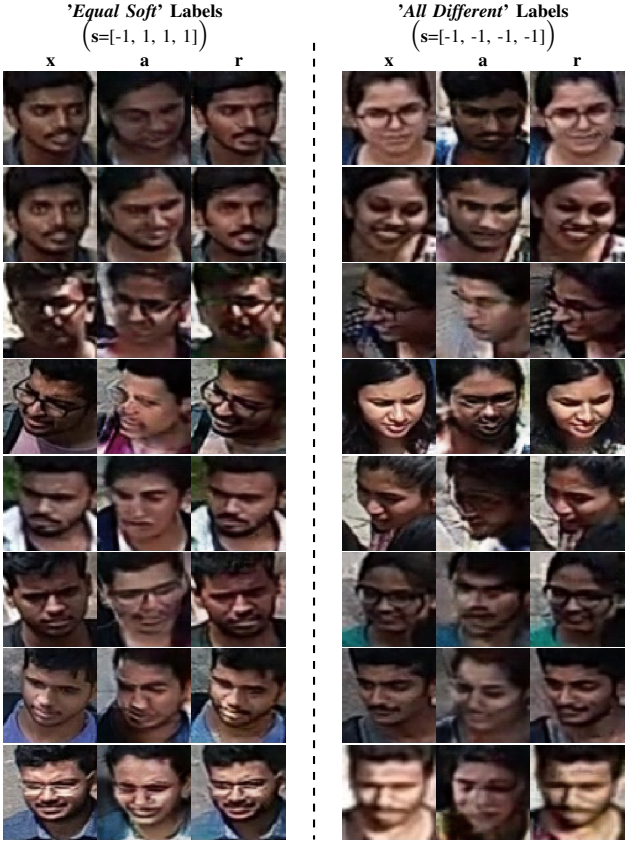
\begin{figure}[ht!]
\begin{center}
\begin{tikzpicture}

\def\sizeImg{1.25}

%%%%%%%%%%%%%%%%%%%%%%%%%%%%%%%%%%%%%%%%%%

\node at (1*\sizeImg, 1.4) [black, rotate=0]  {\scriptsize{\textbf{'\emph{Equal Soft}' Labels}}};   
\node at (1*\sizeImg, 1.1) [black, rotate=0]  {\scriptsize{\big(\textbf{s}=[-1, 1, 1, 1]\big)}};   

\node at (0, 0.75) [black, rotate=0]  {\scriptsize{\textbf{x}}};   
\draw (0,0) node(n1)  {\includegraphics[height=\sizeImg cm]{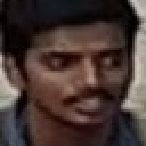}};   
\node at (1*\sizeImg, 0.75) [black, rotate=0]  {\scriptsize{\textbf{a}}};   
\draw (1*\sizeImg,0) node(n1)  {\includegraphics[height=\sizeImg cm]{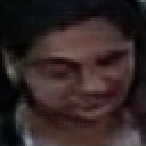}};   
\node at (2*\sizeImg, 0.75) [black, rotate=0]  {\scriptsize{\textbf{r}}};   
\draw (2*\sizeImg,0) node(n1)  {\includegraphics[height=\sizeImg cm]{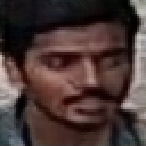}};   

\draw (0,0-1.025*\sizeImg) node(n1)  {\includegraphics[height=\sizeImg cm]{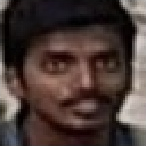}};   
\draw (1*\sizeImg,0-1.025*\sizeImg) node(n1)  {\includegraphics[height=\sizeImg cm]{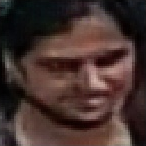}};   
\draw (2*\sizeImg,0-1.025*\sizeImg) node(n1)  {\includegraphics[height=\sizeImg cm]{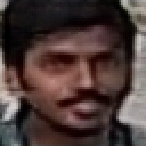}};   

\draw (0,0-2.05*\sizeImg) node(n1)  {\includegraphics[height=\sizeImg cm]{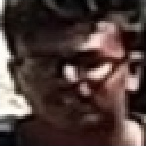}};   
\draw (1*\sizeImg,0-2.05*\sizeImg) node(n1)  {\includegraphics[height=\sizeImg cm]{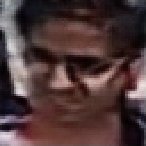}};   
\draw (2*\sizeImg,0-2.05*\sizeImg) node(n1)  {\includegraphics[height=\sizeImg cm]{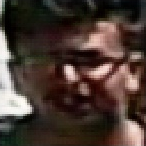}};   

\draw (0,0-3.075*\sizeImg) node(n1)  {\includegraphics[height=\sizeImg cm]{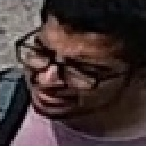}};   
\draw (1*\sizeImg,0-3.075*\sizeImg) node(n1)  {\includegraphics[height=\sizeImg cm]{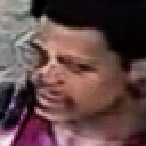}};   
\draw (2*\sizeImg,0-3.075*\sizeImg) node(n1)  {\includegraphics[height=\sizeImg cm]{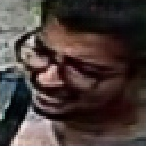}};   

\draw (0,0-4.1*\sizeImg) node(n1)  {\includegraphics[height=\sizeImg cm]{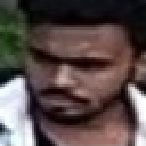}};   
\draw (1*\sizeImg,0-4.1*\sizeImg) node(n1)  {\includegraphics[height=\sizeImg cm]{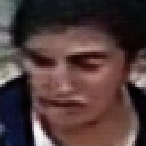}};   
\draw (2*\sizeImg,0-4.1*\sizeImg) node(n1)  {\includegraphics[height=\sizeImg cm]{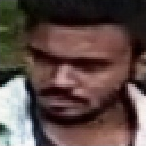}};   

\draw (0,0-5.125*\sizeImg) node(n1)  {\includegraphics[height=\sizeImg cm]{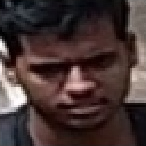}};   
\draw (1*\sizeImg,0-5.125*\sizeImg) node(n1)  {\includegraphics[height=\sizeImg cm]{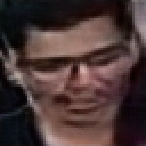}};   
\draw (2*\sizeImg,0-5.125*\sizeImg) node(n1)  {\includegraphics[height=\sizeImg cm]{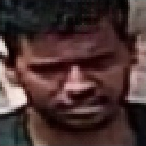}};   

\draw (0,0-6.15*\sizeImg) node(n1)  {\includegraphics[height=\sizeImg cm]{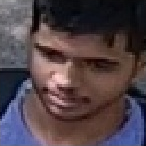}};   
\draw (1*\sizeImg,0-6.15*\sizeImg) node(n1)  {\includegraphics[height=\sizeImg cm]{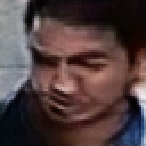}};   
\draw (2*\sizeImg,0-6.15*\sizeImg) node(n1)  {\includegraphics[height=\sizeImg cm]{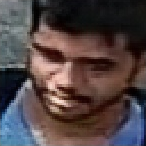}};   

\draw (0,0-7.175*\sizeImg) node(n1)  {\includegraphics[height=\sizeImg cm]{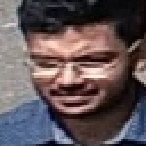}};   
\draw (1*\sizeImg,0-7.175*\sizeImg) node(n1)  {\includegraphics[height=\sizeImg cm]{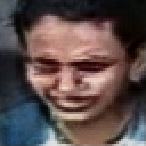}};   
\draw (2*\sizeImg,0-7.175*\sizeImg) node(n1)  {\includegraphics[height=\sizeImg cm]{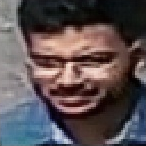}};   

\draw [dashed, thick] (3.5,1) -- (3.5,-9.6);

\def\deltaX{4.5}
\def\deltaY{0}
\node at (1*\sizeImg+\deltaX, 1.4) [black, rotate=0]  {\scriptsize{\textbf{'\emph{All Different}' Labels}}};   
\node at (1*\sizeImg+\deltaX, 1.1) [black, rotate=0]  {\scriptsize{\big(\textbf{s}=[-1, -1, -1, -1]\big)}};  

\node at (0+\deltaX, 0.75) [black, rotate=0]  {\scriptsize{\textbf{x}}};   
\node at (1*\sizeImg+\deltaX, 0.75) [black, rotate=0]  {\scriptsize{\textbf{a}}};   
\node at (2*\sizeImg+\deltaX, 0.75) [black, rotate=0]  {\scriptsize{\textbf{r}}};   

\draw (0+\deltaX,0+\deltaY) node(n1)  {\includegraphics[height=\sizeImg cm]{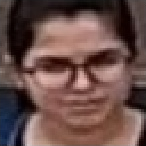}};   
\draw (1*\sizeImg+\deltaX,0+\deltaY) node(n1)  {\includegraphics[height=\sizeImg cm]{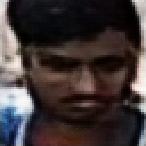}};   
\draw (2*\sizeImg+\deltaX,0+\deltaY) node(n1)  {\includegraphics[height=\sizeImg cm]{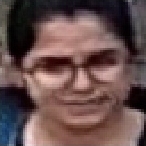}};   

\draw (0+\deltaX,0-1.025*\sizeImg+\deltaY) node(n1)  {\includegraphics[height=\sizeImg cm]{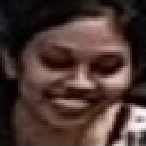}};   
\draw (1*\sizeImg+\deltaX,0-1.025*\sizeImg+\deltaY) node(n1)  {\includegraphics[height=\sizeImg cm]{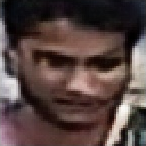}};   
\draw (2*\sizeImg+\deltaX,0-1.025*\sizeImg+\deltaY) node(n1)  {\includegraphics[height=\sizeImg cm]{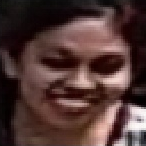}};   

\draw (0+\deltaX,0-2.05*\sizeImg+\deltaY) node(n1)  {\includegraphics[height=\sizeImg cm]{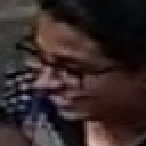}};   
\draw (1*\sizeImg+\deltaX,0-2.05*\sizeImg+\deltaY) node(n1)  {\includegraphics[height=\sizeImg cm]{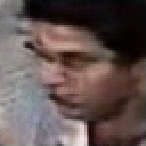}};   
\draw (2*\sizeImg+\deltaX,0-2.05*\sizeImg+\deltaY) node(n1)  {\includegraphics[height=\sizeImg cm]{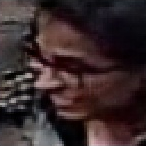}};   

\draw (0+\deltaX,0-3.075*\sizeImg+\deltaY) node(n1)  {\includegraphics[height=\sizeImg cm]{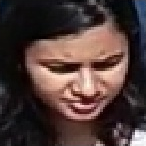}};   
\draw (1*\sizeImg+\deltaX,0-3.075*\sizeImg+\deltaY) node(n1)  {\includegraphics[height=\sizeImg cm]{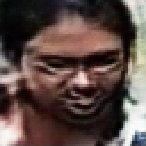}};   
\draw (2*\sizeImg+\deltaX,0-3.075*\sizeImg+\deltaY) node(n1)  {\includegraphics[height=\sizeImg cm]{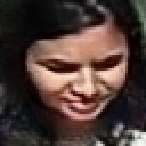}};   

\draw (0+\deltaX,0-4.1*\sizeImg+\deltaY) node(n1)  {\includegraphics[height=\sizeImg cm]{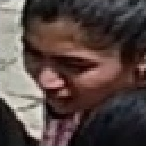}};   
\draw (1*\sizeImg+\deltaX,0-4.1*\sizeImg+\deltaY) node(n1)  {\includegraphics[height=\sizeImg cm]{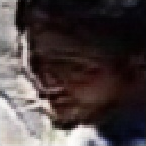}};   
\draw (2*\sizeImg+\deltaX,0-4.1*\sizeImg+\deltaY) node(n1)  {\includegraphics[height=\sizeImg cm]{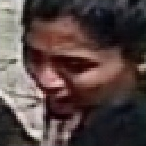}};   

\draw (0+\deltaX,0-5.125*\sizeImg+\deltaY) node(n1)  {\includegraphics[height=\sizeImg cm]{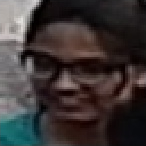}};   
\draw (1*\sizeImg+\deltaX,0-5.125*\sizeImg+\deltaY) node(n1)  {\includegraphics[height=\sizeImg cm]{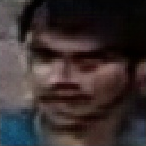}};   
\draw (2*\sizeImg+\deltaX,0-5.125*\sizeImg+\deltaY) node(n1)  {\includegraphics[height=\sizeImg cm]{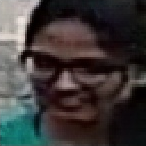}};   

\draw (0+\deltaX,0-6.150*\sizeImg+\deltaY) node(n1)  {\includegraphics[height=\sizeImg cm]{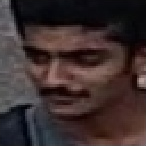}};   
\draw (1*\sizeImg+\deltaX,0-6.150*\sizeImg+\deltaY) node(n1)  {\includegraphics[height=\sizeImg cm]{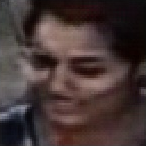}};   
\draw (2*\sizeImg+\deltaX,0-6.150*\sizeImg+\deltaY) node(n1)  {\includegraphics[height=\sizeImg cm]{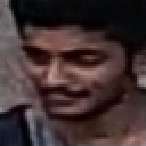}};   

\draw (0+\deltaX,0-7.175*\sizeImg+\deltaY) node(n1)  {\includegraphics[height=\sizeImg cm]{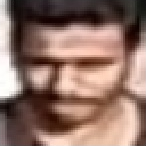}};   
\draw (1*\sizeImg+\deltaX,0-7.175*\sizeImg+\deltaY) node(n1)  {\includegraphics[height=\sizeImg cm]{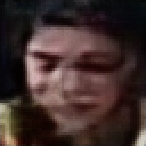}};   
\draw (2*\sizeImg+\deltaX,0-7.175*\sizeImg+\deltaY) node(n1)  {\includegraphics[height=\sizeImg cm]{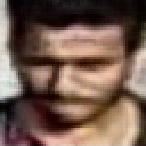}};   

%%%%%%%%%%%%%%%%%%%%%%%%%%%%%%%%%%%%%%%%%%

\end{tikzpicture}
    \caption{Comparison between the de-identification results that are typically attained when the soft labels coherence/discrepancy is enforced. The left column provides some examples where the soft labels configuration between \textbf{x}$_.$/\textbf{a}$_.$ is kept consistent, while the right column illustrates the  results when all soft labels disagree.} 
        \label{fig:labelsConsistency}
    \end{center}
\end{figure}

The consistency of the soft labels generated for the de-identified data and the diversity of the virtual IDs generated per subject were evaluated according to the responses provided by the pairwise labels discriminator \textbf{D}$_a$. For the soft labels, we were interested in confirming if the \{'gender', 'ethnicity', 'hairstyle'\} labels inferred for \textbf{a}$_.$ meet the constraints determined by $\textbf{s}$ in the learning phase. For each \textbf{a}$_i$ element, we obtained the $\textbf{D}_a(\textbf{a}_i, \textbf{x}_i)$ values and measured their Pearson correlation with respect to the ground-truth labels of \textbf{x}, also considering the configuration of $\textbf{s}$. Having drew 50 random samples composed of 90\% of the test samples (with repetition), the linear correlation values are given in Table~\ref{tab:correlation}, which were regarded as good indicators of the soft labels consistency of  \textbf{a}$_.$ elements. Some examples of the \textbf{a}$_.$ elements generated when $\textbf{s}=[\text{'ID'}, \text{'Gender'},\text{'Ethnicity'}, \text{'Hairstyle'}]=[-1,1,1,1]$ ('\emph{Equal Soft}' labels) and $\textbf{s}=[-1,-1,-1,-1]$ ('\emph{All Different}' labels) are shown in Fig.~\ref{fig:labelsConsistency}, enabling to perceive the evidently different features of the \textbf{a} elements according to the configuration used for $\textbf{s}$ in the learning phase. Also, we observed that using the '\emph{Equal Soft}' labels configuration reduces the variability of the synthesised virtual IDs, while also increasing the similarity in appearance between the \textbf{x}$_.$/\textbf{a}$_.$ elements.

\begin{table}[h!]
\centering
     \caption{Pearson correlation between the soft labels inferred for the de-identified elements and the desired properties determined by \textbf{s}.}
     \label{tab:correlation}
\begin{tabular}{|l|C{1.5cm}|C{1.5cm}|C{1.5cm}|}
\hline
\textbf{\scriptsize{Soft Label Consistency}}   & \textbf{\scriptsize{BIODI}} & \textbf{\scriptsize{MARS}}  & \textbf{\scriptsize{P-DESTRE}}\\ \hline

\scriptsize{Gender}  & \scriptsize{0.818} \tiny{$\pm$ 0.096} & \scriptsize{0.890} \tiny{$\pm$ 0.081} & \scriptsize{0.803} \tiny{$\pm$ 0.107} \\ \hline

\scriptsize{Ethnicity}  & \scriptsize{0.702} \tiny{$\pm$ 0.112} & \scriptsize{0.750} \tiny{$\pm$ 0.099} & \scriptsize{0.622} \tiny{$\pm$ 0.144} \\ \hline

\scriptsize{Hairstyle}  & \scriptsize{0.647} \tiny{$\pm$ 0.106} & \scriptsize{0.663} \tiny{$\pm$ 0.102} & \scriptsize{0.594} \tiny{$\pm$ 0.118} \\ \hline

\end{tabular}
\end{table}

\begin{figure}[ht!]
\begin{center}
\begin{tikzpicture}

\def\sizeImg{1.25}

%%%%%%%%%%%%%%%%%%%%%%%%%%%%%%%%%%%%%%%%%%

\node at (0, 0.75) [black, rotate=0]  {\scriptsize{\textbf{x}}};   
\draw (0,0) node(n1)  {\includegraphics[height=\sizeImg cm]{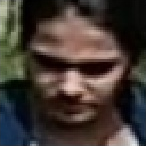}};   
\node at (0.97*\sizeImg, 0.75) [black, rotate=0]  {\scriptsize{\textbf{a}}};   
\draw (0.97*\sizeImg,0) node(n1)  {\includegraphics[height=\sizeImg cm]{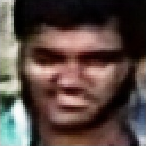}};   
\node at (1.97*\sizeImg, 0.75) [black, rotate=0]  {\scriptsize{\textbf{r}}};   
\draw (1.97*\sizeImg,0) node(n1)  {\includegraphics[height=\sizeImg cm]{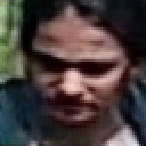}};   

\draw (0,0-1.0*\sizeImg) node(n1)  {\includegraphics[height=\sizeImg cm]{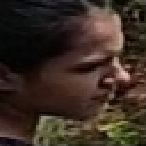}};   
\draw (0.985*\sizeImg,0-1.0*\sizeImg) node(n1)  {\includegraphics[height=\sizeImg cm]{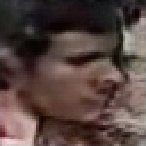}};   
\draw (1.97*\sizeImg,0-1.0*\sizeImg) node(n1)  {\includegraphics[height=\sizeImg cm]{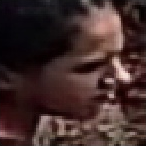}};   

\def\deltaX{0}
\def\deltaY{0}

\node at (-0.85+\deltaX, 0+\deltaY) [black, rotate=90]  {\scriptsize{\text{Session 1}}};   
\node at (-0.85+\deltaX, 0-\sizeImg+\deltaY) [black, rotate=90]  {\scriptsize{\text{Session 2}}};

%%%%%%%%%%%%%%%%%%%%%%%%%%%%%%%%%%%%%%%%%%

\def\deltaX{4.5}
\def\deltaY{0}

\node at (0+\deltaX, 0.75+\deltaY) [black, rotate=0]  {\scriptsize{\textbf{x}}};   
\draw (0+\deltaX,0+\deltaY) node(n1)  {\includegraphics[height=\sizeImg cm]{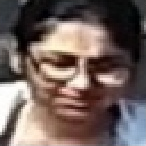}};   
\node at (0.97*\sizeImg+\deltaX, 0.75+\deltaY) [black, rotate=0]  {\scriptsize{\textbf{a}}};   
\draw (0.97*\sizeImg+\deltaX,0+\deltaY) node(n1)  {\includegraphics[height=\sizeImg cm]{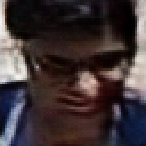}};   
\node at (1.97*\sizeImg+\deltaX, 0.75+\deltaY) [black, rotate=0]  {\scriptsize{\textbf{r}}};   
\draw (1.97*\sizeImg+\deltaX,0+\deltaY) node(n1)  {\includegraphics[height=\sizeImg cm]{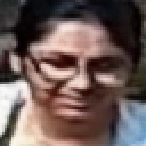}};   

\draw (0+\deltaX,0-1.0*\sizeImg+\deltaY) node(n1)  {\includegraphics[height=\sizeImg cm]{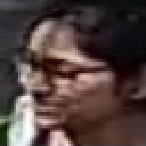}};   
\draw (0.985*\sizeImg+\deltaX,0-1.0*\sizeImg+\deltaY) node(n1)  {\includegraphics[height=\sizeImg cm]{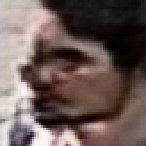}};   
\draw (1.97*\sizeImg+\deltaX,0-1.0*\sizeImg+\deltaY) node(n1)  {\includegraphics[height=\sizeImg cm]{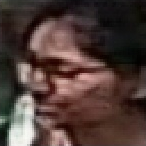}};

\node at (-0.85+\deltaX, 0+\deltaY) [black, rotate=90]  {\scriptsize{\text{Session 1}}};   
\node at (-0.85+\deltaX, 0-\sizeImg+\deltaY) [black, rotate=90]  {\scriptsize{\text{Session 2}}};  

%%%%%%%%%%%%%%%%%%%%%%%%%%%%%%%%%%%%%%%%%%

\def\deltaX{0}
\def\deltaY{-2.85}

\node at (0+\deltaX, 0.75+\deltaY) [black, rotate=0]  {\scriptsize{\textbf{x}}};   
\draw (0+\deltaX,0+\deltaY) node(n1)  {\includegraphics[height=\sizeImg cm]{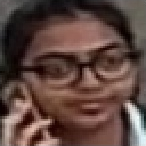}};   
\node at (0.97*\sizeImg+\deltaX, 0.75+\deltaY) [black, rotate=0]  {\scriptsize{\textbf{a}}};   
\draw (0.97*\sizeImg+\deltaX,0+\deltaY) node(n1)  {\includegraphics[height=\sizeImg cm]{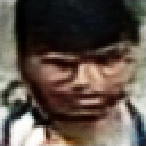}};   
\node at (1.97*\sizeImg+\deltaX, 0.75+\deltaY) [black, rotate=0]  {\scriptsize{\textbf{r}}};   
\draw (1.97*\sizeImg+\deltaX,0+\deltaY) node(n1)  {\includegraphics[height=\sizeImg cm]{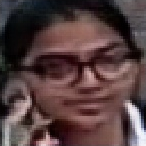}};   

\draw (0+\deltaX,0-1.0*\sizeImg+\deltaY) node(n1)  {\includegraphics[height=\sizeImg cm]{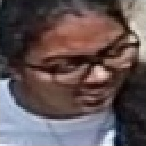}};   
\draw (0.985*\sizeImg+\deltaX,0-1.0*\sizeImg+\deltaY) node(n1)  {\includegraphics[height=\sizeImg cm]{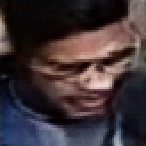}};   
\draw (1.97*\sizeImg+\deltaX,0-1.0*\sizeImg+\deltaY) node(n1)  {\includegraphics[height=\sizeImg cm]{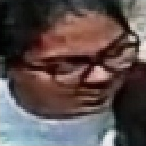}};   

\node at (-0.85+\deltaX, 0+\deltaY) [black, rotate=90]  {\scriptsize{\text{Session 1}}};   
\node at (-0.85+\deltaX, 0-\sizeImg+\deltaY) [black, rotate=90]  {\scriptsize{\text{Session 2}}};  

%%%%%%%%%%%%%%%%%%%%%%%%%%%%%%%%%%%%%%%%%%

\def\deltaX{4.5}
\def\deltaY{-2.85}

\node at (0+\deltaX, 0.75+\deltaY) [black, rotate=0]  {\scriptsize{\textbf{x}}};   
\draw (0+\deltaX,0+\deltaY) node(n1)  {\includegraphics[height=\sizeImg cm]{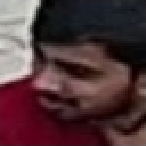}};   
\node at (0.97*\sizeImg+\deltaX, 0.75+\deltaY) [black, rotate=0]  {\scriptsize{\textbf{a}}};   
\draw (0.97*\sizeImg+\deltaX,0+\deltaY) node(n1)  {\includegraphics[height=\sizeImg cm]{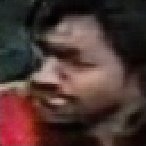}};   
\node at (1.97*\sizeImg+\deltaX, 0.75+\deltaY) [black, rotate=0]  {\scriptsize{\textbf{r}}};   
\draw (1.97*\sizeImg+\deltaX,0+\deltaY) node(n1)  {\includegraphics[height=\sizeImg cm]{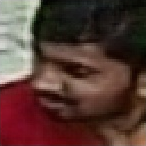}};   

\draw (0+\deltaX,0-1.0*\sizeImg+\deltaY) node(n1)  {\includegraphics[height=\sizeImg cm]{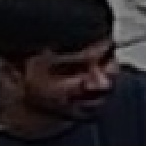}};   
\draw (0.985*\sizeImg+\deltaX,0-1.0*\sizeImg+\deltaY) node(n1)  {\includegraphics[height=\sizeImg cm]{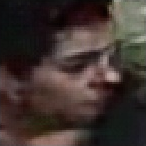}};   
\draw (1.97*\sizeImg+\deltaX,0-1.0*\sizeImg+\deltaY) node(n1)  {\includegraphics[height=\sizeImg cm]{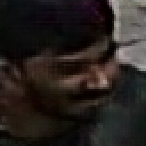}};

\node at (-0.85+\deltaX, 0+\deltaY) [black, rotate=90]  {\scriptsize{\text{Session 1}}};   
\node at (-0.85+\deltaX, 0-\sizeImg+\deltaY) [black, rotate=90]  {\scriptsize{\text{Session 2}}};

\def\deltaY{-6.75}
\def\deltaX{1.2}

\fill [gray!50, rounded corners] (-0.9+\deltaX, 1.3+\deltaY) rectangle (0.9+\deltaX, 1.7+\deltaY);     
\node at (0+\deltaX, 1.5+\deltaY) [black, rotate=0]  {\scriptsize{\textbf{MARS}}};

\draw (0+\deltaX,0+\deltaY) node(n1)  {\includegraphics[width=4 cm]{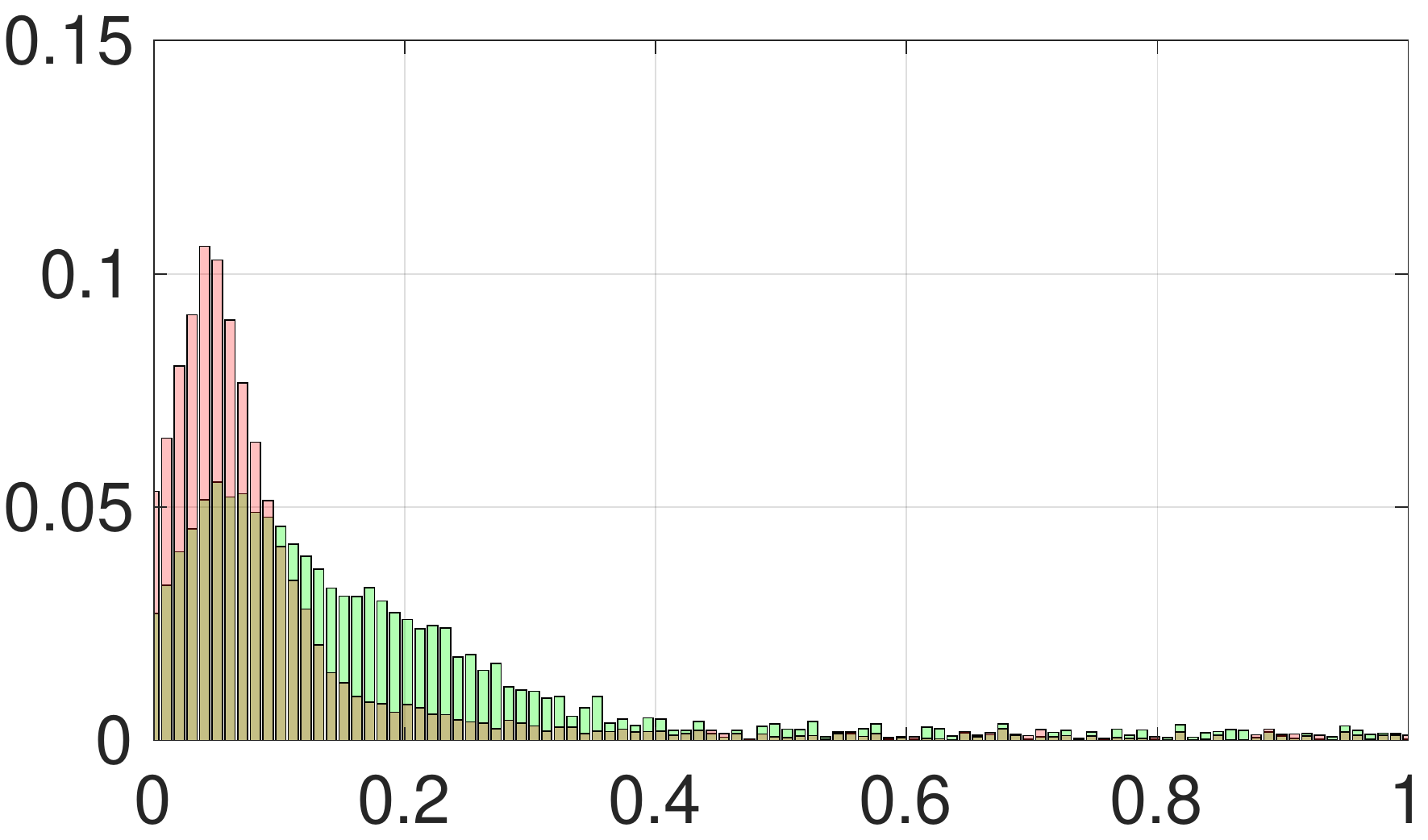}};   
\node at (0+\deltaX, -1.35+\deltaY) [black, rotate=0]  {\scriptsize{Score}};     
\node at (-2.2+\deltaX, 0+\deltaY) [black, rotate=90]  {\scriptsize{Frequency}};

\fill [gray!50, rounded corners] (-0.9+4.5+\deltaX, 1.3+\deltaY) rectangle (0.9+4.5+\deltaX, 1.7+\deltaY);     
\node at (0+4.5+\deltaX, 1.5+\deltaY) [black, rotate=0]  {\scriptsize{\textbf{P-DESTRE}}};     

\draw (4.5+\deltaX,0+\deltaY) node(n1)  {\includegraphics[width=4 cm]{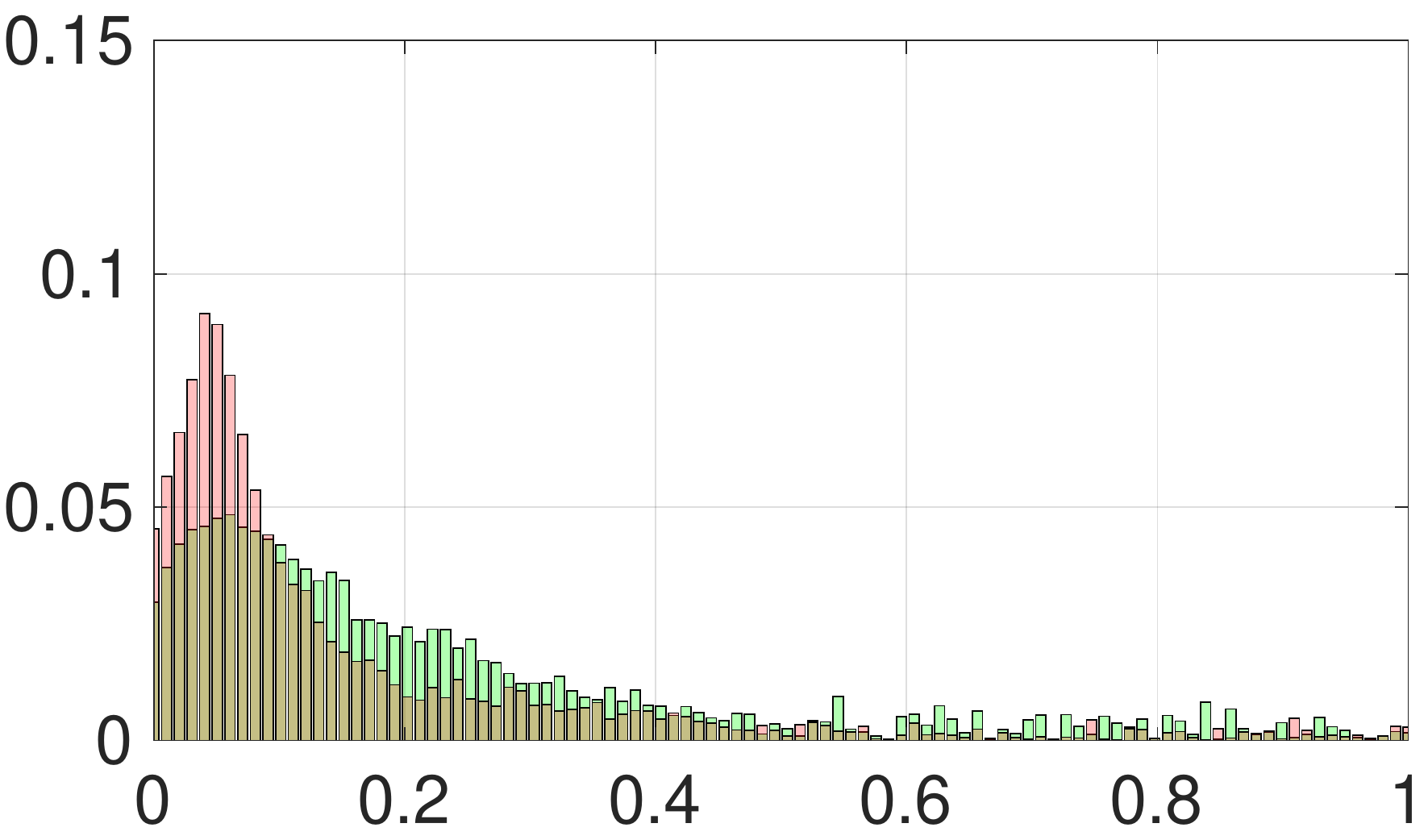}};   
\node at (0+4.5+\deltaX, -1.35+\deltaY) [black, rotate=0]  {\scriptsize{Score}};     
\node at (-2.2+4.5+\deltaX, 0+\deltaY) [black, rotate=90]  {\scriptsize{Frequency}};     

\end{tikzpicture}
    \caption{Top rows: diversity of the de-identified samples for different sessions (sequences) of a single subject. The bottom row provides the decision environments obtained for the MARS (at left) and P-DESTRE (at right) sets, when the \emph{genuine} pairs were exclusively composed of \textbf{a}$_{i,j,.}$/ \textbf{a}$_{i,k,.}$ elements.} 
        \label{fig:sessionDiversity}
    \end{center}
\end{figure}
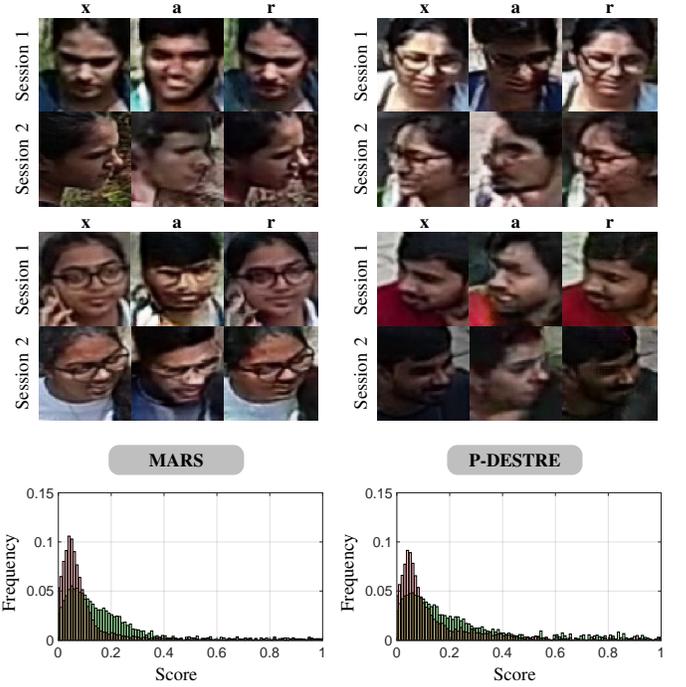

The diversity of IDs generated for different sessions is illustrated in the top rows of Fig.~\ref{fig:sessionDiversity}. Again, the bottom row provides the decision environments when the \emph{genuine} pairs were exclusively composed of  \textbf{a}$_{i,j,.}$/ \textbf{a}$_{i,k,.}$ elements. Here, even though there is an evident separation between both distributions (d' values of 0.491 for MARS and 0.329  for P-DESTRE sets), both genuine distributions were skewed toward the higher values region (i.e., corresponding to the typical \emph{genuine} region), which suggests that in such cases the IDs generated for different sessions of one subject might still share some undesirable patterns.

\subsection{Pose, Background and Facial Expressions Consistency}

\begin{figure*}[ht!]
\begin{center}
\begin{tikzpicture}

\def\sizeImg{1.35}

%%%%%%%%%%%%%%%%%%%%%%%%%%%%%%%%%%%%%%%%%%

\node at (0, 0.85) [black, rotate=0]  {\scriptsize{\textbf{x}}};   
\draw (0,0) node(n1)  {\includegraphics[height=\sizeImg cm]{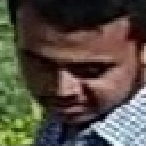}};   
\node at (1*\sizeImg, 0.85) [black, rotate=0]  {\scriptsize{\textbf{a}}};   
\draw (1*\sizeImg,0) node(n1)  {\includegraphics[height=\sizeImg cm]{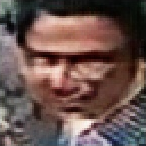}};   
\node at (2*\sizeImg, 0.85) [black, rotate=0]  {\scriptsize{\textbf{r}}};   
\draw (2*\sizeImg,0) node(n1)  {\includegraphics[height=\sizeImg cm]{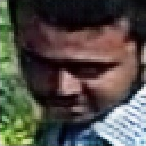}};   

\draw (0,0-1.0*\sizeImg) node(n1)  {\includegraphics[height=\sizeImg cm]{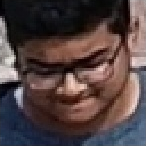}};   
\draw (1*\sizeImg,0-1.0*\sizeImg) node(n1)  {\includegraphics[height=\sizeImg cm]{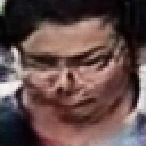}};   
\draw (2*\sizeImg,0-1.0*\sizeImg) node(n1)  {\includegraphics[height=\sizeImg cm]{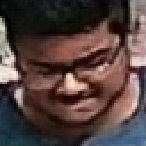}};

\node at (-0.85, 0-\sizeImg/2) [black, rotate=90]  {\scriptsize{\textbf{Background Consistency}}};

\def\deltaX{4.5}
\def\deltaY{0}

\node at (0+\deltaX, 0.85+\deltaY) [black, rotate=0]  {\scriptsize{\textbf{x}}};   
\draw (0+\deltaX,0+\deltaY) node(n1)  {\includegraphics[height=\sizeImg cm]{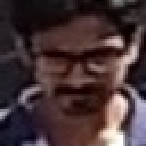}};   
\node at (1*\sizeImg+\deltaX, 0.85+\deltaY) [black, rotate=0]  {\scriptsize{\textbf{a}}};   
\draw (1*\sizeImg+\deltaX,0+\deltaY) node(n1)  {\includegraphics[height=\sizeImg cm]{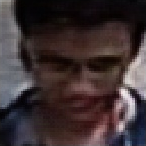}};   
\node at (2*\sizeImg+\deltaX, 0.85+\deltaY) [black, rotate=0]  {\scriptsize{\textbf{r}}};   
\draw (2*\sizeImg+\deltaX,0+\deltaY) node(n1)  {\includegraphics[height=\sizeImg cm]{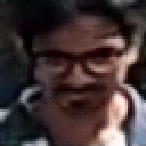}};   

\draw (0+\deltaX,0-1.0*\sizeImg+\deltaY) node(n1)  {\includegraphics[height=\sizeImg cm]{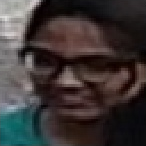}};   
\draw (1*\sizeImg+\deltaX,0-1.0*\sizeImg+\deltaY) node(n1)  {\includegraphics[height=\sizeImg cm]{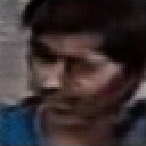}};   
\draw (2*\sizeImg+\deltaX,0-1.0*\sizeImg+\deltaY) node(n1)  {\includegraphics[height=\sizeImg cm]{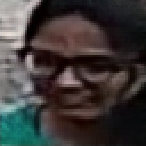}};

\def\deltaX{9.0}
\def\deltaY{0}

\node at (0+\deltaX, 0.85+\deltaY) [black, rotate=0]  {\scriptsize{\textbf{x}}};   
\draw (0+\deltaX,0+\deltaY) node(n1)  {\includegraphics[height=\sizeImg cm]{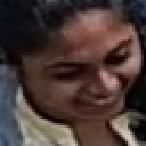}};   
\node at (1*\sizeImg+\deltaX, 0.85+\deltaY) [black, rotate=0]  {\scriptsize{\textbf{a}}};   
\draw (1*\sizeImg+\deltaX,0+\deltaY) node(n1)  {\includegraphics[height=\sizeImg cm]{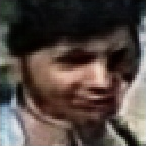}};   
\node at (2*\sizeImg+\deltaX, 0.85+\deltaY) [black, rotate=0]  {\scriptsize{\textbf{r}}};   
\draw (2*\sizeImg+\deltaX,0+\deltaY) node(n1)  {\includegraphics[height=\sizeImg cm]{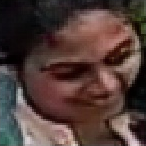}};   

\draw (0+\deltaX,0-1.0*\sizeImg+\deltaY) node(n1)  {\includegraphics[height=\sizeImg cm]{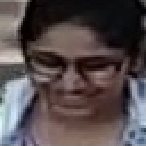}};   
\draw (1*\sizeImg+\deltaX,0-1.0*\sizeImg+\deltaY) node(n1)  {\includegraphics[height=\sizeImg cm]{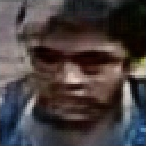}};   
\draw (2*\sizeImg+\deltaX,0-1.0*\sizeImg+\deltaY) node(n1)  {\includegraphics[height=\sizeImg cm]{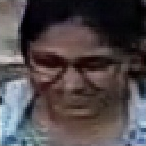}};

\def\deltaX{13.5}
\def\deltaY{0}

\node at (0+\deltaX, 0.85+\deltaY) [black, rotate=0]  {\scriptsize{\textbf{x}}};   
\draw (0+\deltaX,0+\deltaY) node(n1)  {\includegraphics[height=\sizeImg cm]{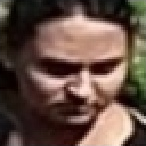}};   
\node at (1*\sizeImg+\deltaX, 0.85+\deltaY) [black, rotate=0]  {\scriptsize{\textbf{a}}};   
\draw (1*\sizeImg+\deltaX,0+\deltaY) node(n1)  {\includegraphics[height=\sizeImg cm]{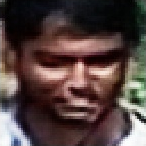}};   
\node at (2*\sizeImg+\deltaX, 0.85+\deltaY) [black, rotate=0]  {\scriptsize{\textbf{r}}};   
\draw (2*\sizeImg+\deltaX,0+\deltaY) node(n1)  {\includegraphics[height=\sizeImg cm]{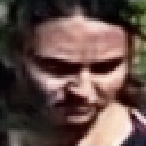}};   

\draw (0+\deltaX,0-1.0*\sizeImg+\deltaY) node(n1)  {\includegraphics[height=\sizeImg cm]{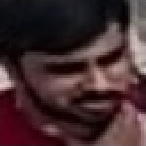}};   
\draw (1*\sizeImg+\deltaX,0-1.0*\sizeImg+\deltaY) node(n1)  {\includegraphics[height=\sizeImg cm]{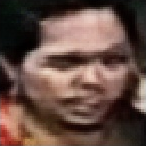}};   
\draw (2*\sizeImg+\deltaX,0-1.0*\sizeImg+\deltaY) node(n1)  {\includegraphics[height=\sizeImg cm]{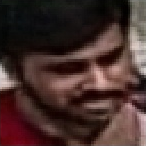}};

%%%%%%%%%%%%%%%%%%%%%%%%%%%%%%%%%%%%%%%%%%

\def\deltaX{0}
\def\deltaY{-3.25}

\node at (0+\deltaX, 0.85+\deltaY) [black, rotate=0]  {\scriptsize{\textbf{x}}};   
\draw (0+\deltaX,0+\deltaY) node(n1)  {\includegraphics[height=\sizeImg cm]{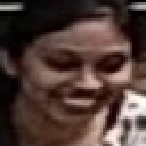}};   
\node at (1*\sizeImg+\deltaX, 0.85+\deltaY) [black, rotate=0]  {\scriptsize{\textbf{a}}};   
\draw (1*\sizeImg+\deltaX,0+\deltaY) node(n1)  {\includegraphics[height=\sizeImg cm]{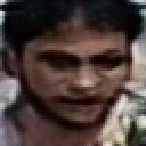}};   
\node at (2*\sizeImg+\deltaX, 0.85+\deltaY) [black, rotate=0]  {\scriptsize{\textbf{r}}};   
\draw (2*\sizeImg+\deltaX,0+\deltaY) node(n1)  {\includegraphics[height=\sizeImg cm]{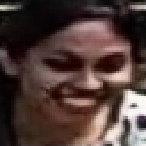}};   

\draw (0+\deltaX,0-1.0*\sizeImg+\deltaY) node(n1)  {\includegraphics[height=\sizeImg cm]{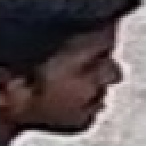}};   
\draw (1*\sizeImg+\deltaX,0-1.0*\sizeImg+\deltaY) node(n1)  {\includegraphics[height=\sizeImg cm]{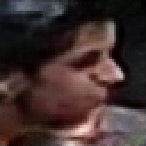}};   
\draw (2*\sizeImg+\deltaX,0-1.0*\sizeImg+\deltaY) node(n1)  {\includegraphics[height=\sizeImg cm]{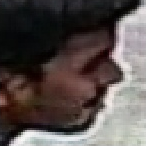}};   

\node at (-0.85, 0-\sizeImg/2+\deltaY) [black, rotate=90]  {\scriptsize{\textbf{Pose Consistency}}};

\def\deltaX{4.5}
\def\deltaY{-3.25}

\node at (0+\deltaX, 0.85+\deltaY) [black, rotate=0]  {\scriptsize{\textbf{x}}};   
\draw (0+\deltaX,0+\deltaY) node(n1)  {\includegraphics[height=\sizeImg cm]{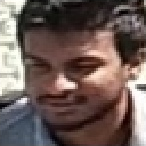}};   
\node at (1*\sizeImg+\deltaX, 0.85+\deltaY) [black, rotate=0]  {\scriptsize{\textbf{a}}};   
\draw (1*\sizeImg+\deltaX,0+\deltaY) node(n1)  {\includegraphics[height=\sizeImg cm]{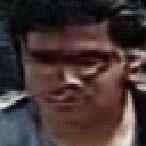}};   
\node at (2*\sizeImg+\deltaX, 0.85+\deltaY) [black, rotate=0]  {\scriptsize{\textbf{r}}};   
\draw (2*\sizeImg+\deltaX,0+\deltaY) node(n1)  {\includegraphics[height=\sizeImg cm]{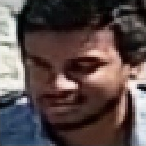}};   

\draw (0+\deltaX,0-1.0*\sizeImg+\deltaY) node(n1)  {\includegraphics[height=\sizeImg cm]{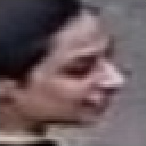}};   
\draw (1*\sizeImg+\deltaX,0-1.0*\sizeImg+\deltaY) node(n1)  {\includegraphics[height=\sizeImg cm]{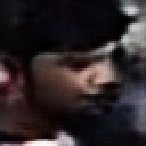}};   
\draw (2*\sizeImg+\deltaX,0-1.0*\sizeImg+\deltaY) node(n1)  {\includegraphics[height=\sizeImg cm]{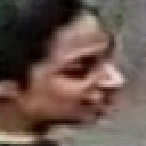}};   

\def\deltaX{9}
\def\deltaY{-3.25}

\node at (0+\deltaX, 0.85+\deltaY) [black, rotate=0]  {\scriptsize{\textbf{x}}};   
\draw (0+\deltaX,0+\deltaY) node(n1)  {\includegraphics[height=\sizeImg cm]{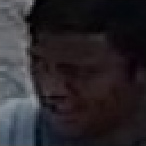}};   
\node at (1*\sizeImg+\deltaX, 0.85+\deltaY) [black, rotate=0]  {\scriptsize{\textbf{a}}};   
\draw (1*\sizeImg+\deltaX,0+\deltaY) node(n1)  {\includegraphics[height=\sizeImg cm]{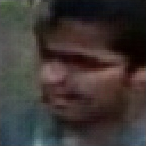}};   
\node at (2*\sizeImg+\deltaX, 0.85+\deltaY) [black, rotate=0]  {\scriptsize{\textbf{r}}};   
\draw (2*\sizeImg+\deltaX,0+\deltaY) node(n1)  {\includegraphics[height=\sizeImg cm]{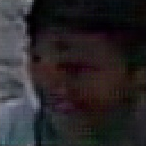}};   

\draw (0+\deltaX,0-1.0*\sizeImg+\deltaY) node(n1)  {\includegraphics[height=\sizeImg cm]{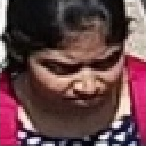}};   
\draw (1*\sizeImg+\deltaX,0-1.0*\sizeImg+\deltaY) node(n1)  {\includegraphics[height=\sizeImg cm]{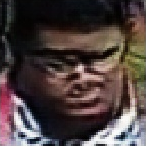}};   
\draw (2*\sizeImg+\deltaX,0-1.0*\sizeImg+\deltaY) node(n1)  {\includegraphics[height=\sizeImg cm]{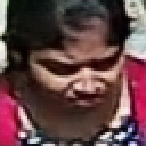}};   

\def\deltaX{13.5}
\def\deltaY{-3.25}

\node at (0+\deltaX, 0.85+\deltaY) [black, rotate=0]  {\scriptsize{\textbf{x}}};   
\draw (0+\deltaX,0+\deltaY) node(n1)  {\includegraphics[height=\sizeImg cm]{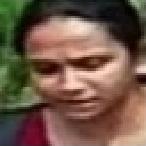}};   
\node at (1*\sizeImg+\deltaX, 0.85+\deltaY) [black, rotate=0]  {\scriptsize{\textbf{a}}};   
\draw (1*\sizeImg+\deltaX,0+\deltaY) node(n1)  {\includegraphics[height=\sizeImg cm]{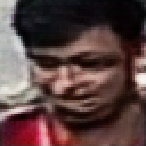}};   
\node at (2*\sizeImg+\deltaX, 0.85+\deltaY) [black, rotate=0]  {\scriptsize{\textbf{r}}};   
\draw (2*\sizeImg+\deltaX,0+\deltaY) node(n1)  {\includegraphics[height=\sizeImg cm]{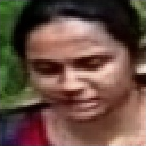}};   

\draw (0+\deltaX,0-1.0*\sizeImg+\deltaY) node(n1)  {\includegraphics[height=\sizeImg cm]{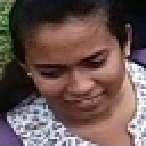}};   
\draw (1*\sizeImg+\deltaX,0-1.0*\sizeImg+\deltaY) node(n1)  {\includegraphics[height=\sizeImg cm]{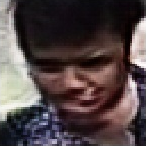}};   
\draw (2*\sizeImg+\deltaX,0-1.0*\sizeImg+\deltaY) node(n1)  {\includegraphics[height=\sizeImg cm]{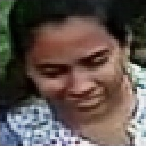}};   

%%%%%%%%%%%%%%%%%%%%%%%%%%%%%%%%%%%%%%%%%%

\def\deltaX{0}
\def\deltaY{-6.5}

\node at (0+\deltaX, 0.85+\deltaY) [black, rotate=0]  {\scriptsize{\textbf{x}}};   
\draw (0+\deltaX,0+\deltaY) node(n1)  {\includegraphics[height=\sizeImg cm]{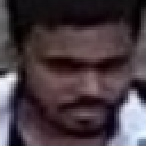}};   
\node at (1*\sizeImg+\deltaX, 0.85+\deltaY) [black, rotate=0]  {\scriptsize{\textbf{a}}};   
\draw (1*\sizeImg+\deltaX,0+\deltaY) node(n1)  {\includegraphics[height=\sizeImg cm]{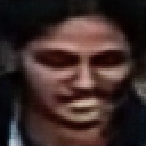}};   
\node at (2*\sizeImg+\deltaX, 0.85+\deltaY) [black, rotate=0]  {\scriptsize{\textbf{r}}};   
\draw (2*\sizeImg+\deltaX,0+\deltaY) node(n1)  {\includegraphics[height=\sizeImg cm]{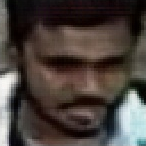}};   

\draw (0+\deltaX,0-1.0*\sizeImg+\deltaY) node(n1)  {\includegraphics[height=\sizeImg cm]{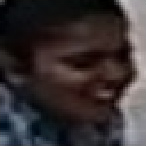}};   
\draw (1*\sizeImg+\deltaX,0-1.0*\sizeImg+\deltaY) node(n1)  {\includegraphics[height=\sizeImg cm]{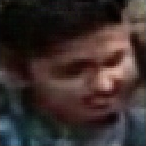}};   
\draw (2*\sizeImg+\deltaX,0-1.0*\sizeImg+\deltaY) node(n1)  {\includegraphics[height=\sizeImg cm]{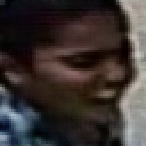}};   

\node at (-0.85, 0-\sizeImg/2+\deltaY) [black, rotate=90]  {\scriptsize{\textbf{Facial Expressions Consistency}}};

\def\deltaX{4.5}
\def\deltaY{-6.5}

\node at (0+\deltaX, 0.85+\deltaY) [black, rotate=0]  {\scriptsize{\textbf{x}}};   
\draw (0+\deltaX,0+\deltaY) node(n1)  {\includegraphics[height=\sizeImg cm]{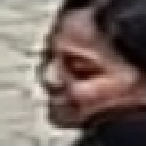}};   
\node at (1*\sizeImg+\deltaX, 0.85+\deltaY) [black, rotate=0]  {\scriptsize{\textbf{a}}};   
\draw (1*\sizeImg+\deltaX,0+\deltaY) node(n1)  {\includegraphics[height=\sizeImg cm]{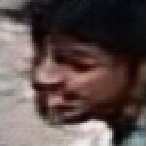}};   
\node at (2*\sizeImg+\deltaX, 0.85+\deltaY) [black, rotate=0]  {\scriptsize{\textbf{r}}};   
\draw (2*\sizeImg+\deltaX,0+\deltaY) node(n1)  {\includegraphics[height=\sizeImg cm]{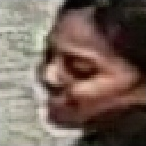}};   

\draw (0+\deltaX,0-1.0*\sizeImg+\deltaY) node(n1)  {\includegraphics[height=\sizeImg cm]{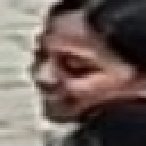}};   
\draw (1*\sizeImg+\deltaX,0-1.0*\sizeImg+\deltaY) node(n1)  {\includegraphics[height=\sizeImg cm]{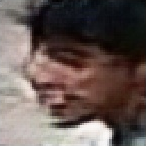}};   
\draw (2*\sizeImg+\deltaX,0-1.0*\sizeImg+\deltaY) node(n1)  {\includegraphics[height=\sizeImg cm]{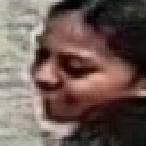}};   

\def\deltaX{9}
\def\deltaY{-6.5}

\node at (0+\deltaX, 0.85+\deltaY) [black, rotate=0]  {\scriptsize{\textbf{x}}};   
\draw (0+\deltaX,0+\deltaY) node(n1)  {\includegraphics[height=\sizeImg cm]{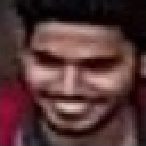}};   
\node at (1*\sizeImg+\deltaX, 0.85+\deltaY) [black, rotate=0]  {\scriptsize{\textbf{a}}};   
\draw (1*\sizeImg+\deltaX,0+\deltaY) node(n1)  {\includegraphics[height=\sizeImg cm]{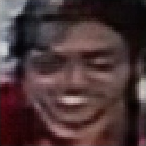}};   
\node at (2*\sizeImg+\deltaX, 0.85+\deltaY) [black, rotate=0]  {\scriptsize{\textbf{r}}};   
\draw (2*\sizeImg+\deltaX,0+\deltaY) node(n1)  {\includegraphics[height=\sizeImg cm]{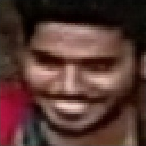}};   

\draw (0+\deltaX,0-1.0*\sizeImg+\deltaY) node(n1)  {\includegraphics[height=\sizeImg cm]{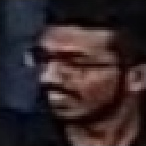}};   
\draw (1*\sizeImg+\deltaX,0-1.0*\sizeImg+\deltaY) node(n1)  {\includegraphics[height=\sizeImg cm]{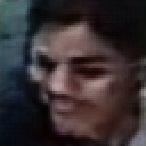}};   
\draw (2*\sizeImg+\deltaX,0-1.0*\sizeImg+\deltaY) node(n1)  {\includegraphics[height=\sizeImg cm]{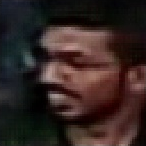}};   

\def\deltaX{13.5}
\def\deltaY{-6.5}

\node at (0+\deltaX, 0.85+\deltaY) [black, rotate=0]  {\scriptsize{\textbf{x}}};   
\draw (0+\deltaX,0+\deltaY) node(n1)  {\includegraphics[height=\sizeImg cm]{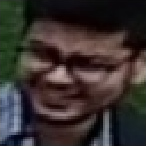}};   
\node at (1*\sizeImg+\deltaX, 0.85+\deltaY) [black, rotate=0]  {\scriptsize{\textbf{a}}};   
\draw (1*\sizeImg+\deltaX,0+\deltaY) node(n1)  {\includegraphics[height=\sizeImg cm]{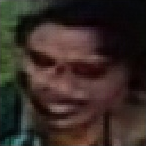}};   
\node at (2*\sizeImg+\deltaX, 0.85+\deltaY) [black, rotate=0]  {\scriptsize{\textbf{r}}};   
\draw (2*\sizeImg+\deltaX,0+\deltaY) node(n1)  {\includegraphics[height=\sizeImg cm]{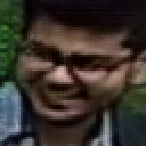}};   

\draw (0+\deltaX,0-1.0*\sizeImg+\deltaY) node(n1)  {\includegraphics[height=\sizeImg cm]{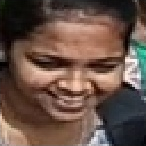}};   
\draw (1*\sizeImg+\deltaX,0-1.0*\sizeImg+\deltaY) node(n1)  {\includegraphics[height=\sizeImg cm]{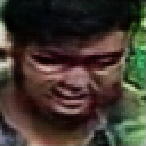}};   
\draw (2*\sizeImg+\deltaX,0-1.0*\sizeImg+\deltaY) node(n1)  {\includegraphics[height=\sizeImg cm]{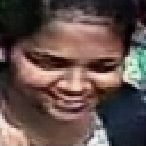}};   

%%%%%%%%%%%%%%%%%%%%%%%%%%%%%%%%%%%%%%%%%%

\end{tikzpicture}
    \caption{Examples illustrating the background consistency (upper row), pose consistency (middle row) and facial expressions consistency (bottom row) between the raw samples \textbf{x} and their de-identified versions \textbf{a}.}
        \label{fig:qualitative}
    \end{center}
\end{figure*}
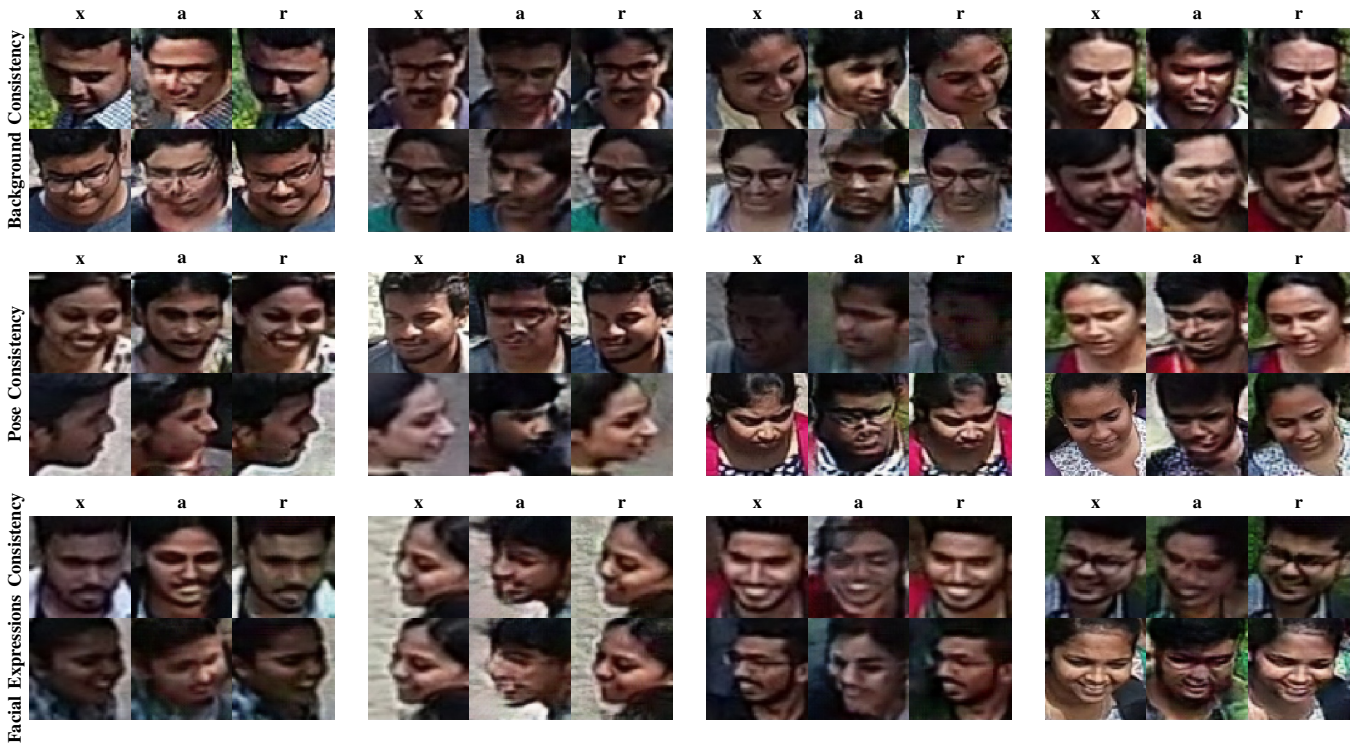

For photo-realism purposes, not only the pose of \textbf{x}$_.$/\textbf{a}$_.$ elements should be consistent, but also the background features in both images should be similar and even the facial expression should agree. According to our loss formulation and experiments, we observed that the distribution loss term $\mathcal{L}_{\text{dis}}$ plays a key role in guaranteeing such kinds of consistencies. To quantitatively perceive the pose consistency, we compared the pose estimation \emph{yaw}, \emph{pitch} and \emph{roll} values obtained by the Deep Head Pose~\cite{Ruiz2018} method for  \textbf{x}/\textbf{a} elements. As the model was not specifically trained for each dataset, errors in pose inference were relatively frequent and covered about 20\% of the samples of the BIODI set, 7\% of the elements in MARS and 28\% of the P-DESTRE images. These cases were rejected under human inspection. For the remaining cases, we measured the absolute difference between the 3D angle values, obtaining average yaw errors of 0.177 $\pm$ 0.091, 0.184 $\pm$ 0.087, 0.140 $\pm$ 0.075 (BIODI, MAR, P-DESTRE), pitch errors of 0.101 $\pm$ 0.068, 0.120 $\pm$ 0.070, 0.113 $\pm$ 0.055 and roll errors 0.021 $\pm$ 0.006, 0.0.25 $\pm$ 0.005, 0.022 $\pm$ 0.004 (in radians), which we regarded to positively confirm the coherence in pose between the \textbf{x}/\textbf{a} elements. The consistency of the remaining factors (background and facial expressions) was only perceived in a subjective way, under visual perception. Figure~\ref{fig:qualitative} illustrates the three types of consistencies addressed in this section.

\subsection{Ablation Experiments And Difficult Cases}

The most frequent variations in the results with respect to changes in the parameterizations of the loss formulation are illustrated in Table~\ref{tab:ablation}. The left column gives the changes in each parameter with respect to the configuration considered optimal (described in sec.~\ref{ssec:datasets}), and the images in the right column illustrate the typical failure cases. In this experiment, every parameter was orthogonally decreased ($\downarrow$) or augmented ($\uparrow$) one order of magnitude in the learning phase. At first, when  the $\omega_{\text{mse}}$ weight was decreased, the reconstructed samples started to appear pixelized and blurred. As an effect of the variation of $\omega_{\text{ano}}$ weight, \textbf{x}/\textbf{a} started to resemble each other in a much more evident way, and in some cases the \textbf{U}$_e$ started to work practically as an identity operator. Decreasing the weights of the adversarial discriminator $\omega_{\text{adv}}$ had a catastrophic effect in the \textbf{a} results, that completely loosen their \emph{face} appearance. When decreasing the $\omega_{\text{div}}$ parameter, the de-identified images started to look alike their \textbf{x} counterparts, while the $\omega_{\text{con}}$ weight stressed the temporal consistency requirements. By decreasing the value of the $\omega_{\text{dis}}$ parameter, the resulting \textbf{a} elements have very different color/brightness distributions with respect to \textbf{x}, which strongly decreases the photo realism. Finally, another catastrophic change occurred when the maximum gradient $\delta_{\text{gp}}$ allowed for adjusting the weights of the adversarial discriminator in the learning phase was increased, which typically causes the divergence of the training phase.

\begin{table}[h!]
\setlength{\tabcolsep}{1pt}
\renewcommand{\arraystretch}{0.5}
\centering
     \caption{Illustration of the typical variations in the results of the proposed method with respect to changes in each term used in the loss formulation.}
     \label{tab:ablation}
\begin{tabular}{|m{1.52cm}|C{7.25cm}|}

\hline
\multicolumn{1}{|c|}{  \scriptsize{\textbf{Parameter}}} & \multicolumn{1}{c|}{  \scriptsize{\textbf{Typical Results}}} \\ \hline

\scriptsize{$\omega_{\text{mse}} \downarrow(\div 10)$} & 
\setlength{\tabcolsep}{0.0cm}
\begin{tabular}{cccccc}
\includegraphics[height=1.1 cm]{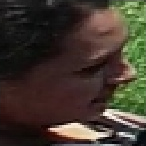} & 
\includegraphics[height=1.1 cm]{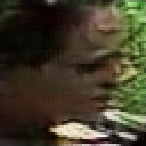} \hspace{0.5cm}&
\includegraphics[height=1.1 cm]{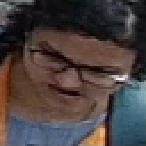} &
\includegraphics[height=1.1 cm]{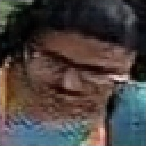}  \hspace{0.5cm}&
\includegraphics[height=1.1 cm]{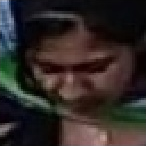} &
\includegraphics[height=1.1 cm]{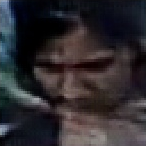} \\
\scriptsize{\textbf{x}} & \scriptsize{\textbf{r}} & \scriptsize{\textbf{x}} & \scriptsize{\textbf{r}} & \scriptsize{\textbf{x}} & \scriptsize{\textbf{r}} \\
\end{tabular}
\\ \hline
\scriptsize{$\omega_{\text{adv}} \downarrow(\div 10)$} & 
\setlength{\tabcolsep}{0.0cm}
\begin{tabular}{cccccc}
\includegraphics[height=1.1 cm]{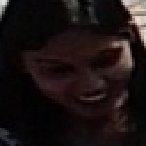} & 
\includegraphics[height=1.1 cm]{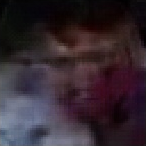} \hspace{0.5cm}&
\includegraphics[height=1.1 cm]{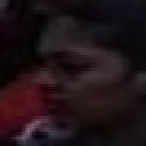} &
\includegraphics[height=1.1 cm]{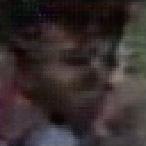}  \hspace{0.5cm}&
\includegraphics[height=1.1 cm]{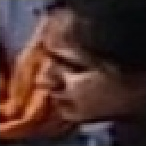} &
\includegraphics[height=1.1 cm]{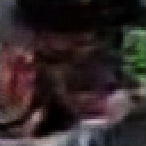} \\
\scriptsize{\textbf{x}} & \scriptsize{\textbf{a}} & \scriptsize{\textbf{x}} & \scriptsize{\textbf{a}} & \scriptsize{\textbf{x}} & \scriptsize{\textbf{a}} \\
\end{tabular}
\\ \hline
\scriptsize{$\omega_{\text{ano}} \downarrow(\div 10)$} & 
\setlength{\tabcolsep}{0.0cm}
\begin{tabular}{cccccc}
\includegraphics[height=1.1 cm]{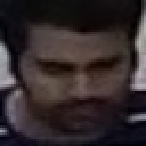} & 
\includegraphics[height=1.1 cm]{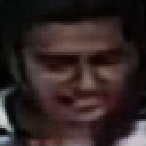} \hspace{1cm}&
\includegraphics[height=1.1 cm]{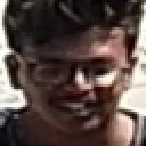} &
\includegraphics[height=1.1 cm]{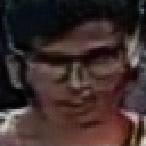}  \hspace{1cm}&
\includegraphics[height=1.1 cm]{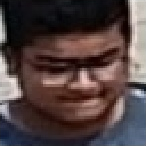} &
\includegraphics[height=1.1 cm]{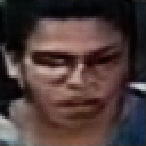} \\
\scriptsize{\textbf{x}} & \scriptsize{\textbf{a}} & \scriptsize{\textbf{x}} & \scriptsize{\textbf{a}} & \scriptsize{\textbf{x}} & \scriptsize{\textbf{a}} \\
\end{tabular}
\\ \hline
\scriptsize{$\omega_{\text{con}} \downarrow(\div 10)$} & 
\setlength{\tabcolsep}{0.0cm}
\begin{tabular}{cccccc}
\includegraphics[height=1.1 cm]{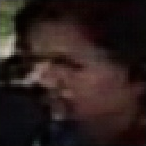} & 
\includegraphics[height=1.1 cm]{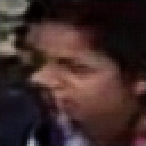} \hspace{1cm}&
\includegraphics[height=1.1 cm]{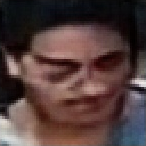} &
\includegraphics[height=1.1 cm]{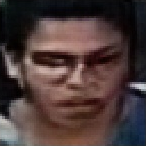}  \hspace{1cm}&
\includegraphics[height=1.1 cm]{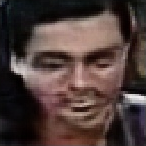} &
\includegraphics[height=1.1 cm]{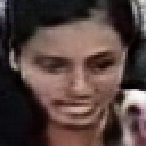} \\
\scriptsize{\textbf{a}$_1$} & \scriptsize{\textbf{a}$_2$} & \scriptsize{\textbf{a}$_1$} & \scriptsize{\textbf{a}$_2$} & \scriptsize{\textbf{a}$_1$} & \scriptsize{\textbf{a}$_2$} \\
\end{tabular}
\\ \hline
\scriptsize{$\omega_{\text{div}} \downarrow(\div 10)$} & 
\setlength{\tabcolsep}{0.0cm}
\begin{tabular}{cccccc}
\includegraphics[height=1.1 cm]{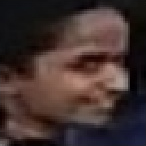} & 
\includegraphics[height=1.1 cm]{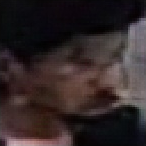} \hspace{1cm}&
\includegraphics[height=1.1 cm]{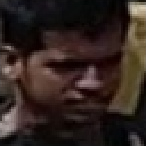} &
\includegraphics[height=1.1 cm]{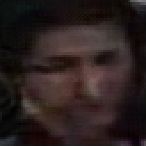}  \hspace{1cm}&
\includegraphics[height=1.1 cm]{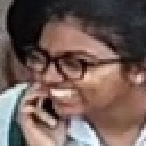} &
\includegraphics[height=1.1 cm]{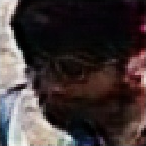} \\
\scriptsize{\textbf{x}} & \scriptsize{\textbf{a}} & \scriptsize{\textbf{x}} & \scriptsize{\textbf{a}} & \scriptsize{\textbf{x}} & \scriptsize{\textbf{a}} \\
\end{tabular}
\\ \hline
\scriptsize{$\omega_{\text{dis}} \downarrow(\div 10)$} & 
\setlength{\tabcolsep}{0.0cm}
\begin{tabular}{cccccc}
\includegraphics[height=1.1 cm]{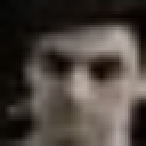} & 
\includegraphics[height=1.1 cm]{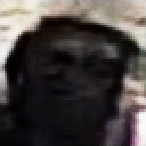} \hspace{1cm}&
\includegraphics[height=1.1 cm]{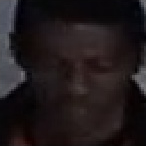} &
\includegraphics[height=1.1 cm]{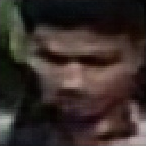}  \hspace{1cm}&
\includegraphics[height=1.1 cm]{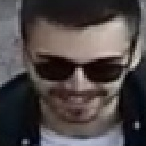} &
\includegraphics[height=1.1 cm]{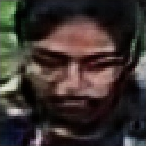} \\
\scriptsize{\textbf{x}} & \scriptsize{\textbf{a}} & \scriptsize{\textbf{x}} & \scriptsize{\textbf{a}} & \scriptsize{\textbf{x}} & \scriptsize{\textbf{a}} \\
\end{tabular}
\\ \hline

\scriptsize{$\delta_{\text{gp}} \uparrow(\times 10)$} & 
\setlength{\tabcolsep}{0.0cm}
\begin{tabular}{cccccc}
\includegraphics[height=1.1 cm]{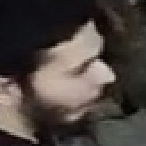} & 
\includegraphics[height=1.1 cm]{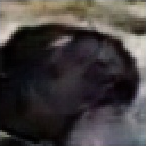} \hspace{1cm}&
\includegraphics[height=1.1 cm]{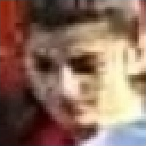} &
\includegraphics[height=1.1 cm]{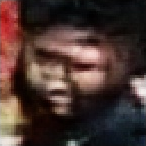}  \hspace{1cm}&
\includegraphics[height=1.1 cm]{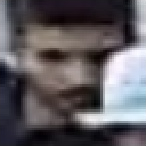} &
\includegraphics[height=1.1 cm]{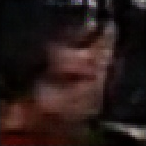} \\
\scriptsize{\textbf{x}} & \scriptsize{\textbf{a}} & \scriptsize{\textbf{x}} & \scriptsize{\textbf{a}} & \scriptsize{\textbf{x}} & \scriptsize{\textbf{a}} \\
\end{tabular}
\\ \hline

\end{tabular}
\end{table}

\section{Conclusions}
\label{sec:Conclusions}
 
This paper addressed the security/privacy balance in visual surveillance environments. While data protection regulations forbid the public disclosure of personal sensitive information, there are scenarios, such as crime scene investigation, where the identification of the subjects is of most importance. Accordingly, we described a solution composed of one \emph{public} module, that detects the faces in each frame and produces the corresponding de-identified versions, where the identity information is surrogated in a photo-realistic and seamless way. Such samples are overlapped in the data stream, which is published without compromising subjects' privacy. This process runs \emph{in situ}, such that no privacy-sensitive information is passed through the network. Next, a \emph{private} module - available exclusively to law enforcement agencies -   is able to reconstruct the original scene and disclosure the actual identity of the subjects exclusively based the public stream.

The proposed solution is landmarks-free and particularly suitable for low resolution data. We designed a two-stage learning process, with a conditional generative adversarial network composed of two different entities (an \emph{encoder} and a \emph{decoder}) that have as common goal to fool an adversarial opponent. This joint optimization process enables them to intrinsically share knowledge about the image features that should be hidden in the encoded data to enable proper reconstruction. The whole process generates highly realistic faces that preserve pose, lighting, background and facial expressions information. Also, it enables full control over the facial attributes that should be preserved between the raw and de-identified streams, which is important to broad the applicability of the system. The experiments were conducted in three visual surveillance datasets, and support the usability of the proposed method. 

\section*{Acknowledgements}

This work is funded by FCT/MEC through national funds and  co-funded by FEDER - PT2020 partnership agreement under the projects UIDB/EEA/50008/2020, POCI-01-0247-FEDER-033395 and C4: Cloud Computing Competence Centre.

{\small
\bibliographystyle{ieee}

\begin{thebibliography}{10}

\bibitem{Agrawal2011}
P.~Agrawal and P.~Narayanan.  Person De-Identification in Videos. \emph{IEEE Transactions on Circuits and Systems for Video Technology}, vol. 21, no.3, pag. 299--310, 2011.

\bibitem{Bao2018}
J.~Bao, D.~Chen, F.~Wen, H.~Li and G.~Hua. Towards open-set identity preserving face synthesis. In proceedings of the \emph{IEEE International Conference on Computer Vision and Pattern Recognition}, doi: \url{10.1109/CVPR.2018.00702}, 2018.

\bibitem{Blanz2004}
V. Blanz, K. Scherbaum, T. Vetter and H.-P. Seidel. Exchanging faces in images. \emph{Computer Graphics Forum}, vol. 23, no. 3, pag. 669--676, 2004.

\bibitem{Bitouk2008}
D. Bitouk, N. Kumar, S. Dhillon, P. Belhumeur and S. Nayar. Face Swapping: Automatically Replacing Faces in Photographs. \emph{ACM Transactions on Graphics}, vol. 27, issue 3, doi: \url{10.1145/1360612.1360638}, 2008.

\bibitem{Boyle2000}
M. Boyle, C. Edwards and S. Greenberg. The Effects of Filtered Video on Awareness and Privacy. In proceedings of the \emph{ACM Conference on Computer Supported Cooperative Work}, pag. 1--10, 2000.

\bibitem{Brkic2017}
K. Brkic, I. Sikiric, T. Hrkac and Z. Kalafatic. I know that person: Generative full body and face de-identification of people in images. In proceedings of the \emph{IEEE Conference on Computer Vision and Pattern Recognition Workshops}, doi: \url{10.1109/CVPRW.2017.173}, 2017.

\bibitem{Butler2015}
D. Butler, J. Huang, F. Roesner and M. Cakmak. The privacy-utility tradeoff for remotely tele-operated robots. In 
proceedings of the \emph{Tenth Annual ACM/IEEE International Conference on Human-Robot Interaction}, doi: \url{10.1145/2696454.2696484}, 2015.

\bibitem{Cao2018}
J. Cao, Y. Li and Z. Zhang. Partially shared multi-task convolutional neural network with local constraint for face attribute learning. In proceedings of the \emph{IEEE Conference on Computer Vision and Pattern Recognition}, pag. 4290--4299, 2018.

\bibitem{Cao2018b}
Q.~Cao, L.~Shen, W.~Xie, O.~Parkhi and A.~Zisserman. VGGFace2: A dataset for recognising faces across pose and age.  In proceedings of the \emph{IEEE Conference on Automatic Face and Gesture Recognition}, doi: \url{10.1109/FG.2018.00020}, 2018.

\bibitem{Chattopadhyay2007}
A. Chattopadhyay and T. Boult. PrivacyCam: a Privacy Preserving Camera Using ucLinox on the Blackfin DSP. In proceedings of the \emph{IEEE Conference on Computer Vision and Pattern Recognition}, doi: \url{10.1109/CVPR.2007.383413}, 2007.

\bibitem{Dale2011}
K. Dale, K. Sunkavalli, M. K. Johnson, D. Vlasic, W. Matusik and H.~Pfister. Video face replacement. \emph{ACM Transactions on Graphics}, vol. 30, no. 6, pag. 1--10, 2011.


\bibitem{Dehghan2015}
A. Dehghan,A., S. ssari, and M. Shah. Gmmcp tracker: Globally optimal generalized maximum multi clique problem for multiple object tracking. In proceedings of the \emph{IEEE Conference on Computer Vision and Pattern Recognition}, doi: \url{10.1109/CVPR.2015.7299036}, 2015. 


\bibitem{Deng2019}
J.~Deng, J.~Guo and S.~Zafeiriou. Single-Stage Joint Face Detection and Alignment. In proceedings of the \emph{IEEE/CVF International Conference on Computer Vision Workshop}, doi: \url{10.1109/ICCVW.2019.00228}, 2019. 

\bibitem{Denemark2016}
T.~Denemark, M.~Boroumand and J.~Fridrich. Steganalysis Features for Content-Adaptive JPEG Steganography. \emph{IEEE Transactions on Information Forensics and Security}, vol. 11, no. 8, pag. 1736--1746, 2016.

\bibitem{Du2014}
L. Du, M. Yi, E. Blasch and H. Ling. GARP-face: Balancing privacy protection and utility preservation in face de-identification. In proceedings of the \emph{IEEE International Joint Conference on Biometrics}, doi: \url{10.1109/BTAS.2014.6996249}, 2014.

\bibitem{Dufaux2008}
F. Dufaux and T. Ebrahimi. Scrambling for Privacy Protection in Video Surveillance Systems. \emph{IEEE Transactions on Circuits and Systems for Video Technology}, vol. 18, no. 8, pag. 1168--1174, 2008.



\bibitem{Felzenszwalb2010}
P. Felzenszwalb, R. Girshick, D. McAllester and  D. Ramanan. Object detection with discriminatively trained part-based models. \emph{IEEE Transactions on Pattern Analysis and Machine Intelligence}, vol. 32, no. 9, pag. 1627--1645, 2010.

\bibitem{Gafni2019}
O.~Gafni, L.~Wolf and Y.~Taigman.  Live Face De-Identification in Video. In proceedings of the \emph{IEEE International Conference on Computer Vision}, pag. 9378--9387, \url{https://arxiv.org/abs/1911.08348v1}, 2019.

\bibitem{Gross2009}
R.~Gross, L.~Sweeney, J.~Cohn, F.~de la Torre and S.~Baker.  Face De-identification. In: Senior A. (eds) Protecting Privacy in Video Surveillance. Springer, doi: \url{10.1007/978-1-84882-301-3_8}, 2009.

\bibitem{Gu2019}
X. Gu, W. Luo, M. Ryoo and Y. Lee.  Password-conditioned Anonymization and Deanonymization with Face Identity Transformers. \emph{ArXiv}, \url{https://arxiv.org/abs/1911.11759}, 2019.

\bibitem{Gulrajani2017}
I. Gulrajani, F. Ahmed, M. Arjovsky, V. Dumoulin and A. Courville. Improved training of Wasserstein GANs. In proceedings of the \emph{Advances in Neural Information Processing Systems} conference, pag. 5769--5779, 2017. 

\bibitem{He2017}
K. He, G. Gkioxari, P. Dollar and R. Girshick, Mask r-CNN. In proceedings of the \emph{IEEE Conference on Computer Vision and Pattern Recognition}, pag. 2961--2969,  2017.

\bibitem{He2019}
Z.~He, W. Zul, M. Kan, S. Shan and X. Chen. AttGAN: Facial Attribute Editing by Only Changing What You Want. \emph{IEEE Transactions on Image Processing}, vol. 28, no. 11, pag. 5464--5478, 2019.

\bibitem{Hukkelas2019}
H.~Hukkelas, R.~Mester and F.~Lindseth. DeepPrivacy: A Generative Adversarial Network for Face Anonymization. In proceedings of the \emph{International Symposium on Visual Computing}, Lecture Notes in Computer Science, vol 11844, doi: \url{10.1007/978-3-030-33720-9_44}, 2019.

\bibitem{Isola2017}
P. Isola, J-Y. Zhu, T. Zhou and A. Efros. Image-to-image translation with conditional adversarial networks. In proceedings of the \emph{IEEE International Conference on Biometrics}, doi: \url{10.1109/ICB.2015.7139096}, 2015.

\bibitem{Jourabloo2015}
A.~Jourabloo, X. Yin and X.~Lu. Attribute preserved face de-identification. In proceedings of the \emph{IEEE International Conference on Computer Vision and Pattern Recognition}, doi: \url{10.1109/CVPR.2017.632}, 2017.

\bibitem{Karras2019}
T. Karras,  S. Laine  and T. Aila. A Style-Based Generator Architecture for Generative Adversarial Networks.  In proceedings of the \emph{IEEE International Conference on Computer Vision and Pattern Recognition}, doi: \url{10.1109/CVPR.2019.00453}, 2019.

\bibitem{Korshunova2017}
I.~Korshunova, W. Shi, J. Dambre and L. Theis. Fast face-swap using convolutional neural networks. In proceedings of the \emph{The IEEE International Conference on Computer Vision}, doi: \url{10.1109/ICCV.2017.397} 2017

\bibitem{Kumar2020}
S. Kumar, E. Yaghoubi, A. Das, B. Harish and H. Proen\c{c}a. The P-DESTRE: A Fully Annotated Dataset for Pedestrian Detection, Tracking, Re-Identification and Search from Aerial Devices. \emph{ArXiv}, \url{https://arxiv.org/abs/2004.02782v1}, 2020.

\bibitem{Li2019}
Y.~Li and S.~Lyu. De-identification Without Losing Faces. In proceedings of the \emph{ACM Information Hiding and Multimedia Security Workshop}, doi: \url{10.1145/3335203.3335719}  2019.

\bibitem{Maximov2020}
M.~Maximov, I.~Elezi and L.~Leal-Taix\' {e}. CIAGAN: Conditional Identity Anonymization Generative Adversarial Networks. In proceedings of the \emph{IEEE International Conference on Computer Vision and pattern Recognition}, doi: \url{https://arxiv.org/abs/2005.09544v1}, 2020. (in press)

\bibitem{Meden2018}
B.~Meden, Z.~Emersic, V.~Struc and P.~Peer.  k-Same-Net: k-Anonymity with Generative Deep Neural Networks for Face De-identification.\emph{MDPI Entropy}, vol. 20, no. 60, doi: \url{10.3390/e20010060}, 2018.

\bibitem{Mrityunjay2011}
Mrityunjay and P. Narayanan. The De-Identification Camera. In proceedings of the \emph{Third National Conference on Computer Vision, Pattern Recognition, Image Processing and Graphics}, pag. 192--195, 2011.

\bibitem{Newton2005}
E.~Newton, L.~Sweeney and B.~Malin.  Preserving privacy by de-identifying facial images.\emph{IEEE Transactions on Knowledge and Data Engineering}, vol. 17, no. 2, pag. 232--243, 2005.

\bibitem{Neustaedter2006}
C. Neustaedter, S. Greenberg and M. Boyle. Blur Filtration Fails to Preserve Privacy for Home-Based Video Conferencing. \emph{ACM Transactions on Computer Human Interaction}, vol. 13, issue 1, pag. 1--36 2006.

\bibitem{Phillips2005}
P. Phillips. Privacy operating characteristic for privacy protection in surveillance applications. Audio- and Video-Based Biometric Person Authentication (T. Kanade, A. Jain, and N. Ratha, eds.), Lecture Notes in Computer Science, Springer pag. 869--878, 2005.

\bibitem{Redmon2018}
J. Redmon and A. Farhadi. YOLOv3: An Incremental Improvement. \url{https://arxiv.org/abs/1804.02767v1}, 2018.

\bibitem{Regmi2018}
K. Regmi and A. Borji. Cross-view image synthesis using conditional GANs. In proceedings of the \emph{IEEE International Conference on Computer Vision and Pattern Recognition}, doi: \url{10.1109/CVPR.2018.00369}, 2018.

\bibitem{Ren2017}
S. Ren, K.~He, R.~Girshick and J. Sun. Faster R-CNN: Towards Real-Time Object Detection with Region Proposal Networks. \emph{IEEE Transactions on Pattern Analysis and Machine Intelligence}, vol. 39, no. 6, pag. 1137--1149, 2017.

\bibitem{Ren2018}
Z.~Ren, Y.~Lee and M.~Ryoo. Learning to Anonymize faces for Privacy Preserving Action Detection. In proceedings of the \emph{European Conference on Computer Vision}, pag. 639--655, 2018.

\bibitem{Ronneberger2015}
O.  Ronneberger, P.~Fischer and T.~Brox.  U-Net: Convolutional Networks for Biomedical Image Segmentation. \url{arXiv:1505.04597}, 2015

\bibitem{Ryoo2017}
M. Ryoo, B. Rothrock, C. Fleming and H. Yang. Privacy-preserving human activity recognition from extreme low resolution. In proceedings of the {Thirty-First AAAI Conference on Artificial Intelligence}, pag. 4255--4262, 2017.

\bibitem{Ruiz2018}
N. Ruiz, E. Chong and J. Rehg. Fine-Grained Head Pose Estimation Without Keypoints. In proceedings of the {EEE Conference on Computer Vision and Pattern Recognition (CVPR) Workshops}, doi: \url{10.1109/CVPRW.2018.00281}, 2018.

\bibitem{Samarzija2014}
B. Samarzija and S. Ribaric. An Approach to the De-Identification of Faces in Different Poses. In Pproceedings of \emph{Special Session on Biometrics, Forensics, De-identification and Privacy Protection}, pag. 21--26, 2014.

\bibitem{Schiff2009}
J. Schiff, M. Meingast, D. K. Mulligan, S. Sastry and K. Goldberg. Respectful Cameras: Detecting Visual Markers in Real-time to Address Privacy Concerns. In proceedings of the \emph{IEEE/RSJ International Conference on Intelligent Robots and Systems}, doi: \url{10.1109/IROS.2007.4399122}, 2009.

\bibitem{Seo2008}
J.~Seo, S.~Hwang and Y-H. Suh. A Reversible Face De-Identification Method based on Robust Hashing. In proceedings of the \emph{International Conference on Consumer Electronics}, doi: \url{10.1109/ICCE.2008.4587904}, 2008.

\bibitem{Senior2009}
A. Senior. Privacy Protection in a Video Surveillance System. \emph{Protecting Privacy in Video Surveillance}, pag. 35--47,  doi: \url{10.1007/978-1-84882-301-3_3}, 2009.

\bibitem{Sweeney2002}
L. Sweeney.  k-anonymity: a model for protecting privacy. \emph{International Journal on Uncertainty, Fuzziness, and Knowledge-Based Systems}, vol. 10, no. 5, pag. 557--570, 2002.

\bibitem{Shen2017}
W.~Shen and R.~Liu. Learning residual images for face attribute manipulation. In proceedings of the \emph{IEEE International Conference on Computer Vision and Pattern Recognition}, doi: \url{10.1109/CVPR.2017.135}, 2017.

\bibitem{Sun2018}
Q. Sun, A. Tewari, W. Xu, M. Fritz, C. Theobalt and B. Schiele. A hybrid model for identity obfuscation by face replacement. In proceedings of the \emph{European Conference on Computer Vision}, pages 570--586, 2018. 

\bibitem{Sun2018b}
Q. Sun, L. Ma, S. Joon, L. Van Gool, B. Schiele and M. Fritz. Natural and effective obfuscation by head inpainting. In proceedings of the  \emph{IEEE International Conference on Computer Vision and Pattern Recognition}, doi: \emph{10.1109/CVPR.2018.00530}, 2018.

\bibitem{Winkler2010}
T. Winkler and B. Rinner. TrustCAM: Security and Privacy-Protection for an Embedded Smart Camera based on Trusted Computing. In proceedings of the \emph{Seventh IEEE International Conference on Advanced Video and Signal Based Surveillance}, pag. 593 -- 600, 2010.

\bibitem{Xiao2018}
T. Xiao, J. Hong and J. Ma. ELEGANT: Exchanging Latent Encodings with GAN for Transferring Multiple Face Attributes. In proceedings of the \emph{European Conference on Computer Vision}, doi: \url{ 10.1007/978-3-030-01249-6_11}, 2018.

\bibitem{Yamac2019}
M. Yamac, M. Ahishali, N. Passalis, J. Raitoharju, B. Sankur  and M. Gabbouj.  Reversible Privacy Preservation using Multi-level Encryption and Compressive Sensing. In proceedings of the \emph{27th European Signal Processing Conference}, doi: \url{10.23919/EUSIPCO.2019.8903056}, 2019.

\bibitem{Yan2019}
B.~Yan, M.~Pei and Z.~Nie.  Attributes Preserving Face De-Identification. In proceedings of the \emph{IEEE International Conference on Computer Vision}, dos: \url{10.1109/ICCVW.2019.00154}, 2019.

\bibitem{Zhang2016}
K.~Zhang,  Z.~Zhang,  Z. ~Li  and  Y.~Qiao,  Joint  face  detection  and alignment   using   multitask   cascaded   convolutional   networks. \emph{IEEE Signal Processing Letters}, vol. 23, no. 10, pag. 1499--1503, 2016.

\bibitem{Zhang2017}
H. Zhang, T. Xu, H. Li, S. Zhang, X. Wang, X. Huang and D. Metaxas. Stack-GAN: Text to photo-realistic image synthesis with stacked generative adversarial networks. In proceedings of the \emph{IEEE International Conference on Computer Vision}, doi: \url{10.1109/ICCV.2017.629}, 2017.

\bibitem{Zheng2016}
L. Zheng, Z. Bie, Y. Sun, J. Wang, C. Su, S. Wang and Q. Tian. MARS: A Video Benchmark for Large-Scale Person Re-Identification. In proceedings of the \emph{European Conference on Computer Vision}, part VI, pag. 868--884, 2016.

\bibitem{Zhong2016}
Y. Zhong, J. Sullivan and H. Li. Face attribute prediction using off-the-shelf deep learning networks. CoRR, \url{abs/1602.03935}, 2016.

\bibitem{Zhu2017}
S. Zhu, R. Urtasun, S. Fidler, D. Lin and C. Loy. Be your own prada: Fashion synthesis with structural coherence. In proceedings of the \emph{IEEE International Conference on Computer Vision}, doi: \url{10.1109/ICCV.2017.186}, 2017.

\end{thebibliography}

}

\end{document}